\documentclass[twoside,11pt]{article}
\pdfoutput=1
\usepackage{jmlr2e-arxiv}

\usepackage{amsmath, amssymb}
\usepackage{algorithm,algorithmic}

\usepackage{multirow}
\usepackage{colortbl}
\usepackage{subfigure}

\definecolor{orange}{rgb}{1,0.9,0.65}
\definecolor{gr}{rgb}{0.9,1,0.6}
\definecolor{bl}{rgb}{0.9,0.8,1}
\definecolor{bg}{rgb}{0.8,0.9,1}
\definecolor{orr}{rgb}{1,0.85,0.4}
\definecolor{grr}{rgb}{0.8,1,0.5}
\definecolor{blr}{rgb}{0.85,0.45,1}
\definecolor{bgr}{rgb}{0.5,0.9,1}
\definecolor{gry}{rgb}{0.92,0.92,0.92}

\def\eop {{\noindent\framebox[0.5em]{\rule[0.25ex]{0em}{0.75ex}}}}
\def\be {\begin{equation}}
\def\ee {\end{equation}}
\def\bea {\begin{eqnarray*}}
\def\eea {\end{eqnarray*}}
\def\beas {\begin{eqnarray*}}
\def\eeas {\end{eqnarray*}}
\newtheorem{theorem-ap}{Theorem}

\newtheorem{claim-ap}{Claim}

\newcommand{\by}{\mathbf{y}}
\newcommand{\bo}{\mathbf{o}}
\newcommand{\bs}{\mathbf{s}}
\newcommand{\ba}{\mathbf{a}}
\newcommand{\bx}{\mathbf{x}}
\newcommand{\bz}{\mathbf{z}}
\newcommand{\bb}{\mathbf{b}}
\newcommand{\R}{\mathbb{R}}

\renewcommand{\S}{{\cal S}}
\newcommand{\X}{{\cal X}}
\newcommand{\Y}{{\cal Y}}

\jmlrheading{1}{2012}{??}{??}{??}{Tamir Hazan, Alexander G. Schwing, David McAllester and Raquel Urtasun}

\ShortHeadings{Blending Learning and Inference in Structured Prediction}{Hazan, Schwing, McAllester and Urtasun}
\firstpageno{1}

\begin{document} 

\title{Blending Learning and Inference in Structured Prediction}

\author{Tamir Hazan \\ tamir@cs.haifa.ac.il \\ University of Haifa \\ Carmel,   Haifa 31905, Israel
\and Alexander G. Schwing \\ aschwing@inf.ethz.ch\\ ETH Zurich \\ Universitaetstrasse 6, 8092 Zurich, Switzerland
\and David McAllester \\ mcallester@ttic.edu \\ TTI Chicago \\ 6045 S. Kenwood Ave., \\Chicago, IL 60637 USA
\and Raquel Urtasun \\rurtasun@ttic.edu \\ TTI Chicago \\ 6045 S. Kenwood Ave., \\Chicago, IL 60637 USA
}


\maketitle

\begin{abstract}

In this paper we derive an efficient algorithm to learn the parameters of structured predictors in general graphical models. This algorithm blends the learning and inference tasks, which results in a significant speedup over traditional approaches, such as conditional random fields and structured support vector machines. For this purpose we utilize the structures of the predictors to describe a low dimensional structured prediction task which encourages local  consistencies within the different structures while learning the parameters of the model. Convexity of the learning task provides the means to enforce the consistencies between the different parts. The inference-learning blending algorithm that we propose is guaranteed to converge to the optimum of the low dimensional primal and dual programs. Unlike many of the existing approaches, the inference-learning blending allows us to learn efficiently high-order graphical models, over regions of any size, and very large number of parameters. We demonstrate the effectiveness of our approach, while presenting state-of-the-art results in stereo estimation, semantic segmentation, shape reconstruction, and indoor scene understanding. 

\end{abstract}

\section{Introduction}
\label{sec:intro}

Structured prediction is an effective framework to reason about real-life problems since it provides the means to map objects $x$  to labels $y$. Typically, the label space has rich internal structure, e.g., semantic segmentations or depth estimations, and the set of possible labels for a given object is typically exponential in its size. Ideally, one would want to make joint predictions on the structured labels instead of simply predicting each element independently, as this additionally accounts for the statistical correlations between label elements, as well as between training objects and their labels. These properties make structured prediction appealing for a wide range of applications in computer vision \cite{s:felzenswalb10, Szeliski07} as well as in natural language processing \cite{Koo10} and computational biology \cite{Yanover07, Sontag08}. 

Learning the parameters of structured predictors greatly influences the prediction accuracy. Several models have been recently proposed, including log-likelihood models such as conditional random fields (CRFs, \cite{Lafferty01}), and structured support vector machines  (structured SVMs) such as maximum-margin Markov networks (M3Ns \cite{Taskar04}) and structured output learning (\cite{Tsochantaridis04}). For CRFs, the parameters estimation is done by minimizing a convex function composed of a negative log-likelihood loss and a regularization term. Learning the parameters with structured SVMs is done by minimizing the convex regularized structured hinge loss.   
  
Despite the convexity of the objective functions, finding the optimal parameters of these models can be computationally expensive since it involves comparing the training labels with the predicted labels, which are inferred out of exponentially many possible labels. When the label structure corresponds to a tree, exact inference can be done efficiently by using belief propagation as a subroutine; The sum-product algorithm is typically used in CRFs and the max-product algorithm in structured SVMs. In general, when the label structure corresponds to a general graph, one cannot compute the objective nor the gradient exactly, except for some special cases in structured SVMs, such as matching and sub-modular functions  (e.g., \cite{Taskar06}). Therefore, one  usually resorts to approximate inference algorithms (cf. \cite{Finley08,Levin06}). However, the approximate inference algorithms are computationally expensive to be used as a subroutine of the learning algorithm, therefore they cannot be applied efficiently to learn the parameters of structured predictors. 
 
In this paper we derive an efficient algorithm that blends the learning and inference tasks, which results in a significant speedup over traditional approaches, such as conditional random fields and structured support vector machines. First, we define the extended log-loss, which relates the log-loss of CRFs and the hinge-loss of structured SVMs through a temperature parameter. As a consequence we show that CRFs smoothly approximate structured SVMs  in low temperatures. We then present a low dimensional upper bound to the extended log-loss that decomposes along the regions of a graphical model. The decomposed upper bound allows to blend the learning and inference tasks, while their consistency is preserved through convexity. We conclude with the dual aspects of learning and inference, showing that learning relates to moment matching constraints and inference relates to probabilities marginalization constraints using the entropy selection rule. 

The rest of the paper is organized as follows. In Section \ref{sec:bg} we review parameter learning methods focusing on its most common models, CRFs and structured SVMs. We present the extended log-loss which relates CRFs and structured SVMs in Section \ref{sec:sp}, and describe the necessary background about graphical models and approximate inference in Section \ref{sec:bg-graphs}. We then describe our extended log-loss upper bound, along with the learning-inference blending algorithm which minimizes this low dimensional structured prediction task with block gradient descent steps, see Section \ref{sec:upper}. We demonstrate the effectiveness of our approach in Section \ref{sec:experiments}, describing our state-of-the-art results in stereo estimation, semantic segmentation, shape reconstruction, and indoor scene understanding. Next, in Section \ref{sec:duality}, we elaborate on the dual aspects of the low dimensional structured prediction task, and relating it to learning pseudo moment matching with inferred beliefs that agree on their marginal probabilities with respect to decomposed entropy selection rule. This perspective gives rise to tighter decomposed bounds to the extended log-loss, that utilize fractional entropy with nonnegative covering numbers in Section \ref{sec:ext}.

\section{Background}
\label{sec:bg}

Structured prediction typically involves objects $x \in \X$ and their labels $y \in \Y$. The structure is usually incorporated into the labels which may be sequences, trees, grids, or other high-dimensional objects with internal structure. For every object $x$, its possible labels are described by a feature function  $\phi_k: \X \times \Y \rightarrow \R$. Our goal is to learn the parameters of the linear prediction rule $$y_{w}(x) = \arg \max_{y \in \Y} \sum_k w_k \phi_k(x,y)$$ with parameters $w \in \R^K$, such that $y_{w}(x)$ is a good approximation to the true label of $x$. Intuitively one would like to learn the parameters of structured predictors by minimizing the training loss $\ell(y,y_{w}(x))$ incurred by using $w$ to predict the label of $x$, given that the true label is $y$. Since the prediction is norm-insensitive this method can lead to over fitting. Therefore, given a training set $(x,y) \in \S$, the parameters $w$ are usually learned by minimizing a norm-dependent loss 
\begin{equation}
\label{eq:reg-loss}
\sum_{(x,y) \in \S} \bar \ell (w,x,y) + \frac{C}{2} \|w\|_2^2.
\end{equation}
The surrogate loss function $\bar{\ell}(w,x,y)$ typically upper bounds the true loss $\ell(y,y_w(x))$. The surrogate loss function determines the learning setting for the prediction problem, e.g., structured SVMs and CRFs. 

Structured SVMs aim at minimizing the surrogate hinge loss, presented by \cite{Taskar04, Tsochantaridis06}:
$$\bar \ell_{hinge}(w,x,y) = \max_{\hat y \in \Y} \Big\{ \ell(y,\hat y) +  \sum_k w_k \phi_k(x,\hat y) - \sum_k w_k \phi_k(x,y)\Big\}.$$        
The structured hinge loss upper bounds the true loss function. It corresponds to a maximum-margin approach that linearly penalizes predictions $y_w(x)$ that violates a training pair $(x,y) \in \S$ by more than $\ell(y,y_{w}(x))$, i.e., $\sum_k w_k \phi_k(x,y) < \ell(y,y_{w}(x)) + \sum_k w_k \phi_k(x,y_{w}(x))$. 

The second loss function that we consider is based on log-linear models, and is commonly used in CRFs, defined by \cite{Lafferty01}. To endure it upper bounds the true loss, we define the loss adjusted (conditional) Gibbs distribution 
\begin{equation}
\label{eq:gibbs}
p_{(x,y)}(\hat y ; w) \propto \exp\Big(\ell(y,\hat y) + \sum_k w_k \phi_k(x,\hat y)\Big).
\end{equation}
The Gibbs distribution provides a probabilistic prediction rule, which scales the different predictions according to their prediction value. The surrogate loss function is then the negative log-likelihood under the parameters $w$
$$\bar \ell_{log}(w,x,y) =  - \log p_{(x,y)}(y ;w).$$
The log-loss upper bounds the structured hinge loss, since $\max_{\hat y} \{\ell(y,\hat y) + \sum_k w_k \phi_k(x,\hat y)\} \le \log (\sum_{\hat y} \exp(\ell(y,\hat y) + \sum_k w_k \phi_k(x,\hat y)))$, and as a result it also upper bounds the true loss $\ell(y,y_w(x))$. In structured SVMs and CRFs a convex loss function and a convex regularization are minimized, and gradient based methods can be used to learn their optimal parameters $w$.

\subsection{One parameter extension of CRFs and Structured SVMs}
\label{sec:sp}

In CRFs one aims to minimize  the regularized negative log-likelihood  of the distribution $p_{(x,y)}(\hat y ;w)$. The regularized log-loss is a convex and smooth function and its parameters learning, using gradient decent, measures the disagreements between the inferred labels and the training labels 
$$\frac{\partial \; \bar \ell_{\scriptsize \mbox{log}}(x,y,w)}{\partial w_k}  =  \sum_{\hat y \in \Y} p_{(x,y)}(\hat y ; w) \phi_k(x,\hat y) - \phi_k(x,y).$$
The computational complexity of CRFs is governed by the gradient computation.

Structured SVMs aim at minimizing the regularized hinge loss $\bar \ell_{hinge}(w,x,y)$. The hinge loss involves the max-function, which is a convex and non-smooth function. However every convex function has subgradients, i.e. supporting hyperplanes to its epigraph (cf. \cite{Rockafellar70}). The subgradients generalize the concept of the gradient since a convex function is smooth if and only if has a single subgradient, namely its gradient. Danskin's theorem (e.g., \cite{Bertsekas03}, Theorem 4.5.1) states that the subgradients of the max-function correspond to probability distributions $p_{(x,y)}(y^* ; w)$ over the optimal set $\Y^* = \arg \max_{\hat y \in \Y} \{\ell(y,\hat y) + \sum_k w_k \phi_k(x,\hat y) \}$. Therefore learning the structured SVMs parameters using the subdifferential of the hinge-loss amounts to measure the disagreements between the inferred labels and the training labels 
$$\frac{\partial \; \bar \ell_{hinge}(x,y,w)}{\partial w_k}  =  \sum_{y^* \in \Y^*} p_{(x,y)}(y^* ; w) \phi_k(x,y^*) - \phi_k(x,y).$$
Unlike the smooth case, a subgradient does not necessarily points towards a direction of descent. Thus subgradient methods are not monotonically decreasing, and their optimal solution is recovered from the algorithm sequence.  

It is convenient to deal with both learning tasks for structured predictors (i.e., structured SVMs and CRFs) as two instances of the same framework. We follow the path of \cite{Pletscher10,Hazan10-nips}, and introduce a temperature parameter to our loss adjusted probability model, namely 
$$p_{(x,y)}(\hat y; w , \epsilon) \propto \exp\Big((\ell(y,\hat y) + \sum_k w_k \phi_k(x,\hat y)) \Big/ \epsilon \Big).$$ 
This parameter controls the variance of the probability distribution: it tends towards the uniform distribution when $\epsilon \rightarrow \infty$, and to the zero-one distribution when $\epsilon \rightarrow 0$. We introduce a temperature extension of the log-loss function $$ \bar \ell_{\epsilon\mbox{-}log} (w,x,y) \stackrel{def}{=} - \epsilon \log p_{(x,y)}(\hat y ; w, \epsilon).$$
Similarly to the log-loss, the extended log-loss upper bounds the structured hinge loss, and consequently it also upper bounds the true loss. The extended log-loss generalizes  the hinge-loss and the log-loss in the same way the norm function $\| \cdot \|_{1/\epsilon}$ generalizes the sum-function $\| \cdot \|_1$ and the max-function $\| \cdot \|_\infty$. In particular, for $\epsilon=1$ the extended log-loss reduces to the log-loss and for $\epsilon=0$ it reduces to the hinge-loss. Moreover, when $\epsilon \rightarrow 0$ the exnteded log-loss smoothly approximates the hinge-loss, in the same way the $\ell_{1/\epsilon}$-norm is a smooth approximation of the $\ell_\infty$-norm. 

One can learn the optimal parameters of the one-parameter extension of CRFs and structured SVMs using gradient descent, which measures the disagreements between the inferred labels and the training labels
\begin{equation}
\label{eq:grad}
\frac{\partial \; \ell_{\epsilon \mbox{-} log}(x,y,w)}{\partial w_k}  = \sum_{\hat y \in \Y} p_{(x,y)}(\hat y ; w, \epsilon) \phi_k(x,\hat y) - \phi_k(x,y), 
\end{equation}
where $p_{(x,y)}(\hat y ; w, \epsilon)$ is the loss adjusted Gibbs distribution over the possible labels $\hat y \in \Y$. When $\epsilon \rightarrow 0$ this probability distribution gets concentrated around its maximal values, since all its elements are raised to the power of a very large number (i.e., $1/\epsilon$). For $\epsilon = 0$ this distribution is supported on the maximal elements $\Y^*$, and we attain a structured SVM subgradient.

\subsection{Structured prediction in graphical models}
\label{sec:bg-graphs}

In many real-life problems the labels $y \in \Y$ are $n$-tuples, $y=(y_1,...,y_n)$, hence there are exponentially many labels in $\Y$. The features usually describe relations between subsets of elements $r \subset \{1,...,n\}$, also called regions. We denote by ${\cal R}_k$ the regions of the feature $\phi_k(x,y)$. The features are functions of their regions labels $y_r \subset \{y_1,...,y_n\}$:
\begin{equation}
\label{eq:feature}
\phi_k(x,y_1,...,y_n) = \sum_{r \in {\cal R}_k} \phi_{k,r}(x,y_r).
\end{equation}
Similarly, we consider region-based loss functions $\ell(y,\hat y) = \sum_{r \in {\cal R}_\ell} \ell_r(y_r,\hat y_r)$. The loss function, as well as the features define hypergraphs whose nodes represent the $n$ labels indexes, and the regions ${\cal R} = \cup_k {\cal R}_k \cup {\cal R}_\ell$ correspond to its hyperedges. A convenient way to represent a hypergraph is by its region graph. A region graph is a directed graph whose nodes represent the regions and its direct edges correspond to the inclusion relation, i.e., a directed edge from node $r$ to $s$ is possible only if $s \subset r$. We adopt the terminology where $P(r)$ and $C(r)$ stand for all nodes that are parents and children of the node $r$, respectively. 

The Hammersley-Clifford theorem (e.g., \cite{Lauritzen96}) asserts that the Gibbs distributions $p_{(x,y)}(\hat y ; w)$ defined in Equation (\ref{eq:gibbs}) corresponds to a Markov random field (MRF) whose statistical independencies are described by the joint hypergraph. These independencies are determined by the Markov property: Two nodes in the graph are conditionally independent when they are separated by observed nodes. \cite{Yedidia05} show that whenever the region graph is bipartite and has no cycles, the Markov property provides a low dimensional representation of the Gibbs distribution using its marginal probabilities $p_{(x,y)}(\hat y_r ; w) = \sum_{\hat y \setminus \hat y_r} p_{(x,y)}(\hat y ; w)$, namely

\begin{equation}
\label{eq:pr}
p_{(x,y)}(\hat y ; w) = \prod_{r \in {\cal R}} p_{(x,y)}(\hat y_r ; w )^{1-|P(r)|}.
\end{equation}
When the bipartite region graph has no cycles one can use the belief propagation algorithm to efficiently infer the marginal probabilities $p_{(x,y)}(\hat y_r ; w, \epsilon)$, for every $\epsilon \ge 0$, without performing exponentially many operations:
\begin{algorithm}
\caption{Belief Propagation}
Set ${\cal K}_r = \{k: r \in {\cal R}_k\}$. For every $(x,y)$ set $\theta_r(\hat y_r) = \ell_r( y_r, \hat y_r) + \sum_{k \in {\cal K}_r} w_k \phi_{k,r}(x,\hat  y_r)$. \\
Repeat until convergence: 
\begin{enumerate}
\item[] $\mu_{\alpha \rightarrow i}(y_i) =  \epsilon \log\Big( \sum_{y_\alpha \setminus y_i} \exp\big((\theta_\alpha(y_\alpha) + \sum_{j \in C(\alpha)\setminus i} \lambda_{j \rightarrow \alpha}(y_j)) \big/ \epsilon \big) \Big)$
\item[] $\lambda_{i \rightarrow \alpha}(y_i) = \theta_i(y_i) + \sum_{\beta \in P(i) \setminus \alpha} \mu_{\beta \rightarrow i}(y_i)$
\end{enumerate}
Output:
\begin{enumerate}
\item[] $b_i(y_i) \propto \exp \big(\theta_i(y_i) + \sum_{\alpha \in P(i)} \mu_{\alpha \rightarrow i}(y_i) \big)$
\item[] $b_\alpha(y_\alpha) \propto  \big(\theta_\alpha(y_\alpha) + \sum_{i \in C(\alpha)} \lambda_{i \rightarrow \alpha}(y_i) \big)$
\end{enumerate}
\end{algorithm}

When restricting to bipartite region graphs it has two types of regions: outer regions, i.e., regions that are not contained by other regions, and inner regions. To distinct between these regions we denote outer regions by $\alpha$ and inner regions by $i$. The marginal probabilities $p_{(x,y)}(\hat y_r ; w, \epsilon)$ appear in the beliefs $b_r(\hat y_r)$. In general, when the region graph has cycles the belief propagation algorithm is not guaranteed to output the marginal probabilities. Nevertheless, when it converges it provides beliefs that agree on their marginal probabilities, namely $\sum_{y_\alpha \setminus y_i} b_\alpha(y_\alpha) = b_i(y_i)$. In some cases   the belief propagation algorithm infers beliefs $b_r(\hat y_r)$ which approximate well the marginal probabilities, while in other cases it produces non-accurate results or might fail to converge. Recently, there was an extensive effort trying to fix the drawbacks of the belief propagation algorithm, and convergence of belief propagation type algorithms is attained using techniques from convex duality, e.g., \cite{Heskes06, Hazan10}. 

\begin{algorithm}
\caption{Norm-Product Belief Propagation}
Set $\hat c_i = c_i + \sum_{\alpha \in P(i)} c_\alpha$. Repeat until convergence: 
\begin{enumerate}
\item[] $\mu_{\alpha \rightarrow i}(y_i) =  \epsilon c_\alpha \log \Big( \sum_{y_\alpha \setminus y_i} \exp \big((\theta_\alpha(y_\alpha) + \sum_{j \in C(\alpha)\setminus i} \lambda_{j \rightarrow \alpha}(y_j)) \big/ /\epsilon c_\alpha \big) \Big)$
\item[] $\lambda_{i \rightarrow \alpha}(y_i) = \frac{c_\alpha}{\hat c_i} \Big( \theta_i(y_i) + \sum_{\beta \in P(i)} \mu_{\beta \rightarrow i}(y_i) \Big)  - \mu_{\alpha \rightarrow i}(y_i)$
\end{enumerate}
Output:
\begin{enumerate}
\item[] $b_i(y_i) \propto \exp \big(\theta_i(y_i) + \sum_{\alpha \in P(i)} \mu_{\alpha \rightarrow i}(y_i) \big)^{1/\epsilon \hat c_i}$
\item[] $b_\alpha(y_\alpha) \propto  \exp \big(\theta_\alpha(y_\alpha) + \sum_{i \in C(\alpha)} \lambda_{i \rightarrow \alpha}(y_i) \big)^{1/\epsilon c_\alpha}$
\end{enumerate}
\end{algorithm}
The norm-product algorithm reduces to belief propagation when setting its coefficients to $c_r = 1-|P(r)|$. These coefficients also appear when constructing a probability distribution from its marginal probabilities in graph without cycles, and are known as the Bethe coefficients. We refer the interested reader to \cite{Wainwright08, Koller09} for more details. 

The norm-product is guaranteed to converge whenever $\epsilon, c_r \ge 0$ to beliefs that agree on their marginal probabilities. Typically its inferred beliefs approximate the marginal probabilities as well as the belief propagation approximations. Thus in its various forms it can be used to approximately learn the parameters of structured predictors as well as its gradient, described in Equation (\ref{eq:grad}). However, iteratively executing the norm-product algorithm as a sub-procedure to compute the gradient is computationally intractable and this method was not been widely used. In the following we explore convex upper bounds to the extended log-loss. These upper bounds are decomposed according to the graphical model, thus provide the means to blend the learning and inference tasks. This provides with the means to efficiently learn the parameters of a graphical model, based on (dual) decomposition, which targets moment matching instead of the time consuming estimations of the probability $p_{(x,y)}(\hat y ; w, \epsilon)$.

\section{Loss upper bounds and decompositions} 
\label{sec:upper}

The computational complexity of structured prediction depends on the extended log-loss. In complex models, the labels $y=(y_1,..,y_n)$ may enumerate structures which are exponential in $n$. Restricting to graph based features, we decompose the extended log-loss with respect to their corresponding regions. These low dimensional decomposition upper bounds the extended log-loss, thus its minimization implicitly also minimizes the structured prediction task. 
\begin{theorem}
\label{theorem:upper}
Consider region based features, defined in Equation (\ref{eq:feature}) and their corresponding region graphs. Assume the loss function decomposes with respect to its regions $\R_\ell$. Let ${\cal K}_r$ to be the set of $\{k : r \in {\cal R}_k\}$ and denote by $P(r)$ and $C(r)$ the parents and children of a region in the joint region graph. Consider, for every $(x,y) \in \S, r \in {\cal R}, p \in P(r)$, the real valued vector $\lambda_{(x,y),r \rightarrow p}(y_r)$ and the potential function $\theta_{(x,y),r}(\hat y_r ; w) = \ell_r(y_r , \hat y_r) + \sum_{k \in {\cal K}_r} w_k \phi_{k,r}(x,\hat y_r)$. Define the parameterized beliefs
$$b_{(x,y),r}(\hat y_r ; w,\lambda,\epsilon) \propto \exp \Big((\theta_{(x,y),r}(\hat y_r ; w) + \sum_{c \in C(r)} \lambda_{(x,y),c \rightarrow r}(\hat y_c) - \sum_{p \in P(r)} \lambda_{(x,y),r \rightarrow p}(\hat y_r)) \Big/\epsilon \Big)$$ 
Then the loss functions $\bar \ell_{r,\epsilon\mbox{-}log} (w,x,y_r) = - \epsilon \log b_{(x,y),r} \big(\hat y_r ; w, \lambda, \epsilon \big)$ upper bound the extended log-loss, i.e., 
$$\bar \ell_{\epsilon\mbox{-}log} (w,x,y)  \le \sum_{r \in {\cal R}} \bar \ell_{r,\epsilon\mbox{-}log} (w,x,y_r).$$ 
The low dimensional structured prediction program $\min_{w, \lambda} \sum_{(x,y) \in \S} \sum_{r \in {\cal R}} \bar \ell_{r,\epsilon\mbox{-}log} (w,x,y_r) + \frac{C}{2} \|w\|_2^2$ is an unconstrained convex function of $w,\lambda$ thus it attains its minimum when the (sub)gradients vanish. 
\end{theorem}
{\bf Proof:} 
We consider the case $\epsilon > 0$, while the case $\epsilon = 0$ follows using a limit argument. Following the loss definition,  
$$
\bar \ell_{\epsilon\mbox{-}log} (w,x,y)  = -\sum_{r \in {\cal R}} \theta_{(x,y),r}(y_r ; w) + \epsilon \log \Big(\sum_{\hat y} \exp(\sum_{r \in {\cal R}} \theta_{(x,y),r}(\hat y_r ; w) /\epsilon ) \Big) 
$$
Consider the parametrized potential function 
$$\theta_{(x,y),r}(\hat y_r ; w, \lambda) = \theta_{(x,y),r}(\hat y_r ; w) + \sum_{c \in C(r)} \lambda_{(x,y),c \rightarrow r}(\hat y_c) - \sum_{p \in P(r)} \lambda_{(x,y),r \rightarrow p}(\hat y_r).
$$ 
Since $\sum_{r \in {\cal R}} \sum_{p \in P(r)} \lambda_{(x,y), r \rightarrow p}(y_r) - \sum_{r  \in {\cal R}} \sum_{c \in C(r)} \lambda_{(x,y), c \rightarrow r}(y_c) \equiv 0$ the sum of low dimensional extended log-loss takes the form 
$$ 
\sum_{r \in {\cal R}} \ell_{r,\epsilon\mbox{-}log} (w,x,y_r) = -\sum_{r \in {\cal R}} \theta_{(x,y),r}( y_r ; w) + \epsilon  \sum_{r \in {\cal R}}  \log \Big(\sum_{\hat y_r} \exp(\theta_{(x,y),r}(\hat y_r ; w, \lambda) /\epsilon ) \Big) 
$$
The theorem then follows as the regions span the set of variables $\cup_{r \in {\cal R}} r = \{1,...,n\}$, thus:
$$\sum_{\hat y} \exp(\sum_{r \in {\cal R}} \theta_{(x,y),r}(\hat y_r ; w)) = \sum_{\hat y} \prod_{r \in {\cal R}}  \exp(\theta_{(x,y),r}(\hat y_r ; w, \lambda)) \le \prod_{r \in {\cal R}} \sum_{\hat y_r} \exp(\theta_{(x,y),r}(\hat y_r ; w, \lambda))$$
$\Box$

Performing block coordinate descent on the low dimensional structured prediction program objective in Theorem \ref{theorem:upper} requires minimizing a block of variables while holding the rest fixed. We begin by describing how to infer the optimal set of variables $\lambda_{(x,y),r \rightarrow p}(\hat y_r)$ that are related to a region and its parents in the graphical model. 
\begin{lemma}
\label{lemma:lambda}
Inference, for every region $r$, of the optimal $\lambda_{(x,y),r \rightarrow p}(\hat y_r)$ for every $p \in P(r), \hat y_r \in \Y_r, (x,y) \in \S$ in the low dimensional structured prediction program of Theorem \ref{theorem:upper} follows the update rules
\begin{eqnarray*}
 \mu_{(x,y), p \rightarrow r}(\hat y_r) &=& \epsilon \log \Big(\sum_{\hat y_p \setminus \hat y_r} \exp \big( (\theta_{(x,y),p}(\hat y_p ; w) + \sum_{c \in C(p) \setminus r} \lambda_{(x,y),c \rightarrow p} (\hat y_c) - \sum_{p' \in P(p)} \lambda_{(x,y),p \rightarrow p'}(\hat y_p)) \big/ \epsilon \big) \Big)\\ 
\lambda_{(x,y),r \rightarrow p} (\hat y_r) &=&  \frac{1}{1+|P(r)|} \Big(\theta_{(x,y),r}(\hat y_r ; w) + \sum_{c \in C(r)} \lambda_{(x,y),c \rightarrow r}(\hat y_c) + \sum_{p' \in P(r)} \mu_{(x,y),p' \rightarrow r} (\hat y_r)  \Big) - \mu_{(x,y),p \rightarrow r} (\hat y_r)
\end{eqnarray*}
Moreover, since the program is not strictly convex, the optimal solutions can be achieved for every additive shifts, namely $\lambda_{(x,y),r \rightarrow p} (\hat y_r) - c_{(x,y),r \rightarrow p}$ are also optimal solutions for every constant $c_{(x,y),r \rightarrow p}$.  
\end{lemma}
{\bf Proof:} The loss minimization program in Theorem \ref{theorem:upper} is convex and unconstrained, therefore the optimum is achieved when the (sub)gradient vanishes. Using the definition of the parametrized potential function in Theorem \ref{theorem:upper} proof, for $\epsilon = 0$ we define $b_{(x,y),r}(\hat y_r ; w, \lambda, \epsilon)$ to be a probability distribution over the maximal elements $\Y_r^* = \mbox{argmax}_{\hat y_r \in \Y_r} \{\theta_{(x,y),r}(\hat y_r ; w, \lambda) \}$. Then the gradient with respect to $\lambda_{(x,y),r \rightarrow p}(\hat y_p)$ takes the form
$$\frac{\partial }{\partial \lambda_{(x,y),r \rightarrow p}(\hat y_r)} = \sum_{\hat y_p \setminus \hat y_r} b_{(x,y),p}(\hat y_p ; w, \lambda, \epsilon) - b_{(x,y),r}(\hat y_r ; w, \lambda, \epsilon)$$ 

The optimal dual variables are those for which the gradient vanishes, i.e., the corresponding beliefs agree on their marginal probabilities. When setting $\mu_{(x,y),p \rightarrow r}(\hat y_r)$ as above, the marginalization of $b_{(x,y),p}(\hat y_p ; w, \lambda, \epsilon)$ satisfy
$$\sum_{\hat y_p \setminus \hat y_r} b_{(x,y),p}(\hat y_p ; w, \lambda, \epsilon) \propto \exp \Big( (\mu_{(x,y),p \rightarrow r}(\hat y_r) + \lambda_{(x,y),r \rightarrow p}(\hat y_r)) \Big/ \epsilon \Big) $$
Therefore, by taking the logarithm, the gradient vanishes whenever the beliefs numerators agree up to an additive constant 
$$
\mu_{(x,y),p \rightarrow r}(\hat y_r) + \lambda_{(x,y),r \rightarrow p}(\hat y_r) = \theta_{(x,y),r}(\hat y_r ; w, \lambda)$$
whereas $\theta_{(x,y),r}(\hat y_r ; w, \lambda)$ depends on $\sum_{p \in P(r)} \lambda_{(x,y),r \rightarrow p}(\hat y_r)$. To isolate this quantity we sum both sides with respect to $p \in P(r)$, thus we are able to obtain 
$$(1+|P(r)|) \sum_{p \in P(r)} \lambda_{(x,y),r \rightarrow p}(\hat y_r) = |P(r)| (\theta_{(x,y),r}(\hat y_r ; w) + \sum_{c \in C(r)} \lambda_{(x,y),c \rightarrow r}(\hat y_c) ) - \sum_{p \in P(r)} \mu_{(x,y),p \rightarrow r}(\hat y_r)$$ 
Plugin it into the above equation results in the desired block dual descent update rule, i.e., $ \lambda_{(x,y),r \rightarrow p}(\hat y_r)$ for which the partial derivatives vanish.  \eop \\

The above lemma describes an analytic solution for the optimal $\lambda_{(x,y),r \rightarrow p}(\hat y_r)$, that are computed in the  block coordinate steps of the algorithm. In practice, block coordinate descent with analytic steps provides a significant speedup over conventional gradient methods and can be parallelized and distributed easily, as shown by \cite{Schwing11}. Unfortunately, we are not able to analytically find the optimal $w_k$ while holding the rest fixed, thus we perform a step in the direction of the negative (sub)gradient. 

\begin{lemma}
\label{lemma:theta}
Learning the optimal parameters $W$ of the low dimensional structured prediction program in Theorem \ref{theorem:upper} follow the (sub)gradient  
$$
\frac{\partial}{\partial w_k} =  \sum_{(x,y) \in \S} \sum_{r \in {\cal R}}  \Big(\sum_{\hat y_r \in \Y_r} b_{(x,y),r}(\hat y_r ; w, \lambda, \epsilon) \phi_{k,r}(x,\hat y_r) - \phi_{k,r}(x,y_r) \Big) + C w_k. 
$$
\end{lemma} 
 {\bf Proof:} Recall the definition of $ b_{(x,y),r}(\hat y_r ; w, \lambda, \epsilon)$, for $\epsilon > 0$, in Theorem \ref{theorem:upper}. For $\epsilon = 0$ we use its definition in Lemma \ref{lemma:lambda} and Danskin theorem. \eop \\
 
The computational complexity of the gradient depends on the structure of the features, namely the number of regions and their labels. Therefore our framework prefers features with small regions and reasonable number of labels $\hat y_r$. Another computational issue relates to the step size $\eta$ for decreasing the objective along the gradient of $w_k$. In general, the coordinate descent scheme verifies that the chosen step size $\eta$ reduces the objective. Theoretically, we can use the fact that the gradient is Lipschitz continuous to predetermine a step size that guarantees descent. However, in practice it gives worse performance than searching for a step size, dividing $\eta$ by a constant factor until descent is guaranteed. 

Lemmas \ref{lemma:lambda} and \ref{lemma:theta} describe the inference and learning steps for minimizing the low dimensional structured prediction program in Theorem \ref{theorem:upper}. Since the program is convex, the order of the minimizing steps is not important, and as long as all inference and learning parameters are optimized the minimal value is attained. For example, one can minimize the inference variables $\lambda$ till they do not change before optimizing the learning parameters $w$. Since these update rules follow the norm-product belief propagation, this approach is equivalent to performing the approximate inference heuristic described in Section \ref{sec:bg-graphs}. Thus our low dimensional structured prediction in Theorem \ref{theorem:upper} provides the objective function for this heuristic. However, this heuristic is computationally intractable as it requires to infer $\lambda$ till convergence for every descent step for learning $w$. Since convexity ensures that the minimization does not depend on the order of the minimizing steps, it also provides a principled way to blend the learning and inference steps. Particularly, it may learn the $w$ parameters using inferred beliefs that do not agree of their marginal probabilities. For this purpose our algorithm infers the parametrized beliefs differently than the (outer) beliefs that are computed by the approximate inference heuristics in Section  \ref{sec:bg-graphs}. This blending property is important in practice, since in the beginning of the algorithm runtime, where the given $w$ are far from the optimum, one needs not spend time on computing consistent beliefs. Figure  \ref{fig:alg} summarizes the inference-learning blending algorithm.

\begin{figure}[t]
\fbox{
\begin{minipage}[c]{17cm}
{\bf Blending learning and inference with low dimensional structured prediction} \\ 
Consider the low dimensional structured prediction program in Theorem \ref{theorem:upper}. 

\begin{enumerate}
\item Repeat until convergence: 
\item For every $(x,y) \in \S$, $r \in {\cal R}$, $\hat y_r \in \Y_r$, $p \in P(r)$:
\item[] Set $\theta_{(x,y),r}(\hat y_r; w) = \ell_r( y_r, \hat y_r) + \sum_{k \in {\cal K}_r} w_k \phi_{k,r}(x,\hat  y_r)$.
\begin{eqnarray*} 
\hspace{-0.8cm} \mu_{(x,y), p \rightarrow r}(\hat y_r) &=&  \epsilon \log \Big(\sum_{\hat y_p \setminus \hat y_r} \exp \big((\theta_{(x,y),p}(\hat y_p; w) +  \!\!\!\! \sum_{c \in C(p) \setminus r} \!\!\!\! \lambda_{(x,y),c \rightarrow p} (\hat y_c) - \!\!\!\! \sum_{p' \in P(p)} \!\!\!\! \lambda_{(x,y),p \rightarrow p'}(\hat y_p)) \big/ \epsilon \big) \Big)\\
\hspace{-0.8cm} \lambda_{(x,y),r \rightarrow p} (\hat y_r) &=&  \frac{1}{1+|P(r)|} \Big( \theta_{(x,y),r}(\hat y_r; w)  + \!\!\!\! \sum_{c \in C(r)} \!\!\!\! \lambda_{(x,y),c \rightarrow r}(\hat y_c) + \!\!\!\! \sum_{p' \in P(r)}  \!\!\!\! \mu_{(x,y),p' \rightarrow r} (\hat y_r) \Big) - \mu_{(x,y),p \rightarrow r} (\hat y_r)
\end{eqnarray*}
\item Set $b_{(x,y),r}(\hat y_r ; w,\lambda, \epsilon) \propto  \exp \big( (\theta_{(x,y),r}(\hat y_r; w) +  \sum_{c \in C(r)} \lambda_{(x,y),c \rightarrow r}(\hat y_c) - \sum_{p \in P(r)} \lambda_{(x,y),r \rightarrow p}(\hat y_r)) \big/ \epsilon \big)$. 
\item[] $w_k \leftarrow w_k - \eta \Big( \sum_{(x,y) \in \S} \sum_{r \in {\cal R}}  \Big(\sum_{\hat y_r} b_{(x,y),r}(\hat y_r ; w, \lambda, \epsilon) \phi_{k,r}(x,\hat y_r) - \phi_{k,r}(x,y_r) \Big) + C w_k \Big)$. 
\end{enumerate}
\end{minipage}
}
\caption{The inference step is described in Lemma \ref{lemma:lambda} and the learning step is described in Lemma \ref{lemma:theta}. The step size $\eta$ is set to guarantee convergence (e.g., corresponding to the Lipschitz constant or the Armijo rule.) Convexity of the program in Theorem \ref{theorem:upper} ensures that the blending  converges to consistent inferred beliefs, see Theorem \ref{theorem:conv} and Section \ref{sec:duality}. }
\label{fig:alg}
\end{figure}

The block coordinate descent algorithm is guaranteed to converge, as it monotonically decreases the objective in Theorem \ref{theorem:upper}, which is lower bounded by its dual. However, convergence to the global minimum cannot be guaranteed in all cases. In particular, for $\epsilon=0$ coordinate descent on the low dimensional structured SVM program is not guaranteed to converge to its global minimum. To converge to the global minimum in this case one can use subgradient methods, but despite their theoretical guarantees they turn to be slow in practice. Since the primal program is not strictly convex in $\lambda$, even when we are guaranteed to converge to the global minimum, when $\epsilon > 0$, the sequence of variables $\lambda_{(x,y),r \rightarrow p}(\hat y_r)$ generated by the algorithm is not guaranteed to be bounded. As a trivial example, adding an arbitrary constant to the variables, $\lambda_{(x,y),r \rightarrow p}(\hat y_r) + c$, does not change the objective value, hence the algorithm can generate monotonically decreasing unbounded sequences. However, the beliefs generated by the algorithm are bounded and guaranteed to converge to the unique solution of the dual program. The convergence properties of the algorithm are summarized in the following claim. 
 
\begin{theorem}
\label{theorem:conv}
The learning-inference blending algorithm in Figure \ref{fig:alg} for low dimensional structured prediction is guaranteed to converge. Moreover, if $\epsilon > 0$, then the value of its objective is guaranteed to converge to the global minimum, and its sequence of beliefs are guaranteed to converge to the unique solution of the dual program.   
\end{theorem}
{\bf Proof:} 
The block coordinate descent algorithm in Figure \ref{fig:alg} iteratively apply Lemmas \ref{lemma:lambda} and \ref{lemma:theta} thus monotonically decreases the low dimensional structured prediction program in Theorem \ref{theorem:upper}. This program is bounded by its dual program (see Section \ref{sec:duality}), therefore the value of its objective is guaranteed to converge. 

Whenever $\epsilon > 0$, the dual objective (see Section \ref{sec:duality}) is strictly concave in $b_{(x,y),r}(\hat y_r) , z_k$ subject to linear marginalization constraints and the linear moment constraints  $$z_k =  \sum_{(x,y) \in \S} ( \sum_{r \in {\cal R}_k} \sum_{\hat y_r} b_{(x,y),r}(\hat y_r) \phi_{k,r}(x,\hat y_r) - \phi_k (x,y) ).$$ Hence the claim properties are a direct consequence of \cite{Tseng87}. \eop \\

The convergence of the block coordinate descent depends on the step size $\eta$, which requires to reduce the objective. This can be done by the Armijo rule, or by using the fact that the function $z^2$ is strongly convex (e.g., \cite{Tseng87}) and its gradient is Lipschitz continuous (e.g., \cite{Nesterov04}). In practice, Theorem \ref{theorem:conv} describes how to measure the convergence of the algorithm.

\section{Experimental evaluation}
\label{sec:experiments}

\begin{figure}
\begin{small}
\begin{center}
\begin{tabular}{|c||c|c|c|c||c|c|c|c|}
	\hline
&	\multicolumn{4}{|c||}{{\bf Gaussian noise}} &	\multicolumn{4}{|c|}{{\bf Bimodal noise}}\\
	\hline 
 &  $I_1$    &   $I_2$  & $I_3$ & $I_4$   &  $I_1$    &   $I_2$  & $I_3$ & $I_4$  \\
	\hline \hline
LBP-SGD  &  2.7344& 2.4707 & 3.2275 & 2.3193  & 5.2905 & 4.4751 & 6.8164 & 7.2510\\ \hline
LBP-SMD & 2.7344 & 2.4731 & 3.2324 & 2.3145 & 5.2954  & 4.4678 & 6.7578 & 7.2583\\ \hline
LBP-BFGS  & 2.7417  & 2.4194  & 3.1299 & 2.4023 & 5.2148 & 4.3994 & 6.0278 & 6.6211 \\ \hline
MF-SGD & 3.0469  & 3.0762  & 4,1382 & 2.9053 & 10.0488 & 41.0718 & 29.6338 &  53.6035\\ \hline
MF-SMD & 2.9688 & 3.0640 &3.8721 & 14.4360 & -- & -- & -- & --\\ \hline
MF-BFGS & 3.0005 & 2.7783 & 3.6157 & 2.4780 & 5.2661 & 4.6167 & 6.4624& 7.2510\\ \hline
Ours   & {\bf  0.0488} & {\bf 0.0073} & {\bf 0.1294} & {\bf 0.1318} & {\bf 0.0537} & {\bf 0.0244}  & {\bf 0.1221} & {\bf  0.9277} \\ \hline
\end{tabular}
\end{center}
\vspace{-0.4cm}
\caption{{\bf Gaussian and bimodal noise}: Comparison of our approach to loopy belief propagation and mean field approximations when optimizing using BFGS, SGD and SMD. Note that our approach significantly outperforms all the baselines. MF-SMD did not work for Bimodal noise.}
\label{fig:denoise_table}
\end{small}
\end{figure}

In this section we evaluate our approach in a wide range of computer vision problems including de-noising, stereo estimation, semantic segmentation, shape reconstruction and 3D indoor scene understanding. Our approach enables us to learn  a large number of parameters efficiently and results in state-of-the-art performance in all these tasks. 
A more detailed version of these results can be found in \cite{Salzmann12,Yamaguchi12,Schwing12-cvpr,Yao12}.

\subsection{Image Denoising}

We performed experiments on 2D grids since they are widely used to represent images, and have many  cycles.  We first investigate the role of $\epsilon$ in the accuracy and running time of our algorithm, described in Fig,. \ref{fig:alg}. We used a $10 \times 10$ binary image and randomly generated $10$ corrupted samples flipping every bit with $0.2$ probability. We trained the model using $\epsilon=\{1, 0.5, 0.01, 0\}$, ranging the low-dimensional extended log-loss from $\epsilon=1$ (low dimensional CRFs) to $\epsilon=0$ (low dimensional  structured SVM) and its smooth version ($\epsilon=0.01$). The runtimes are $323,324,326,294$ seconds for $\epsilon = \{1, 0.5, 0.01, 0\}$ respectively. As $\epsilon$ gets smaller the runtime slightly increases, but it decreases for $\epsilon=0$ since the $\ell_\infty$ norm is  efficiently computed using the max function. However, for $\epsilon=0$ it is hard to determine the optimality as the max-function is non-smooth thus a dual solution is not uniquely recovered. When the approximated structured SVM converges, the gap between the primal objective and dual objective was $1.3$, while the dual beliefs were recovered according to the subgradient, i.e., the maximal arguments. In contrast, for $\epsilon>0$ the primal-dual gap was $10^{-5}$, while the the dual beliefs were uniquely recovered using the gradient.

We generated test images in a similar fashion. When using the same $\epsilon$ for training and testing we obtained $2$ misclassifications for $\epsilon>0$  and $109$ for $\epsilon=0$. We conjecture that this comes from the existence of multiple maximal arguments in the primal objective when $\epsilon=0$, or equivalently from its non-smooth corners.
We also evaluated the quality of the solution using different values of $\epsilon$  for training and inference, following \cite{Wainwright06}.
When predicting with smaller $\epsilon$ than the one used for learning the results are marginally worse than when predicting with the same $\epsilon$. However, when predicting with larger $\epsilon$, the results get significantly worse, e.g., learning  with $\epsilon=0.01$ and predicting with $\epsilon=1$ results in $10$ errors, and only $2$ when  $\epsilon=0.01$. 

The main advantage of our algorithm is that it can efficiently learn many parameters in a graphical model. We now compared, in a similarly generated dataset of size $5 \times 5$,  a model learned with different parameters for every edge and vertex ($\approx 300$ parameters) and a model learned with  parameters shared among the vertices and edges (2 parameters for edges and 2 for vertices) used by \cite{Kumar04}. Using large number of parameters  increases performance:  sharing parameters resulted in $16$ misclassifications, while optimizing over the  $300$ parameters resulted in $2$ errors.  WE note that our algorithm avoids overfitting in this case.

We now compare our algorithm in Fig. \ref{fig:alg} to standard CRF solvers that use different approaches to compute the gradient. 
We use the binary image dataset of \cite{Kumar04} that consists of 4 different $64 \times 64$ base images. Each base image was corrupted 50 times with each type of noise.  
Following \cite{Murphy06}, we trained different models to denoise each individual image, using 40 examples for training and 10 for test.
We compare our approach to the result of approximating the conditional likelihood using loopy belief propagation (LBP) and mean field approximation (MF). For each of these approximations, we use stochastic gradient descent (SGD), stochastic meta-descent (SMD) and BFGS to learn the parameters. We do not report pseudolikelihood  (PL) results since it did not work. Note that the same behavior of PL was noticed by \cite{Murphy06}. 
To reduce the computational complexity and the chances of convergence, \cite{Kumar04,Murphy06} forced their parameters to be shared across all nodes such that $\forall i, \theta_i=\theta^{(n)}$ and $\forall i, \forall j \in N(i), ~ \theta_{ij} = \theta^{{e}}$. 
In contrast, we can exploit the full flexibility of the graph and learn more than $10,000$ parameters. Note that this is computationally prohibitive with the baselines.
For the local features we simply use the pixel values, and for the node potentials we use an Ising model with only bias features such that $\phi_{i,j} = [1, -1; -1, 1]$. For all experiments we use $\epsilon = 1$.
For the baselines, we use the code, features and optimal parameters of \cite{Murphy06}.


\begin{figure}[t]
\begin{center}
\begin{tabular}{|cccc|cccc|}
\hline
\includegraphics[width=0.12\columnwidth]{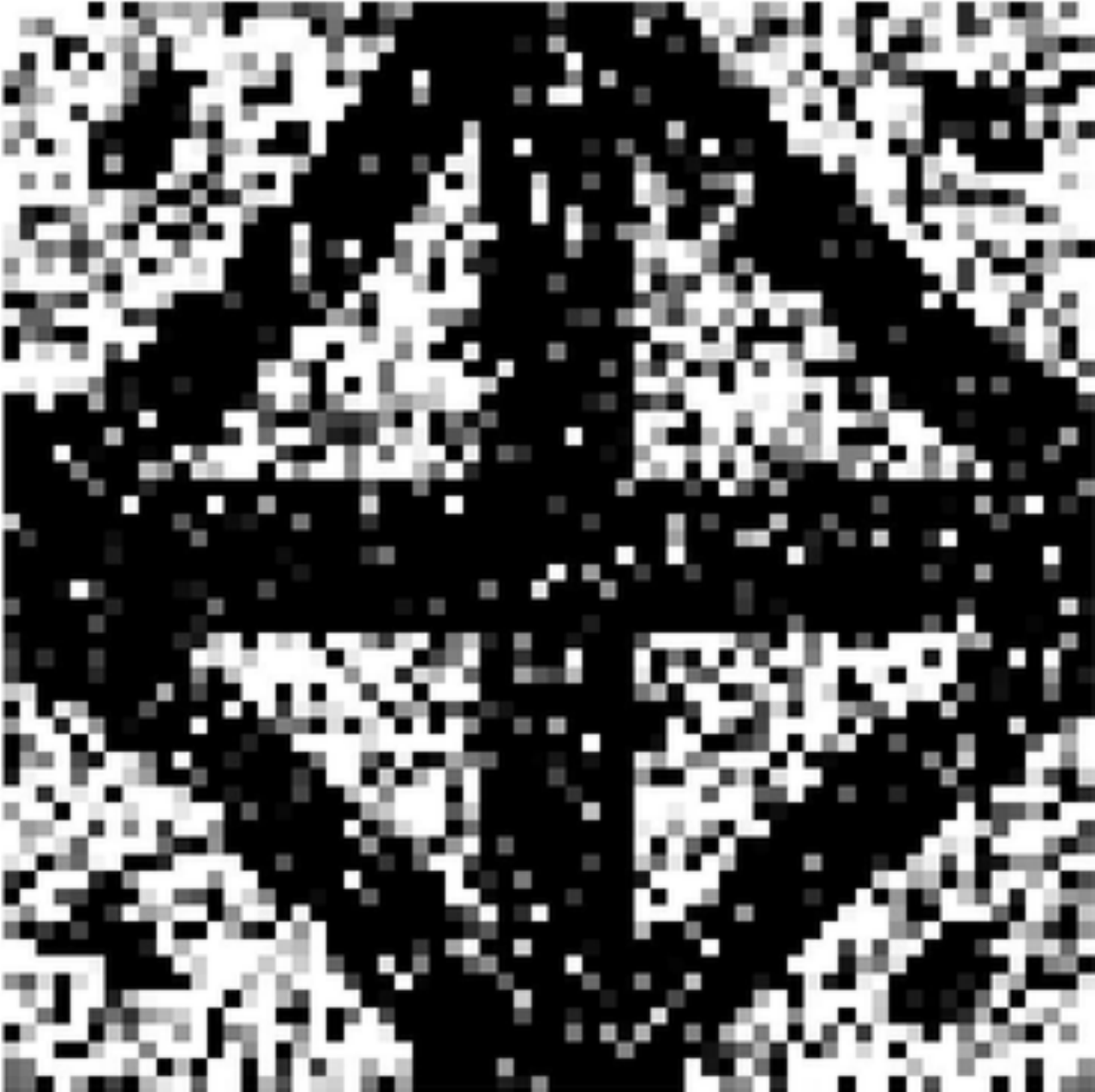} & \hspace{-0.5cm}
\includegraphics[width=0.12\columnwidth]{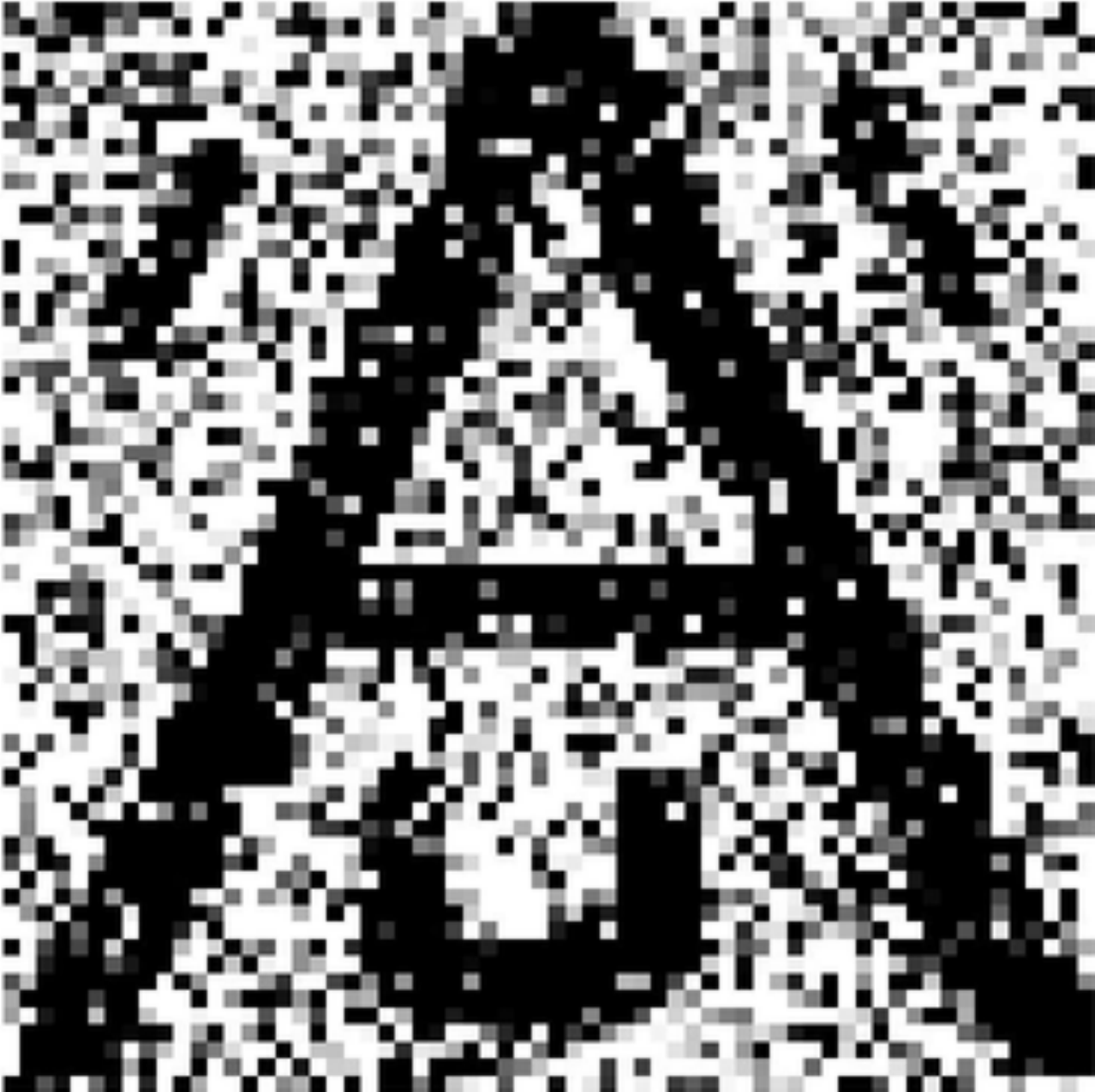} &\hspace{-0.5cm}
\includegraphics[width=0.12\columnwidth]{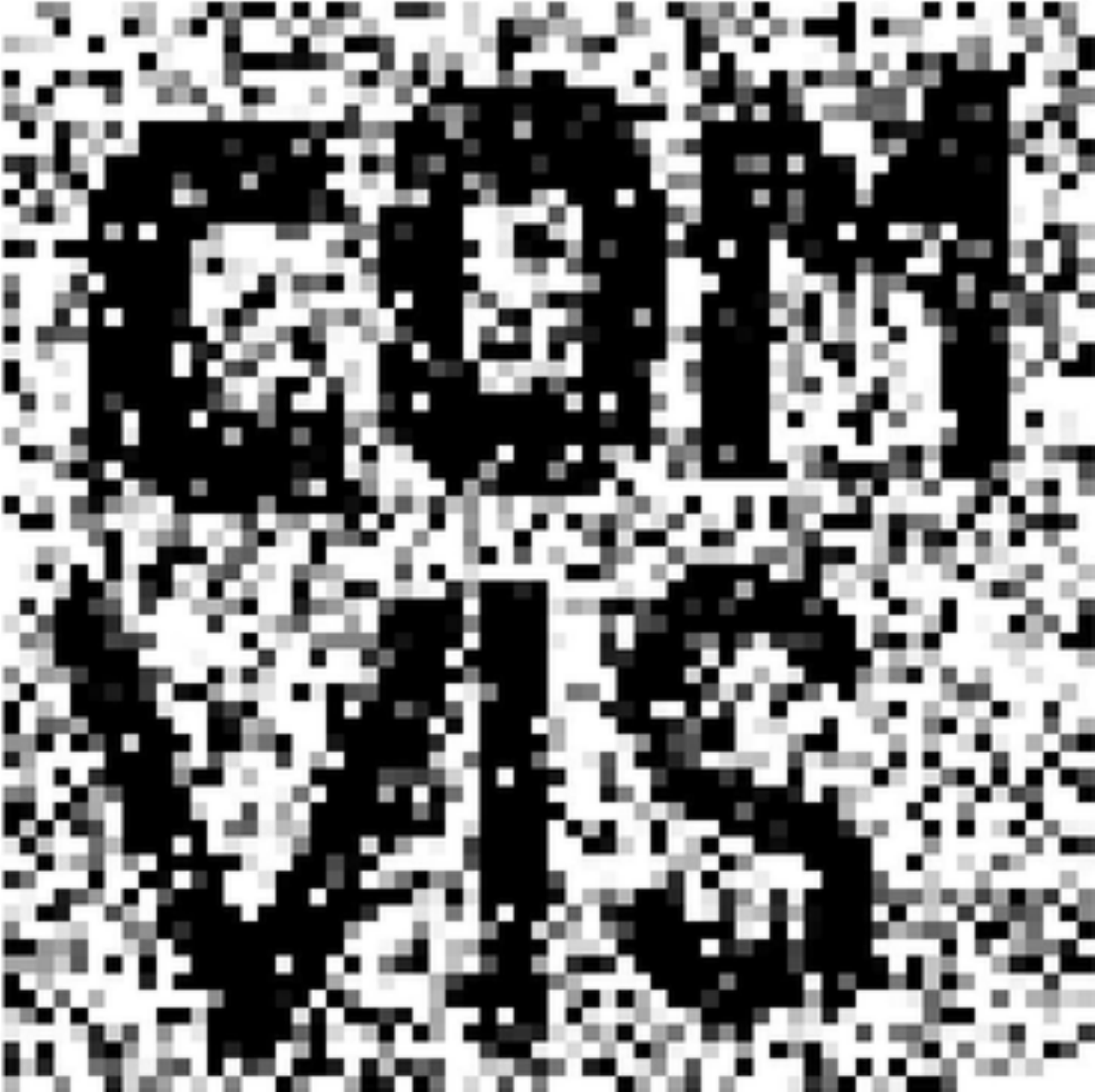} &\hspace{-0.5cm}
\includegraphics[width= 0.12\columnwidth]{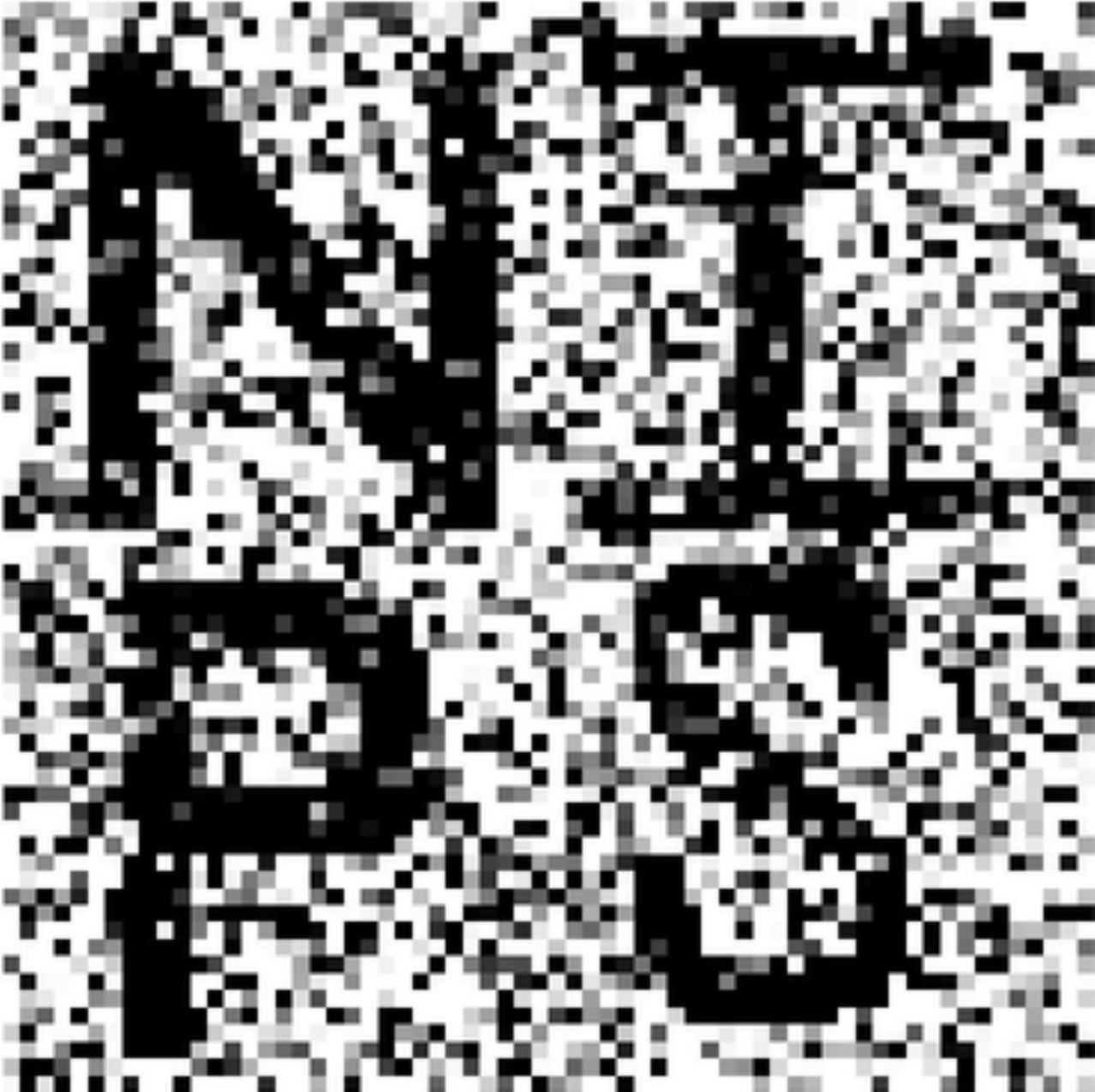} &\hspace{-0.2cm}
\includegraphics[width=0.12\columnwidth]{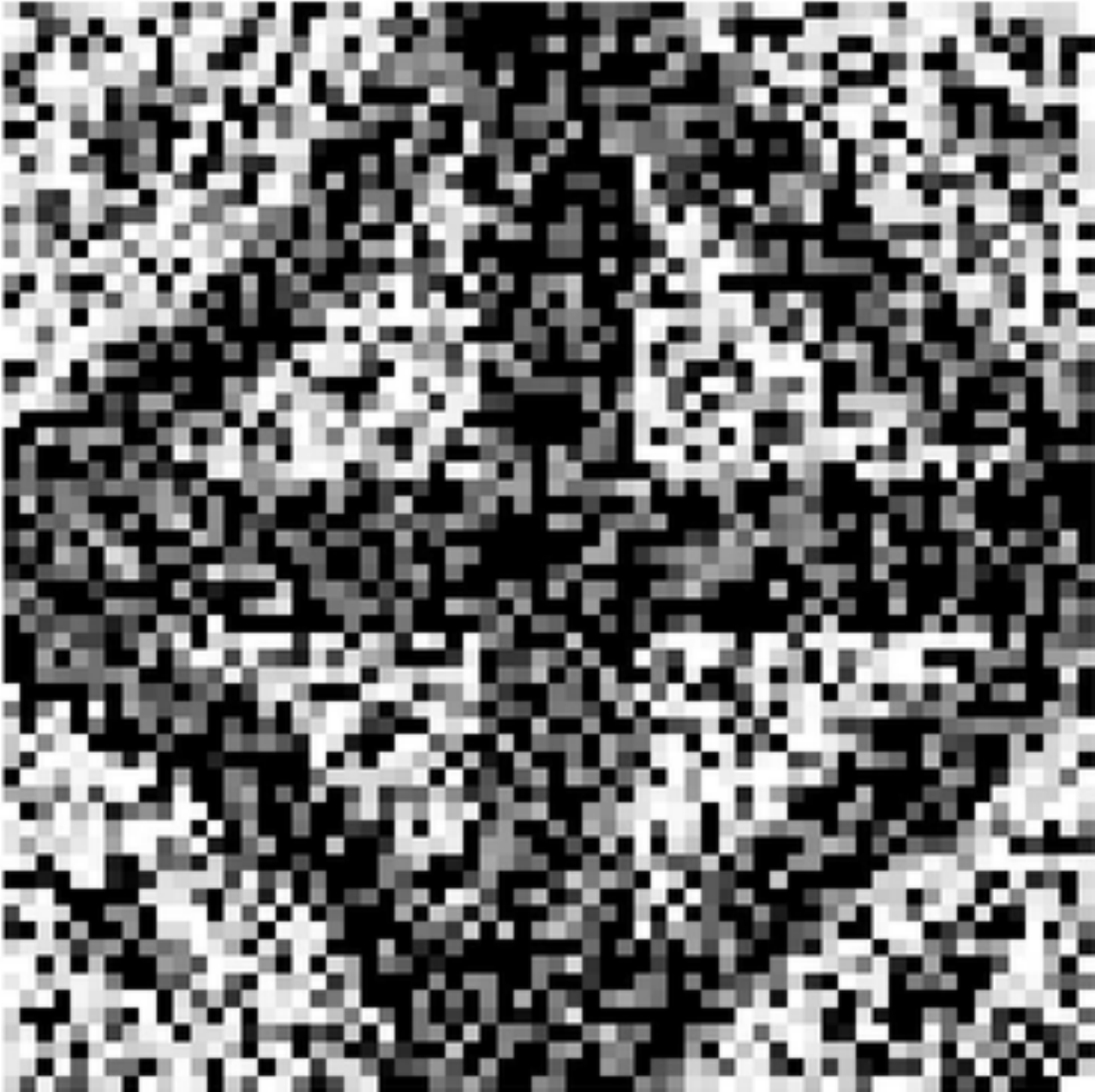} &\hspace{-0.5cm}
\includegraphics[width=0.12\columnwidth]{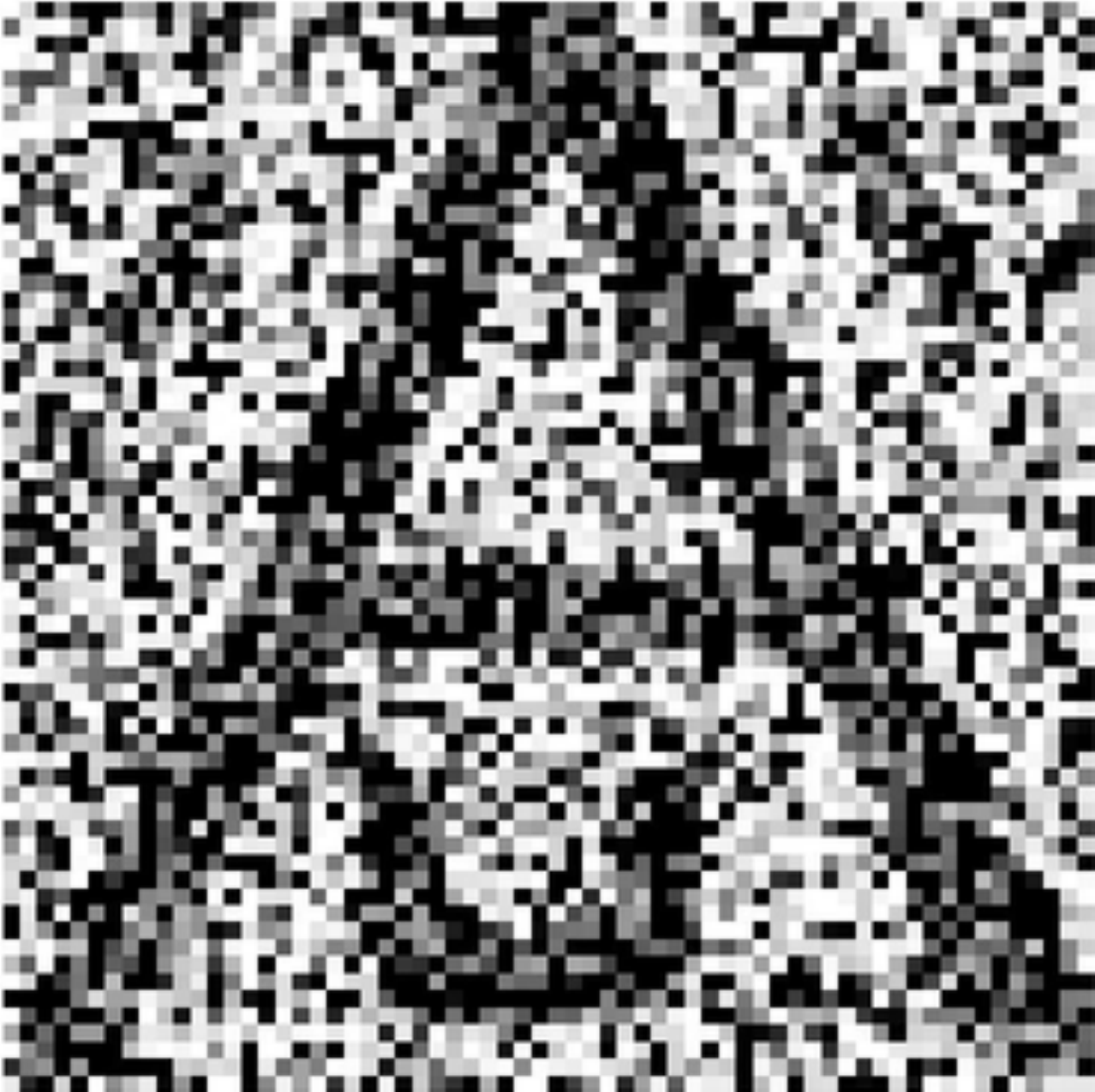} &\hspace{-0.5cm}
\includegraphics[width=0.12\columnwidth]{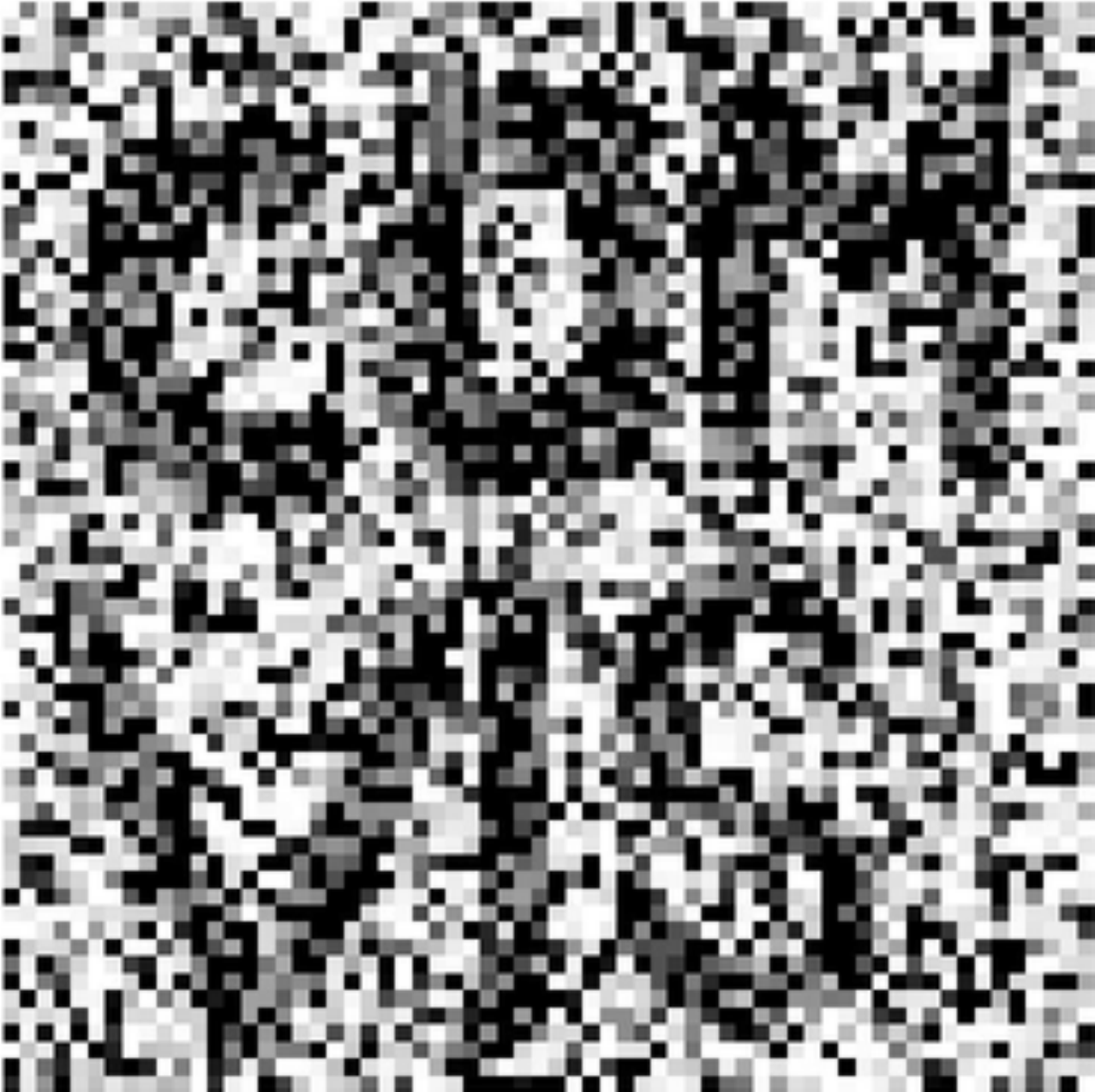} &\hspace{-0.5cm}
\includegraphics[width=0.12\columnwidth]{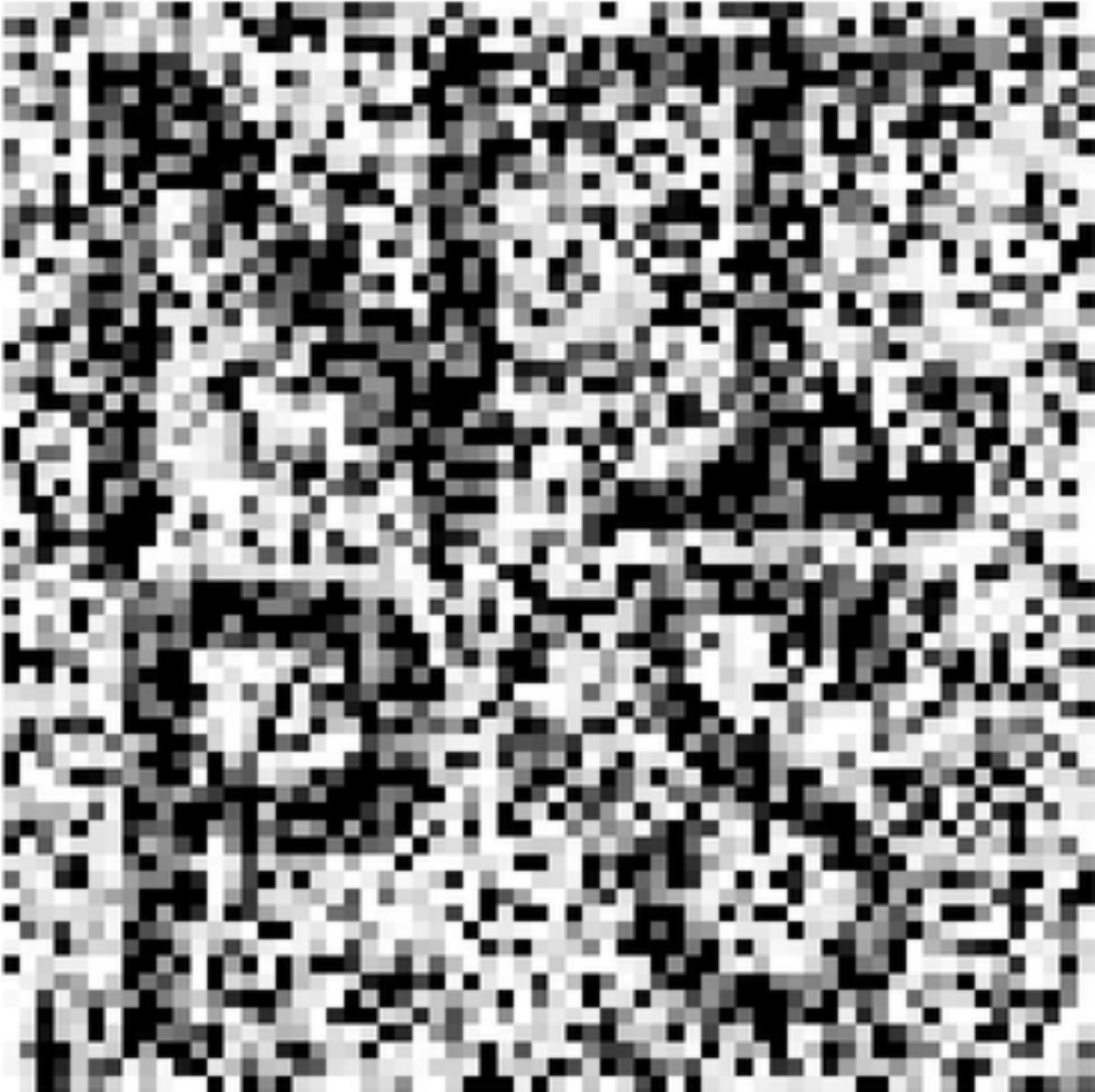}\hspace{-0.2cm}
\\
\includegraphics[width= 0.12\columnwidth]{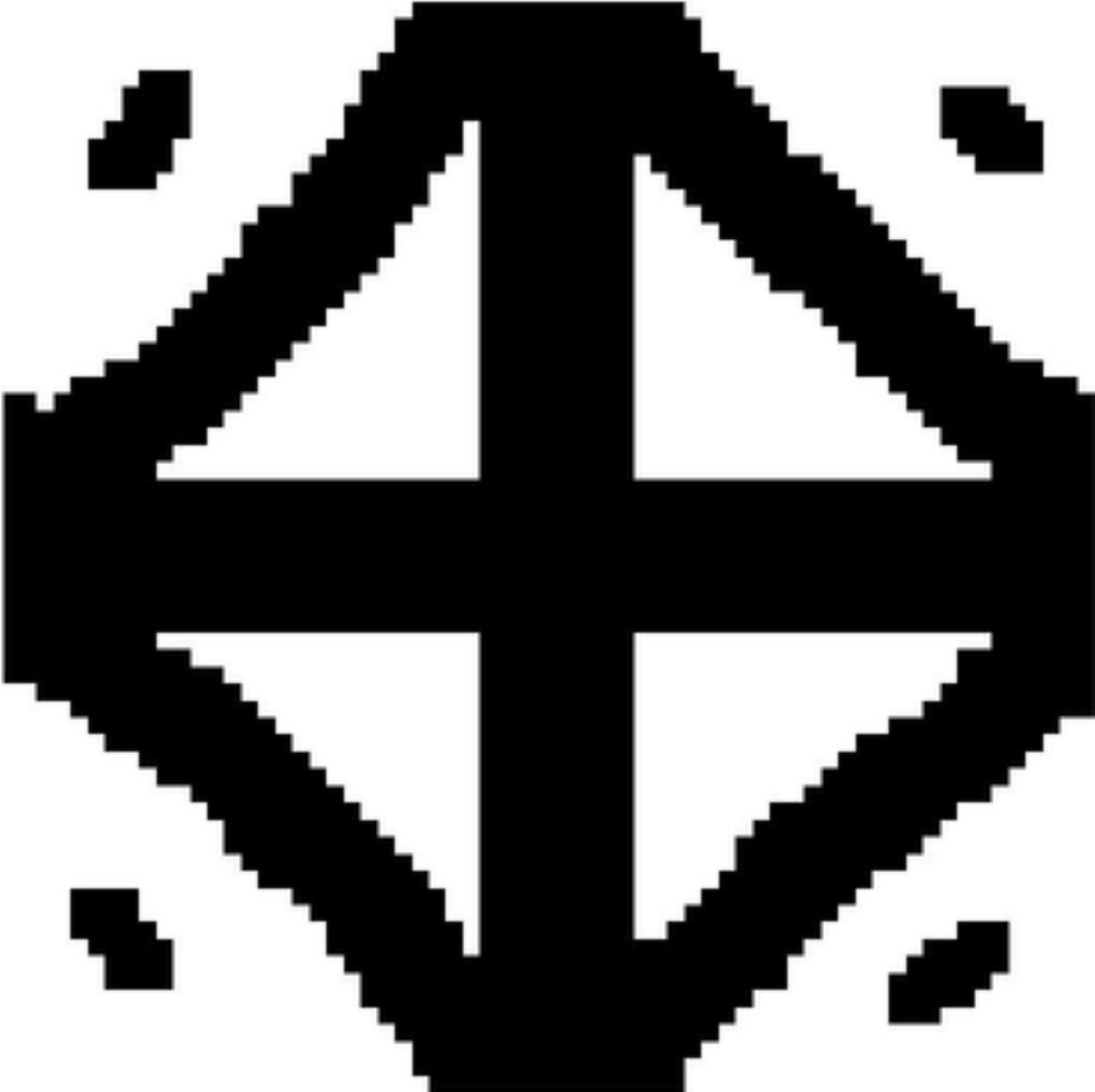} &\hspace{-0.5cm}
\includegraphics[width= 0.12\columnwidth]{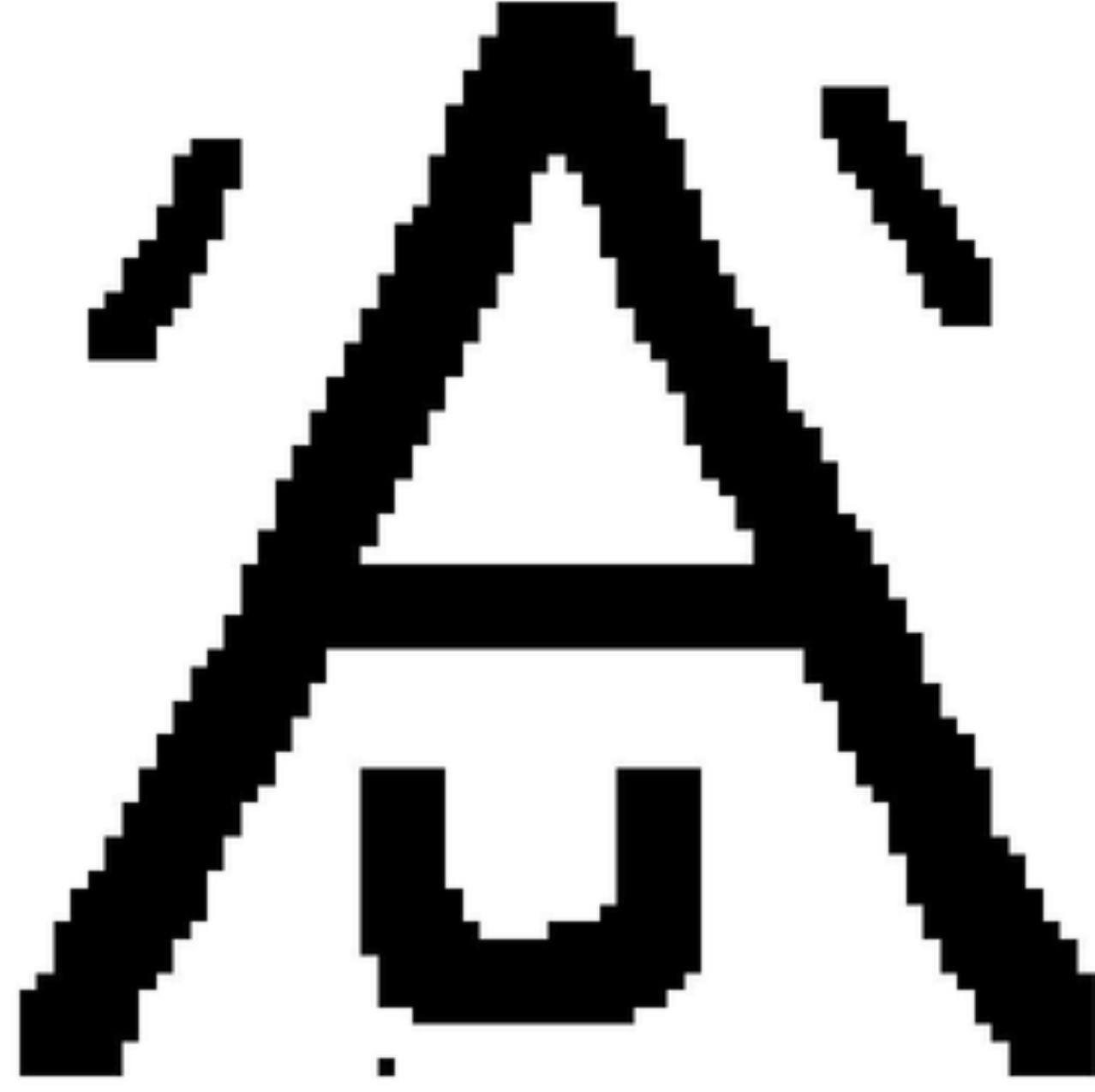} &\hspace{-0.5cm}
\includegraphics[width= 0.12\columnwidth]{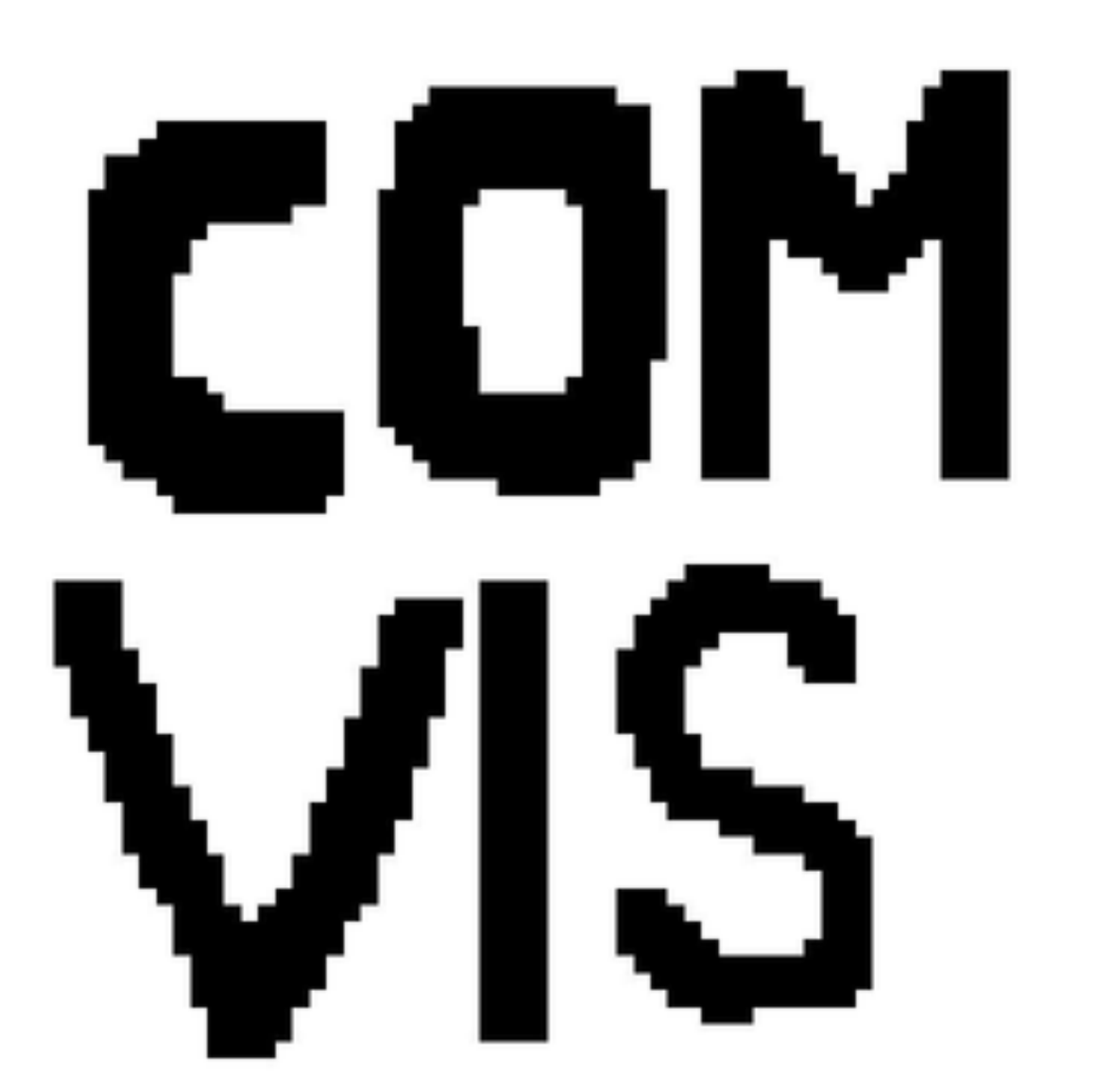} &\hspace{-0.5cm}
\includegraphics[width= 0.12\columnwidth]{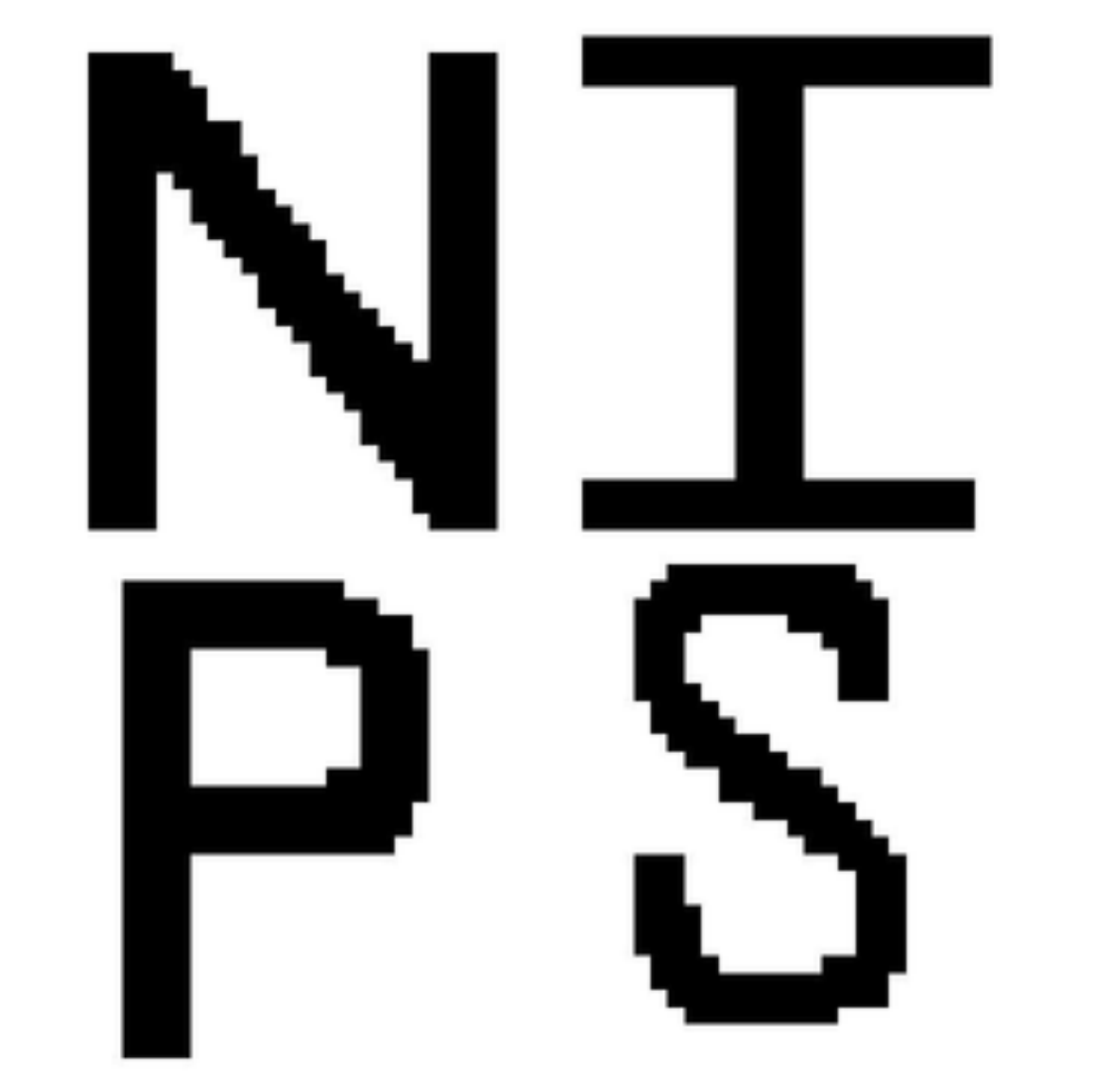} &\hspace{-0.2cm}
\includegraphics[width=0.12\columnwidth]{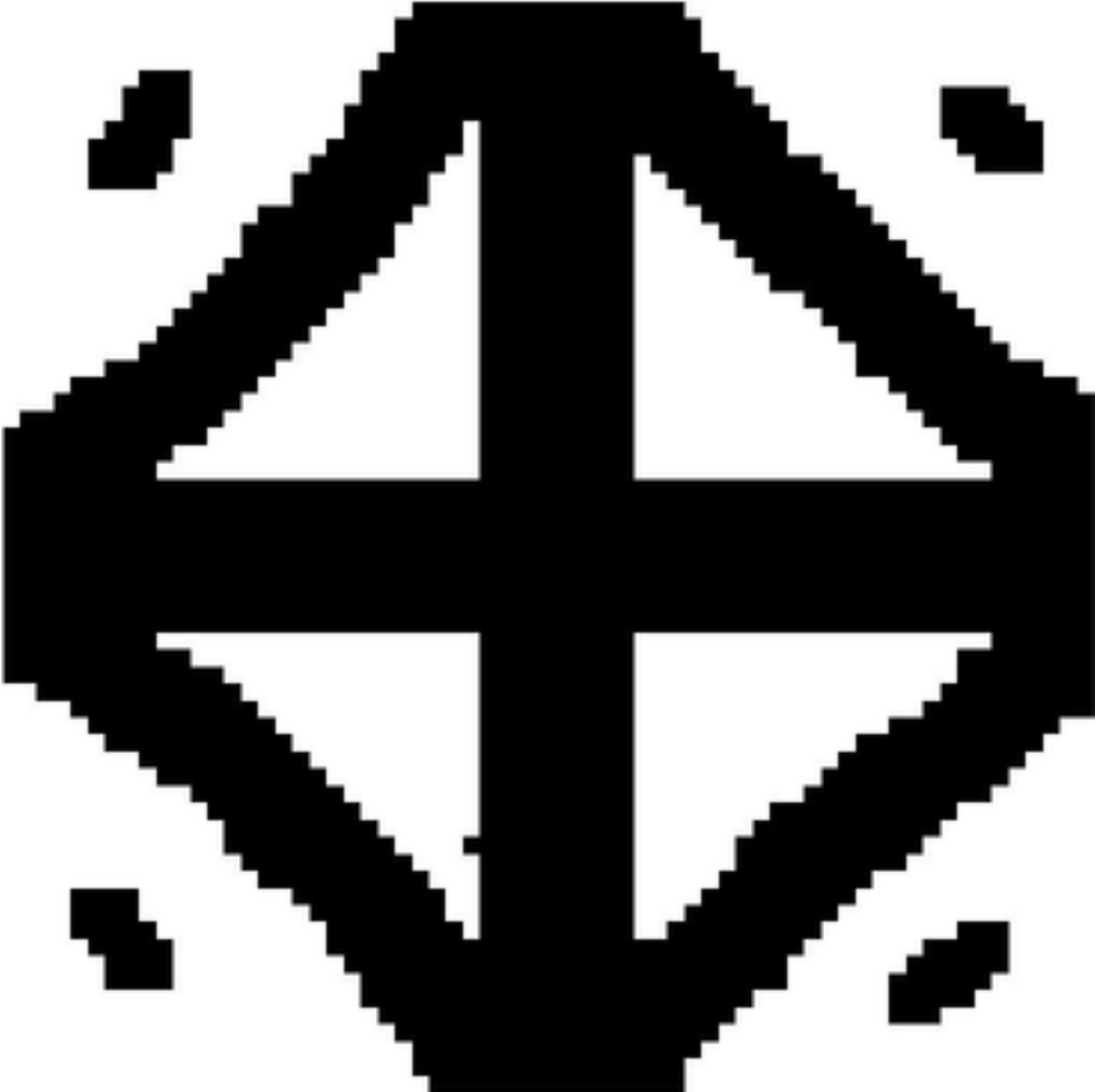} &\hspace{-0.5cm}
\includegraphics[width=0.12\columnwidth]{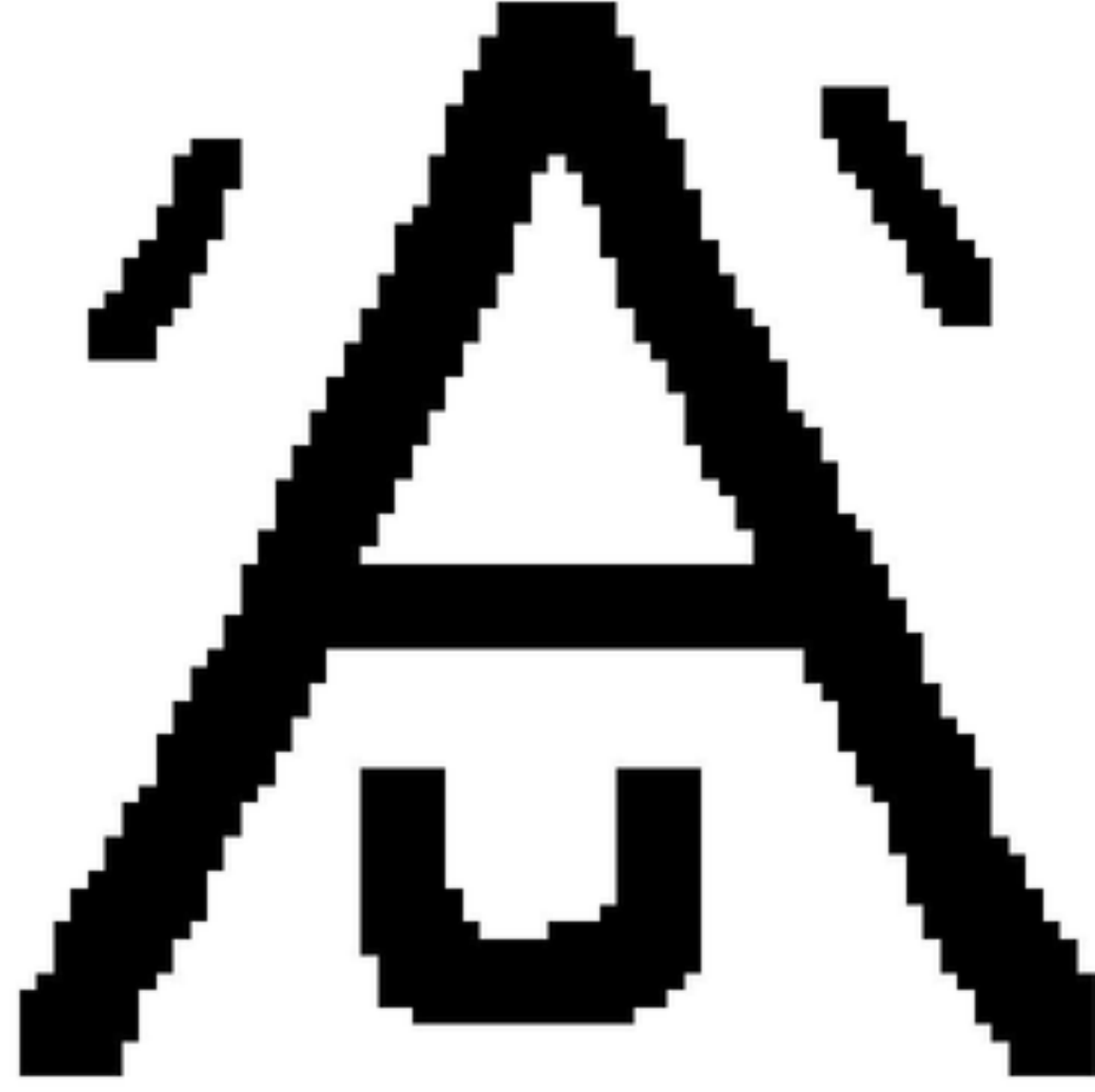} &\hspace{-0.5cm}
\includegraphics[width=0.12\columnwidth]{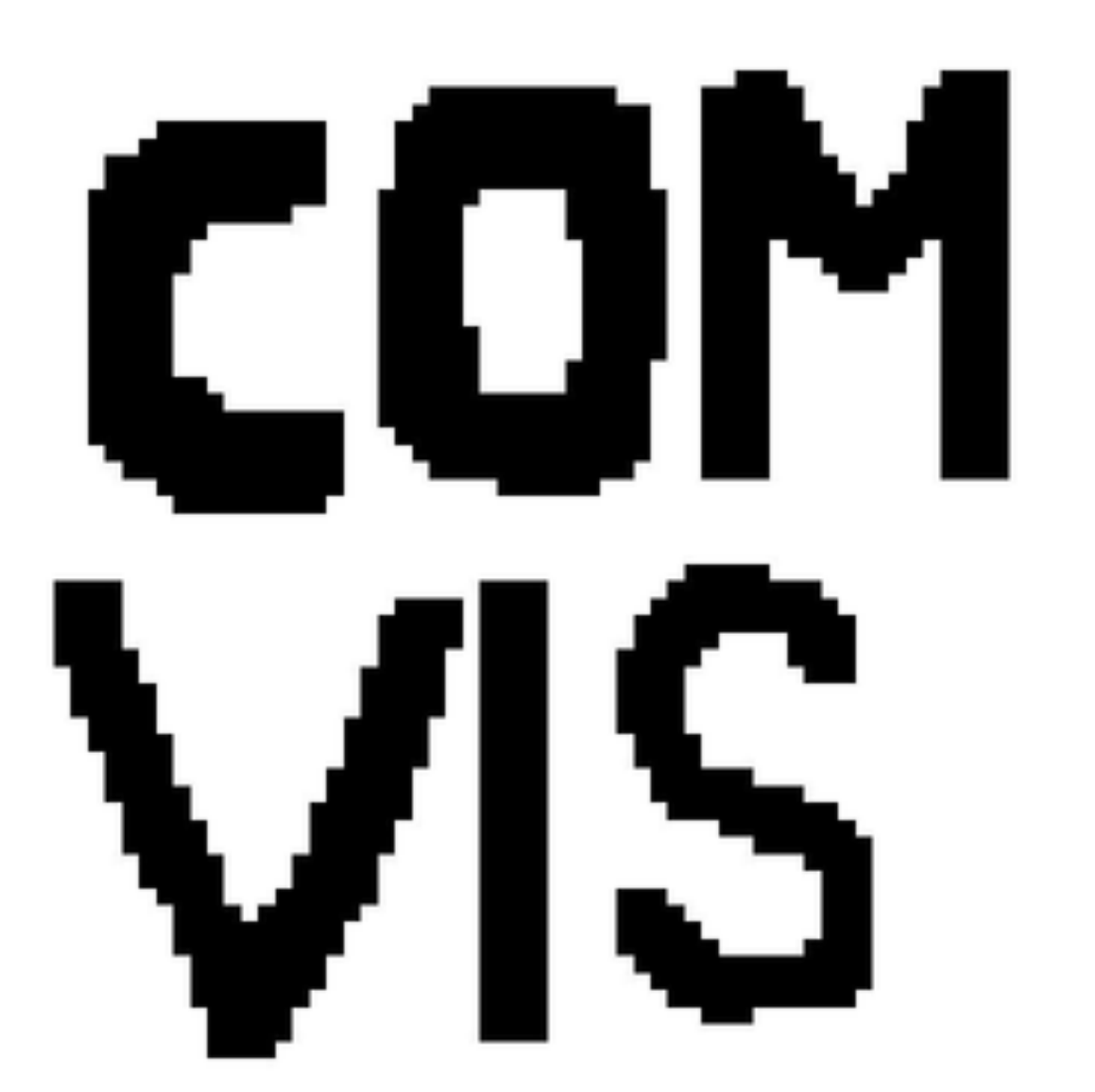} &\hspace{-0.5cm}
\includegraphics[width=0.12\columnwidth]{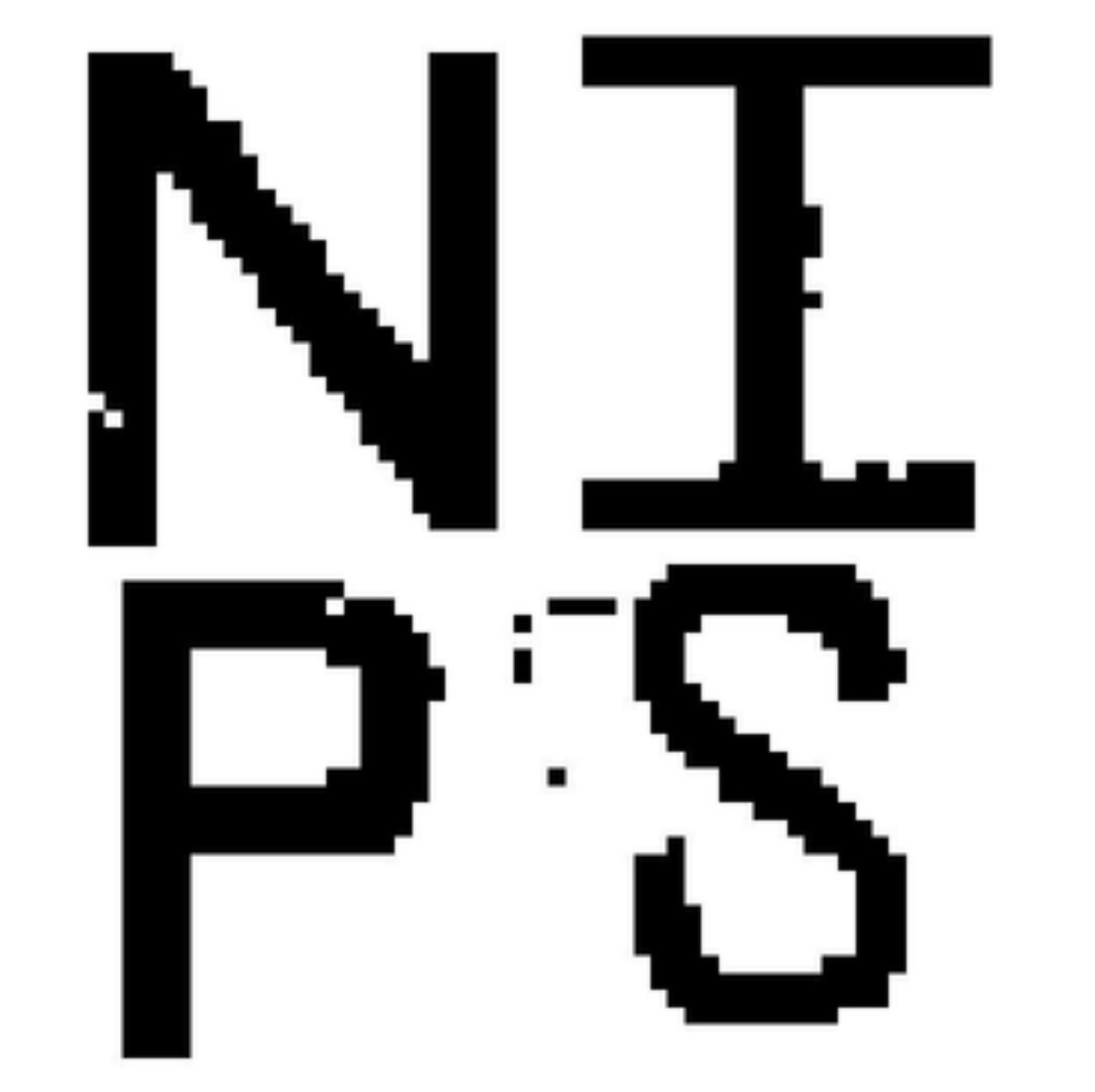}\hspace{-0.2cm}
\\
\hline
\end{tabular}
\end{center}
\caption{{\bf Denoising results}: Gaussian (left) and Bimodal (right) noise.} 
\label{fig:denoise}
\end{figure}

Under the first noise model, each pixel was corrupted via i.i.d. Gaussian noise with mean 0 and standard deviation of 0.3. 
Fig. \ref{fig:denoise_table}  depicts test error in $(\%)$ for the different base images (i.e., $I_1,\dots, I_4$). Note that our approach outperforms considerably the loopy belief propagation and mean field approximations for all optimization criteria (BFGS, SGD, SMD).  For example, for the first base image the error of our approach is $0.0488 \%$, which is equivalent to a 2 pixels error on average. In contrast the best baseline gets 112 pixels wrong on average.
Fig. \ref{fig:denoise} (left) depicts test examples as well as our denoising results. Note that our approach is able to cope with large amounts of noise.

\begin{figure}[t]
\begin{small}
\begin{center}
\begin{tabular}{cc}
\includegraphics[width=0.4\columnwidth]{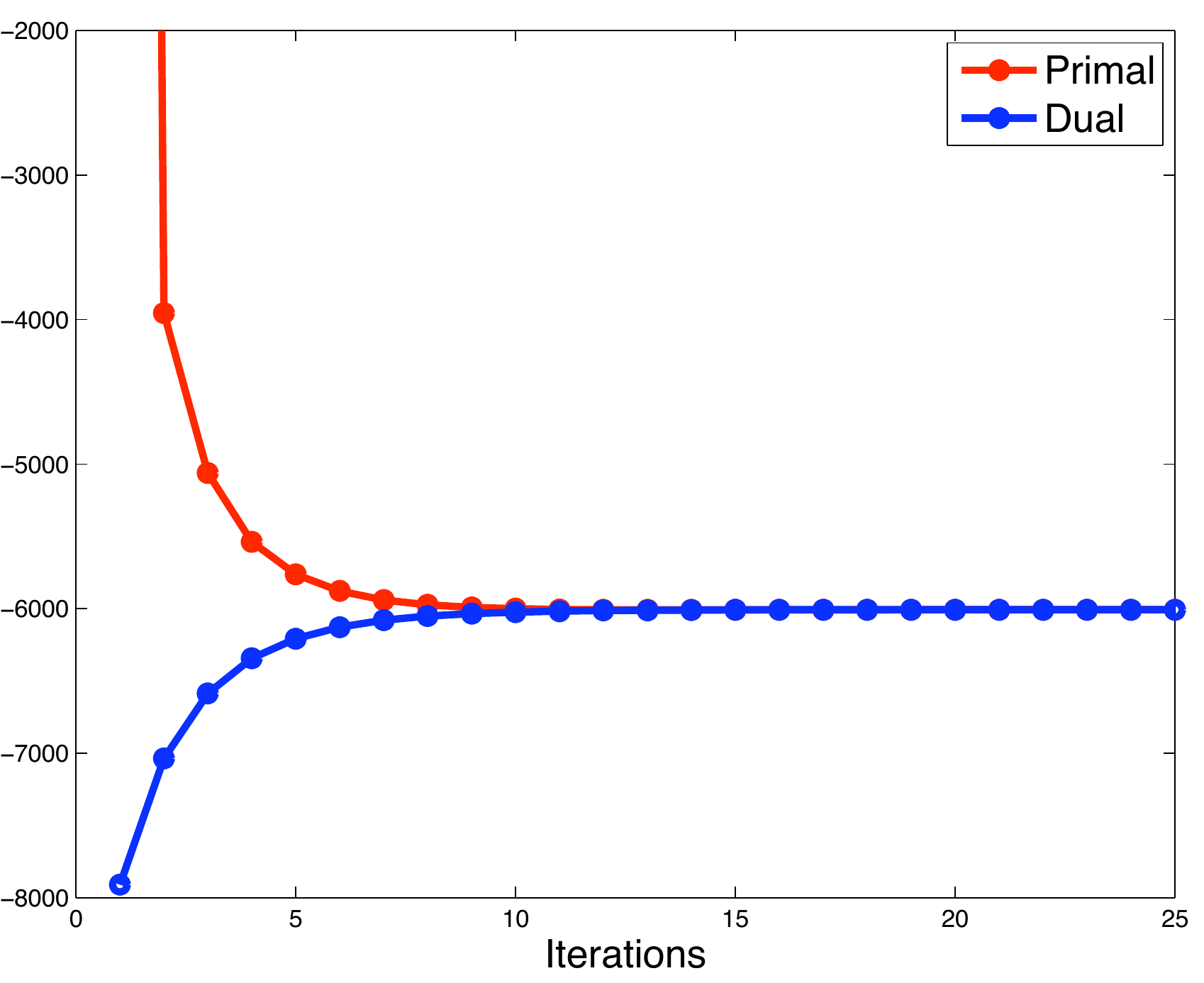} & \hspace{0.5cm}
\includegraphics[width=0.4\columnwidth]{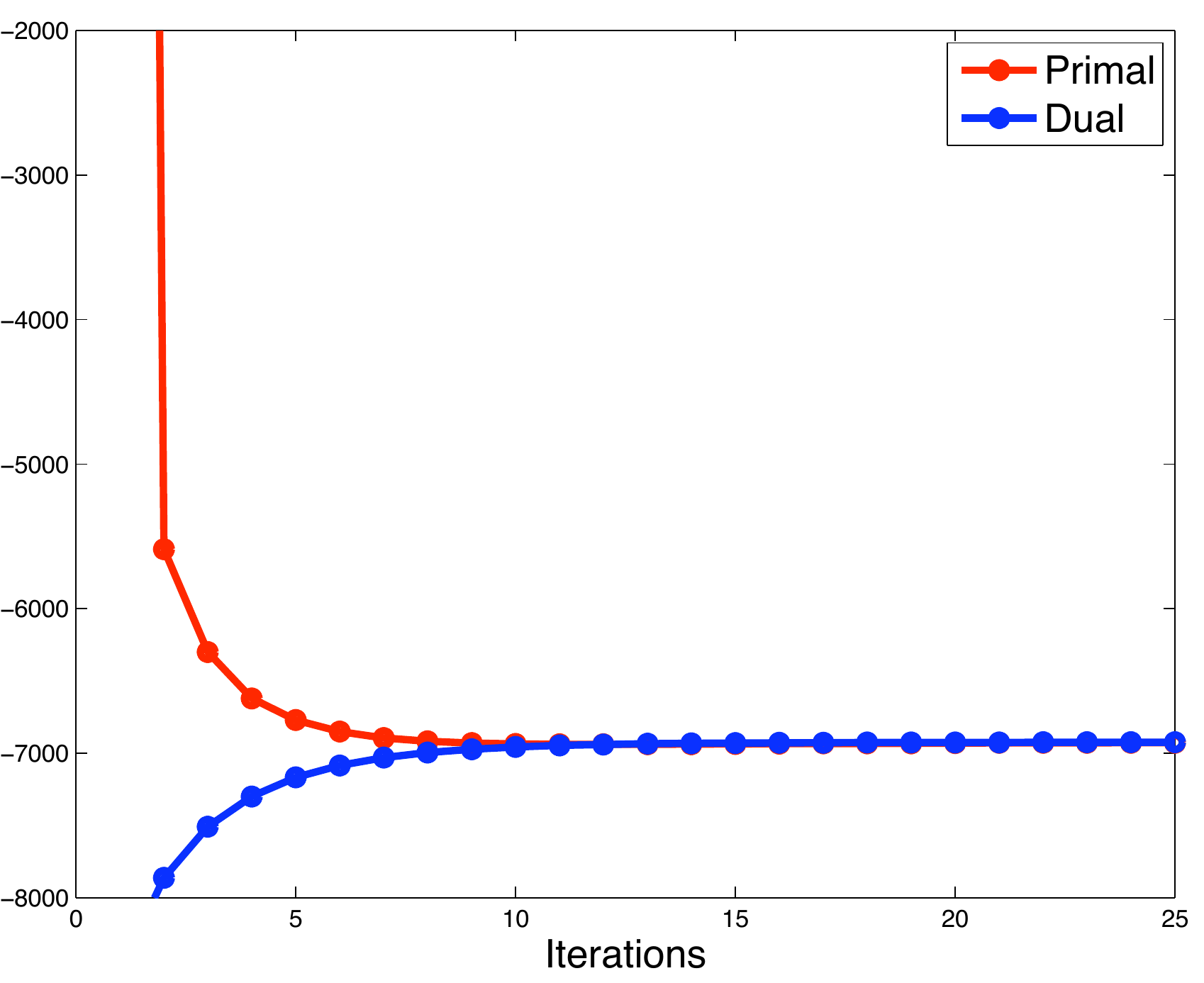} 
\\
(Gaussian) &
(Bimodal)
\end{tabular}
\end{center}
\caption{{\bf Convergence}. Primal and dual train errors when for $I_1$ is corrupted with Gaussian and Bimodal noise. Our algorithm is able to converge in a few iterations.}
\label{fig:primaldual}
\end{small}
\end{figure}

Under the second noise model, each pixel was corrupted with an independent mixture of Gaussians. For each class, a mixture of 2 Gaussians with equal mixing weights was used, yielding the Bimodal noise. The mixture model parameters were $(0.08,0.03)$ and $(0.46,0.03)$ for the first class and $(0.55,0.02)$ and $(0.42,0.10)$ for the second class, with $(a,b)$ a Gaussian with mean $a$ and standard deviation $b$.
Fig. \ref{fig:denoise_table}  depicts test error in $(\%)$ for the different base images. As before, our approach outperforms all the baselines. We do not report MF-SMD results since it did not work.
Denoised images  are shown in Fig. \ref{fig:denoise} (right).
We now  show how our algorithm converges in a few iterations. Fig. \ref{fig:primaldual} depicts the primal and dual training errors as a function of the number of iterations. Note that our algorithm converges, and the dual and primal values are very tight after a few iterations.


\subsection{Stereo estimation}

The problem of stereo estimation consists of two images of a scene, for which we wish to calculate the depth for each pixel in these images.  Assuming that the cameras are calibrated and the images are rectified, the problem can be reduced for each pixel to a 1-D search  along the corresponding epipolar line. 
Over the past few decades we have witnessed a great improvement in
performance of stereo algorithms. Most modern approaches frame the
problem as  inference on a graphical model. 
Most methods \cite{BirchfieldTomasi,Hong04,Bleyer05,KSK06,Deng05,Yang08,Trinh09}
assume a fixed set of superpixels on a reference image, say the left image, 
and model the surface under each superpixel as a slanted plane.
The graphical model typically has a robust data term scoring the assigned plane in terms of
a matching score induced by the plane on the pixels contained in the superpixel.
This data term often incorporates an explicit treatment of occlusion --- pixels in one image
that have no corresponding pixel in the other image \cite{Kanade00,Kolmogorov02,Deng05,Bleyer10}.
Slanted-plane models also typically include a robust smoothness term expressing the belief that the planes assigned to
adjacent superpixels should be similar.
Despite recent advances in learning graphical models, most approaches hand tuned their parameters.

In recent work \cite{Yamaguchi}, we have proposed an approach to stereo estimation that is computational efficient in both learning and inference. We incorporate a better model of occlusion  than existing approaches by modeling explicitly occlusion boundaries between adjacent superpixels. This allow us to incorporate potentials that reason about boundary ownership as well as whether junctions are physically possible. We now briefly discuss the graphical model as well as the potentials we employ. 

We represent the stereo estimation problem as the one of inference in a hybrid Markov random field that contains a mixture of discrete and continuous random variables. 
The continuous random variables represent, for each segment, the disparities of all pixels contained in that segment in the form of a  3D slanted plane. The discrete random variables indicate for each pair of neighboring segments, whether they are co-planar, they form a hinge or there is a depth discontinuity (indicating which plane is in front of which). 
Let $o_{i,j}  \in \{co, hi, lo, ro \}$ be a discrete random variable representing whether two neighboring planes are  coplanar, form a hinge or an occlusion boundary.  Here, $lo$ implies that plane $i$ occludes plane $j$, and $ro$ represents that plane $j$ occludes plane $i$. 
We define our  hybrid conditional random field  as follows 
\begin{equation}
p(\by,\bo) = \frac{1}{Z} 
\prod_{\vartheta}\psi_{\vartheta}(\by_{\vartheta})\prod_\zeta\psi_{\zeta}(\bo_{\zeta})\prod_{\tau} \psi_\tau(\by_\tau, \bo_\tau)
\vspace{-0.2cm}
\label{eq:mrf}
\end{equation}
where $\by$ represents the set of all 3D slanted planes, $\bo$ the set of all discrete random variables, and  $\psi_\vartheta, \psi_\zeta, \psi_\tau$ encode potential functions over sets of continuos, discrete or mixture of both types of  variables.  Note that $\by$ contains three random variables for every segment in the image,  and there is a random variable $o_{i,j}$ for each pair of neighboring segments. 

\begin{figure}[t]
\begin{center}
\begin{tabular}{cc}
\includegraphics[width=0.45\columnwidth]{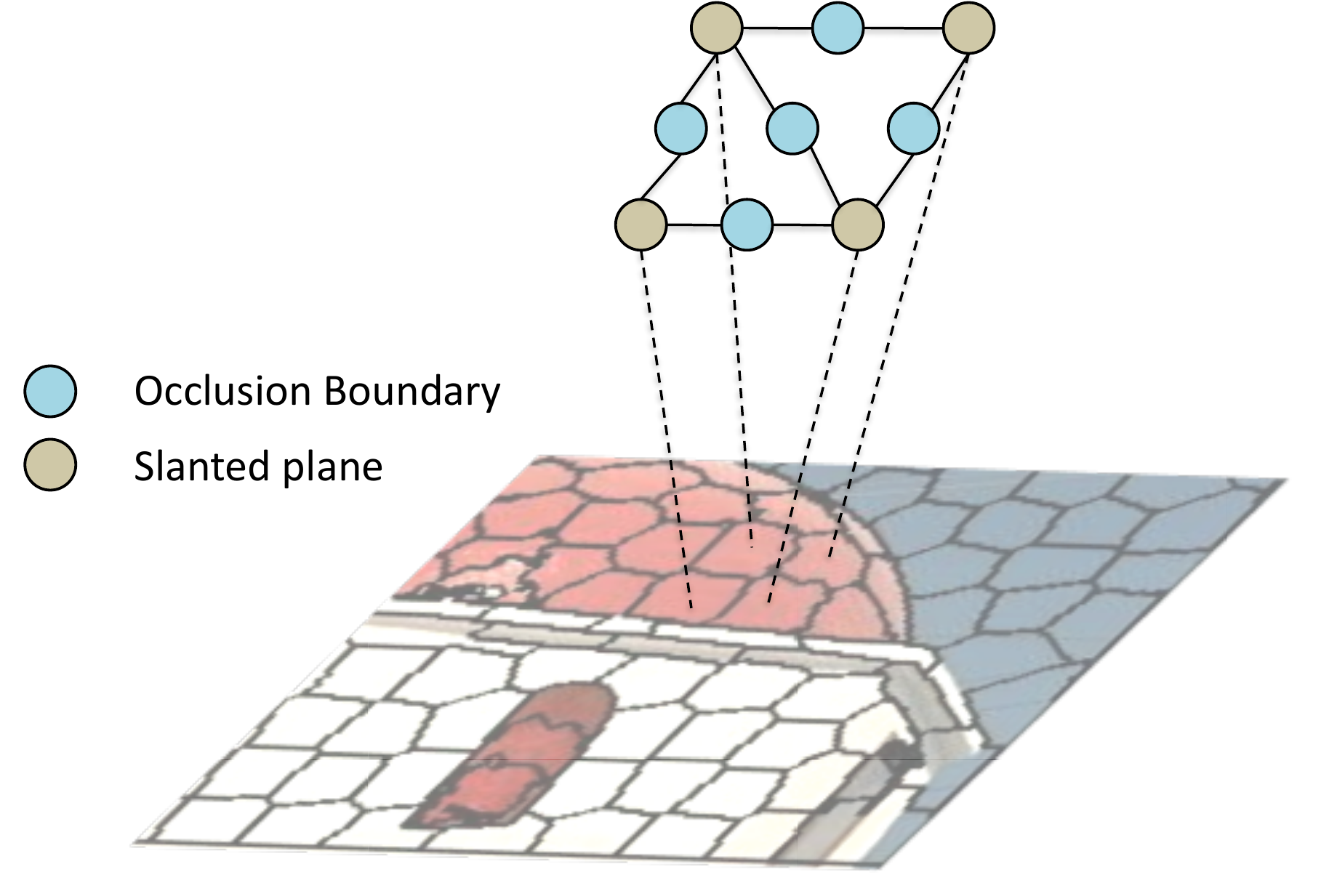} &
\includegraphics[width=0.4\columnwidth]{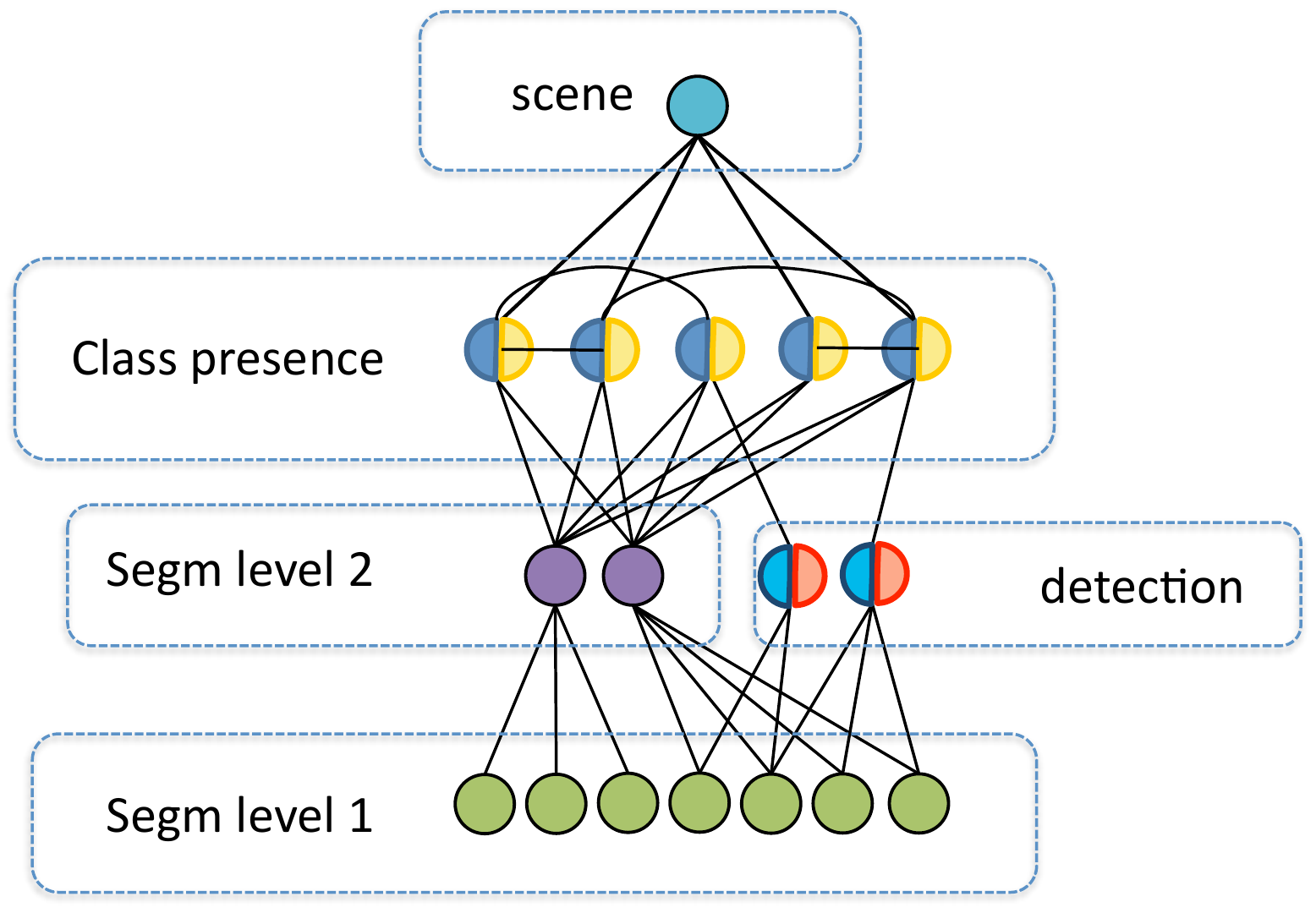} \\ 
 (Stereo) & (Recognition)
\end{tabular}
\end{center}
\caption{{\bf Graphical models} for (a) stereo (b) recognition. }
\label{fig:graph}
\end{figure}

We now briefly describe the potentials we employ, and  refer the reader to \cite{Yamaguchi12} for more details.
We utilize a truncated quadratic disparity potential, $\phi_i^{\mathrm{seg}}(\mathbf{y}_i)$,   which encodes that the plane should agree with the results of the matching along the epipolar lines. We additional incorporate 3-way boundary potentials $\phi_{ij}^{\mathrm{bdy1}}(o_{ij}, \by_i, \by_j)$ linking our discrete and continuous variables. In particular, these potentials express the fact  that when two neighboring planes are hinge or coplanar they should agree on the boundary, and when a segment occludes another segment, the boundary should be explained by the occluder.
We impose a regularization on the type of occlusion boundary, where we prefer simpler explanations (i.e., coplanar is preferable than hinge which is more desirable than occlusion). This is encoded in $\phi_{ij}^{\mathrm{bdy2}}(o_{ij}, \by_i, \by_j)$.
We introduce a potential $\phi^{\mathrm{occ}}_{ij}(\by_{\mathrm{front}}, \by_{\mathrm{back}})$ which ensures that the discrete occlusion labels match well the disparity  observations, as well as an additional potential $\phi^{\mathrm{neg}}_{ij}(\by_i)$ which penalizes negative disparities.
Following work on occlusion boundary reasoning \cite{Malik87,Hoiem07}, we utilize higher order potentials to encode whether junctions of three and four planes are possible. This is encoded in $\phi_{ijk}^{\mathrm{jct}}(o_{ij}, o_{jk}, o_{ik})$  and  $ \phi_{pqrs}^{crs}(o_{pq}, o_{qr}, o_{rs}, o_{ps}) $ respectively.
Finally, we employ a simple color potential to reason about segmentation, which is defined in terms of the $\chi$-squared distance between color histograms of neighboring segments. 
This potential, $\phi_{ij}^{\mathrm{col}}(o_{ij})$,  encodes the fact that we expect segments which are coplanar to have similar color statistics (i.e., histograms), while the entropy of this distribution is higher when the planes form an  occlusion boundary or a hinge. Fig. \ref{fig:graph} (left) illustrates the graphical model. 
Thus our hybrid graphical model is defined in terms of the following energy function
\begin{eqnarray}
E(\by,\bo) &=& \sum_i^{|\by|} w^{\mathrm{seg}}\phi_i^{\mathrm{seg}}(\mathbf{y}_i) + \sum_{(i,j) \in {\cal E}_{bdy}} w^{\mathrm{bdy1}}\phi_{ij}^{\mathrm{bdy1}}(o_{ij}, \by_i, \by_j) +\sum_{(i,j) \in  {\cal E}_{bdy}} w^{\mathrm{occ}}\phi^{\mathrm{occ}}_{ij}(\by_{\mathrm{front}}, \by_{\mathrm{back}}) \nonumber \\
&& + \sum_{i,j \in {\cal E}} w^{\mathrm{bdy2}}\phi_{ij}^{\mathrm{bdy2}}(o_{ij}, \by_i, \by_j) +  \sum_i^{|\by|} w^{\mathrm{neg}}\phi^{\mathrm{neg}}_{ij}(\by_i) + \sum_{(i,j,k)\in  {\cal E}_{jct}} w^{\mathrm{jct}}\phi_{ijk}^{\mathrm{jct}}(o_{ij}, o_{jk}, o_{ik}) \nonumber \\
&&   + \sum_{(p,q,r,s)\in  {\cal E}_{crs}} w^{\mathrm{crs}}  \phi_{pqrs}^{crs}(o_{pq}, o_{qr}, o_{rs}, o_{ps})  \nonumber 
\end{eqnarray}

We learn the  the weights $w^{\mathrm{bdy1}}$, $w^{\mathrm{occ}}$, $w^{\mathrm{bdy2}}$, $w^{\mathrm{neg}}$, $w^{\mathrm{jct}}$, $w^{\mathrm{crs}}$ with our approach, and set   $\epsilon=1$ and $C$ to be equal to the number of training examples.  
We also employ particle convex belief propagation (PCBP) \cite{Peng11} for inference. PCBP is an iterative algorithm that works as follows: For each random variable, particles are sampled around the current solution. These samples act as labels in a discretized graphical model which is solved to convergence using convex belief propagation~\cite{Hazan10}. The current solution is then updated with the MAP estimate obtained on the discretized graphical model. This process is repeated for a fixed number of iterations. In our implementation, we use the distributed message passing algorithm of~\cite{Schwing11} 
to solve the discretized graphical model at each iteration. 

We perform experiments on the challenging KITTI dataset \cite{Geiger12}, which  is the only real-world stereo dataset with accurate ground truth. 
It is composed of 194 training and 195 test high-resolution images ($1237.1 \times 374.1$ pixels) captured from an autonomous driving platform driving around in a urban environment. The ground truth is generated by means of a Velodyne sensor which is calibrated with the stereo pair. This results in  semi-dense ground truth covering approximately  30 \% of the pixels. We employ 20 images for training, and utilize the remaining 174 images for validation purposes.  

\begin{figure}[t]
\vspace{-0.3cm}
\begin{center}

%
%
%

\vspace{0mm}
\includegraphics[width=4cm]{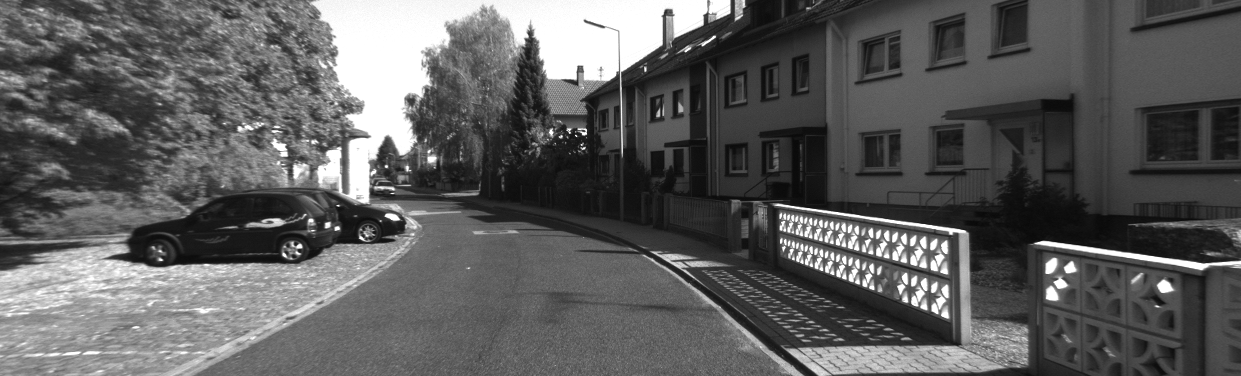}
\includegraphics[width=4cm]{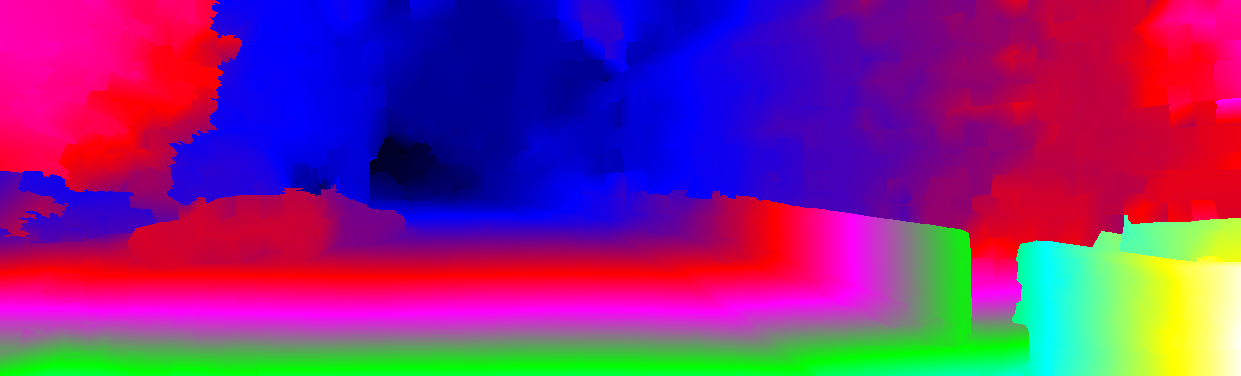}
\includegraphics[width=4cm]{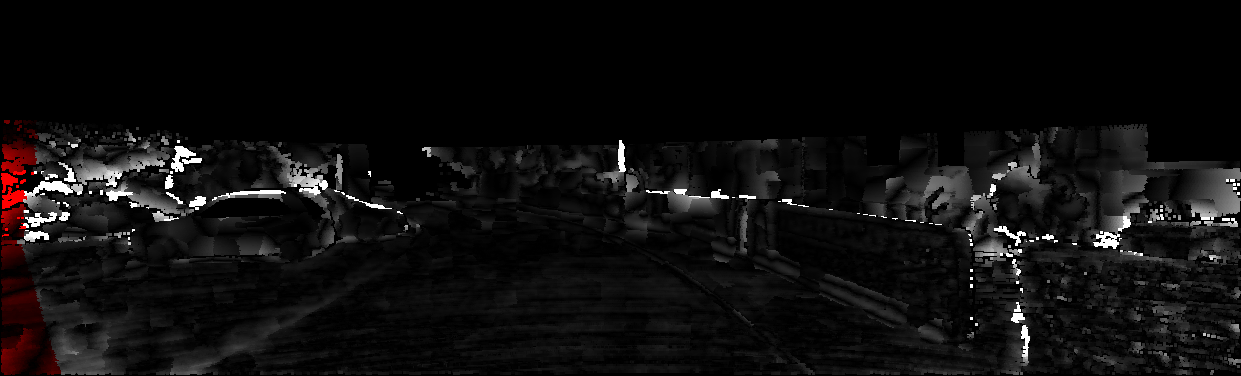}

\includegraphics[width=4cm]{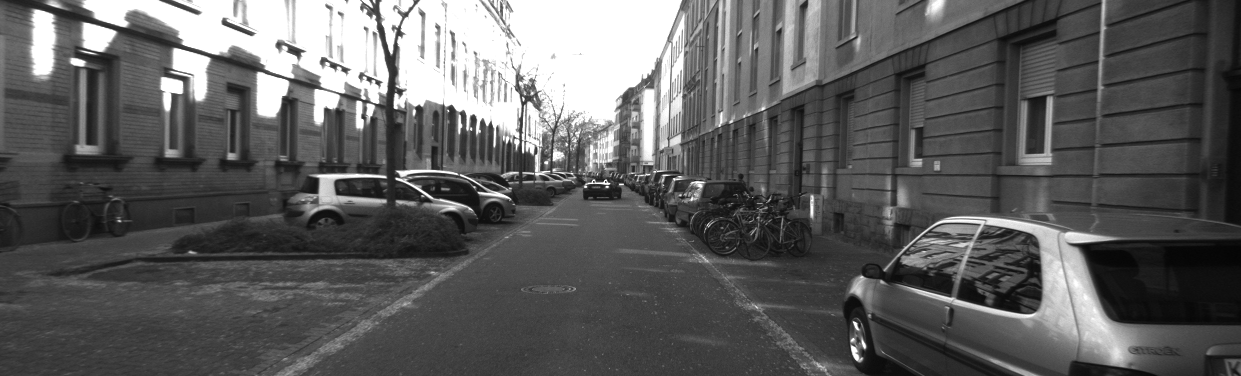}
\includegraphics[width=4cm]{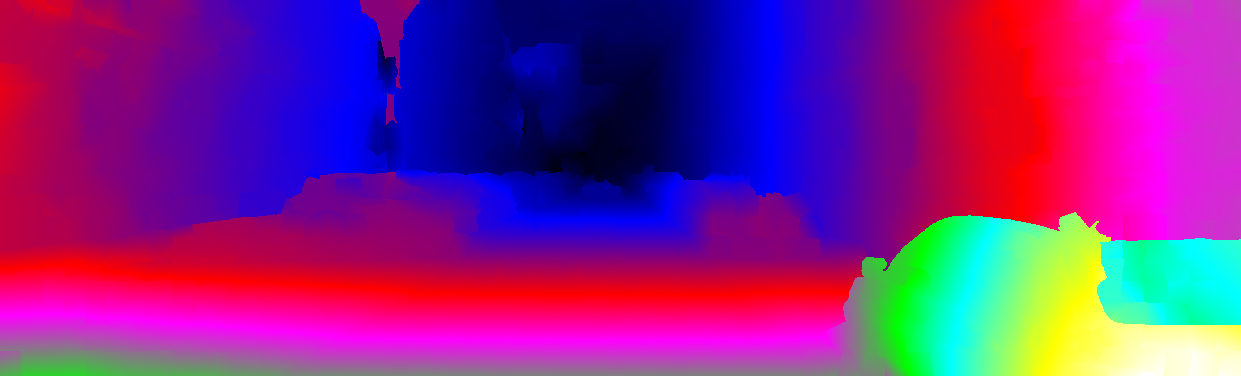}
\includegraphics[width=4cm]{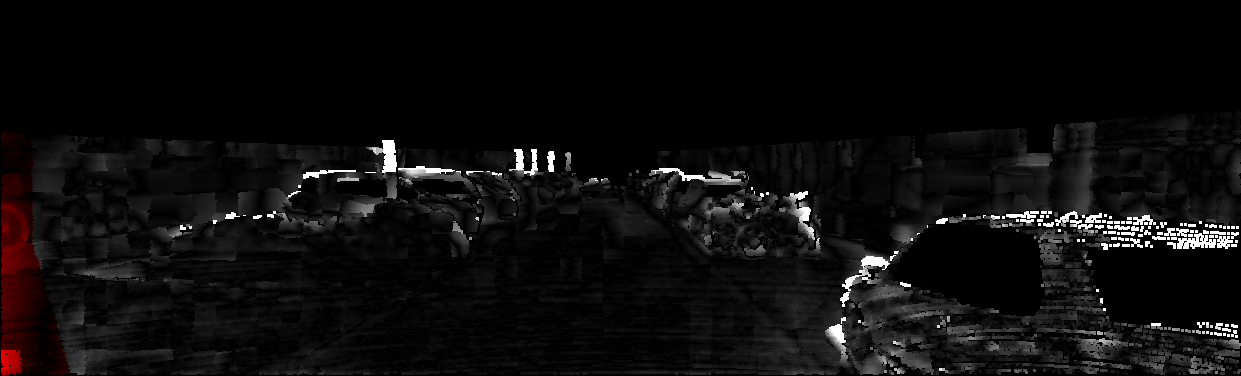}
\end{center}
\vspace{-0.7cm}
\caption{ KITTI examples. (Left) Original. (Middle) Disparity. (Right) Disparity errors.}
\label{fig:example}
\vspace{-0.2cm}
\end{figure}

We employ two different metrics. The first one measures  the average number of non-occluded pixels which error is bigger than  a fixed threshold.
To test the extrapolation capabilities of the different approaches, the second metric computes the same metric, but including the occluding pixels as well.
We employ this metrics as our loss. 
Table \ref{tab:kitti}  depict results of our approach and the baselines in terms of the two metrics. 
Note that our approach significantly outperforms all the baselines in all settings (i.e., thresholds bigger than 2, 3, 4 and 5 pixels). 
 Fig. \ref{fig:example} depicts an illustrative set of KITTI examples. Despite the challenges, our approach does a good job at estimating disparities. 
 
\begin{table}[t]
\begin{center}
\begin{scriptsize}
\begin{tabular}{| c | c | c |c | c | c | c | c | c |}
\hline
& \multicolumn{2}{|c|}{  $>$ 2 pixels} & \multicolumn{2}{|c|}{  $>$ 3 pixels}& \multicolumn{2}{|c|}{  $>$ 4 pixels}& \multicolumn{2}{|c|}{  $>$ 5 pixels}\\ \hline
& {\bf Non-Occ} & {\bf All} & {\bf Non-Occ} & {\bf All}& {\bf Non-Occ} & {\bf All}& {\bf Non-Occ} & {\bf All}
\\\hline
GC+occ \cite{Kolmogorov01} & 39.76 \% & 40.97 \% & 33.50 \% & 34.74 \%& 29.86 \% & 31.10 \% & 27.39 \% & 28.61 \%
\\\hline
OCV-BM \cite{Bradski00} & 27.59 \% & 28.97 \%  & 25.39 \% & 26.72 \%& 24.06 \% & 25.32 \% & 22.94 \% & 24.14 \%
 \\\hline
CostFilter \cite{Rhemann11} & 25.85 \% & 27.05 \% & 19.96 \% & 21.05 \%& 17.12 \% & 18.10 \% & 15.51 \% & 16.40 \%
\\\hline
GCS \cite{Cech07} & 18.99 \% & 20.30 \% & 13.37 \% & 14.54 \% & 10.40 \% & 11.44 \% & 8.63 \% & 9.55 \%
\\\hline
GCSF \cite{Cech11} & 20.75 \% & 22.69 \%   & 13.02 \% & 14.77 \% & 9.48 \% & 11.02 \% & 7.48 \% & 8.84 \%
 \\\hline
SDM \cite{Kostkova03} & 15.29 \% & 16.65 \% & 10.98 \% & 12.19 \% & 8.81 \% & 9.87 \% & 7.44 \% & 8.39 \%
\\ \hline
ELAS \cite{Geiger10} & 10.95 \% & 12.82 \% & 8.24 \% & 9.95 \% & 6.72 \% & 8.22 \% & 5.64 \% & 6.94 \%
\\\hline
OCV-SGBM \cite{Hirschmueller08} & 10.58 \% & 12.20 \% & 7.64 \% & 9.13 \% & 6.04 \% & 7.40 \% & 5.04 \% & 6.25 \%
\\ \hline 
ITGV \cite{Ranftl2012} & 8.86 \% & 10.20 \% & 6.31 \% & 7.40 \% & 5.06 \% & 5.97 \% & 4.26 \% & 5.01 \%
\\ \hline 
{\bf Ours} & {\bf 6.25} \% & {\bf 7.78} \% & {\bf 4.13} \% & {\bf 5.45} \%  & {\bf 3.18} \% & {\bf 4.32} \% & {\bf 2.66} \% & {\bf 3.66} \%
\\\hline
\end{tabular}
\end{scriptsize}
\end{center}
\vspace{-0.2cm}
\caption{Comparison with the state-of-the-art on the test set of KITTI \cite{Geiger12}}
\label{tab:kitti}
\end{table}

\subsection{Semantic Segmentation}

While there has been significant progress in solving tasks such as  image labeling~\cite{s:ladicky10b}, object detection~\cite{s:felzenswalb10} and scene classification~\cite{s:xiao10},
existing approaches could benefit from solving these problems jointly~\cite{s:heitz08}.
For example,  segmentation should be easier if we know where the object of interest is. Similarly, if we know the type of the scene, we can narrow down the classes we are expected to see,  e.g., if we are looking at the sea, we are more likely to see a boat than a cow. Conversely, if we know which semantic regions (e.g., sky, road) and which objects are present in the scene, we can more accurately infer the scene type. 
Holistic scene understanding aims at 
recovering multiple related aspects of a scene so as to provide a deeper understanding of the scene as a whole. 

In recent work, \cite{Yao12}, we have proposed an approach to holistic scene understanding that simultaneously reasons about regions,  location, class and spatial extent of objects,  as well as the type of scene. 
We frame the holistic problem as a structured prediction problem in a graphical model   defined over hierarchies of regions of different sizes,  as well as  auxiliary variables encoding the scene type,  the presence of a given class in the scene,
 and  the correctness of the  bounding boxes output by an object detector.
 For objects with well-defined shape (e.g., cow, car), we additionally incorporate a shape prior that takes the form of a soft mask learned from training examples.
 Unlike existing approaches  that reason at the (super-) pixel  level, we employ~\cite{j:Arbelaez11} to obtain (typically large) regions which respect boundaries well. This enables us to represent the problem using only a small number of variables.
 Learning and inference are efficient in our model as the  auxiliary variables we utilize allow us to decompose the inherent high-order potentials into pairwise potentials between a few variables with small number of states (at most the number of classes). 
 
 We now briefly describe the graphical model as well as the potentials employed. 
 We refer the reader to Fig.~\ref{fig:graph} (right) for  an overview of our model, and to \cite{Yao12} for more details and results.
Let $x_i \in \{1, \cdots, C \}$ be a random variable representing the class label of the i-th segment in the lower level of the hierarchy, while $y_j \in \{1, \cdots, C \}$ is a random variable associated with the class label of the j-th segment of the second level of the hierarchy.
Following recent approaches 
\cite{s:ladicky10b,s:lee10}, we represent the detection problem with a set of candidate bounding boxes.
Let $b_l \in \{0,1\}$ be a binary random variable associated with a candidate detection, taking value  $0$ when the detection is a false detection.
We use the detector of \cite{s:felzenswalb10} to generate candidate detections, which provides us with an object class, a score, the location and aspect ratio of the bounding box, as well as the root mixture component ID that has generated the detection.
The latter gives us information about the expected shape of the object.
 Let $z_k \in \{0,1\}$ be a random variable which takes value $1$ if class $k$ is present in the image, and let  $\bs \in \{1,\dots, C_l\}$ be a random variable representing the scene type. 

We define our {\it holistic  conditional random field}  as
\begin{equation}
p(\ba) = p(\bx,\by,\bz,\bb,s) = \frac{1}{Z} \prod_{type}  \prod_{r}\psi^{type}_{r}(\ba_{r})
\vspace{-0.2cm}
\label{eq:crf}
\end{equation}
where $\ba = (\bx,\by,\bz,\bb,s)$ represents the set of all segmentation random variables, $\bx$ and $\by$, 
 the set of  $C$ binary random variables $\bz$ representing the presence of the different classes in the scene,  the set of all candidate detections $\bb$, and $\psi_\alpha^{type}$ encodes potential functions over sets of variables. Note that the variables in a region $r$  can be of the same task (e.g.,  two segments) or different tasks (e.g., detection and segmentation).

We compute the unary potential for each region at segment $\phi_i^{\mathrm{xseg}}({x}_i)$ and super-segment  level $\phi_i^{\mathrm{yseg}}({y}_j)$  by averaging the TextonBoost  \cite{s:ladicky10b}  pixel  potentials inside each region.
We use  $P^n$ potentials \cite{Kohli09}, $\phi_{ij}^{\mathrm{xy}}(x_i,y_j) $, to encourage that segments and supersegments agree on their class labels. Additionally, unary potentials $\phi_i^{\mathrm{det}}({b}_i)$ represent the score of the detector for that hypothesis squash by a sigmoid. 
We train a classifier for each scene type and represent its score in $\phi^{\mathrm{scene}}({s})$.
We additionally incorporate the shape prior by placing a mask representing the typical shape of the training examples that fell in that  DPM component, and encouraging the segments inside the bounding box to take the same label as the detector, with strength proportional to the mask value on that segment. This is encoded in $\phi_{ij}^{\mathrm{shape}}(b_i,x_j)$.
We also incorporate statistics of class occurrance and co-occurances as unary and pairwise potentials $\phi_i^{\mathrm{stats}}({z}_i)$ and $\phi_{ij}^{\mathrm{co-occ}}(z_i,z_j)$ respectively. $\phi_{ij}^{\mathrm{zy}}(z_i,y_j)$ ensures that the classes that are inferred to be present in the scene are compatible with the classes that are chosen at the segment level, while $\phi_{ij}^{\mathrm{bz}}(b_i,z_j)$ ensures that when a bounding box is on, its class is also present in the scene.
Finally, $\phi_{ij}^{\mathrm{sz}}(s,z_j)$  encodes statistics of class occurrences for each scene type.
The energy of the holistic graphical model is then defined as 

\begin{eqnarray}
E(\bx,\by,\bz,\bb,s) &=& w^{\mathrm{scene}}\phi^{\mathrm{scene}}({s})  + \sum_i^{|\bx|} w^{\mathrm{xseg}}\phi_i^{\mathrm{xseg}}({x}_i) + \sum_i^{|\by|} w^{\mathrm{yseg}}\phi_i^{\mathrm{yseg}}({y}_i)  +
\sum_i^{|\bz|} w^{\mathrm{stats}}\phi_i^{\mathrm{stats}}({z}_i)  \nonumber \\
&& 
+ \sum_i^{|\bb|} w^{\mathrm{det}}\phi_i^{\mathrm{det}}({b}_i) + \sum_{(i,j) \in {\cal E}_{z}} w^{\mathrm{co-occ}}\phi_{ij}^{\mathrm{co-occ}}(z_i,z_j) + \sum_{(i,j) \in {\cal E}_{xy}} w^{\mathrm{xy}}\phi_{ij}^{\mathrm{xy}}(x_i,y_j) 
 \nonumber \\
&&  +\sum_{(i,j) \in {\cal E}_{bz}} w^{\mathrm{bz}}\phi_{ij}^{\mathrm{bz}}(b_i,z_j) + 
\sum_{(i,j) \in {\cal E}_{bx}} w^{\mathrm{shape}}\phi_{ij}^{\mathrm{shape}}(b_i,x_j) +
\sum_{i=1}^{|\bz|} w^{\mathrm{sz}}\phi_{ij}^{\mathrm{sz}}(s,z_j)
 \nonumber  \\
&& + \sum_{(i,j) \in {\cal E}_{zy}} w^{\mathrm{zy}}\phi_{ij}^{\mathrm{zy}}(z_i,y_j) 
   \nonumber 
\end{eqnarray}

We employ our approach to learn the weights with $\epsilon=1$ and $C=0.02$. 
To deal with our holistic setting, we employ  a holistic loss which takes into account all tasks. We define it to be a weighted sum of losses, each one designed for a particular task, e.g., detection, segmentation.
In order to do efficient learning, it is important that the losses decompose as a sum of functions on small subsets of variables.
Here, we define loss functions which decompose into unitary terms. 
In particular, we define the segmentation loss at each level of the hierarchy to be the percentage  of wrongly predicted pixels. This decomposes as sums of unitary terms (one for each segment).
We utilize a 0-1 loss for the variables encoding the classes that are present in the scene, which also decomposes as the sum of unitary 0-1 losses on each $z_k$.
We define a 0-1 loss over the scene type, and a PASCAL loss over the detections which decomposes as the sum of losses for each detection. 

We test our approach on the tasks of semantic segmentation on the MSRC-21 dataset ~\cite{s:shotton08}. 
We employ~\cite{j:Arbelaez11} to obtain regions which respect boundaries well, and   set the watershed threshold  to be $0.08$ and $0.16$ for the two layers in the hierarchy. To create the unitary potentials for the scenes, we use a standard bag-of-words spatial pyramid with $1,2$ and $4$ levels over a $1024$ sparse coding dictionary on SIFT features, colorSIFT, RGB histograms and color moment invariants, and train a linear  one-vs-all SVM classifier. 
We use the detector of~\cite{s:felzenswalb10} to generate candidate detections. 
For each detector we lowered the  threshold to produce over-detections.
We follow Felzenswalb et al.'s entry in PASCAL'09 to compute the soft shape masks. For each class we ran the detector on the training images and chose those that overlaped with groundtruh more than $0.5$ in the intersection over union measure. For each positive detection we also recorded the winning component. We compute the mask for each component by simply averaging the groundtruth class regions inside the assigned groundtruth boxes. Prior to averaging, all bounding boxes were warped to the same size, i.e., the size of the root filter of the component. To get the shape mask  for each detection we warped the average mask of the detected component to the predicted bounding box.

MSRC-21 contains classes such as \emph{sky, water}, as well as more shape-defined classes such as \emph{cow, car}. We manually annotated bounding boxes for the latter classes, with a total of $15$ classes and $934$ annotations.
We also annotated $21$ scenes, taking the label of the salient object in the image, if there is one, or a more general label such as ``city'' or ``water'' otherwise.
We follow the standard error measure of  average per-class accuracy as well as average per-pixel accuracy, denoted as global~\cite{s:ladicky10}. We used the standard train/test split~\cite{s:shotton08} to train the full model, the pixel unary potential, object detector and scene classifier.

\begin{table*}[!ht]
{\scriptsize
\begin{center}
\renewcommand{\arraystretch}{0.945}
\addtolength{\tabcolsep}{-2.95pt}
\begin{tabular}{l|ccccccccccccccccccccc|c c}
 & \rotatebox{90}{building} & \rotatebox{90}{grass} & \rotatebox{90}{tree} & \rotatebox{90}{cow} & \rotatebox{90}{sheep} & \rotatebox{90}{sky} & \rotatebox{90}{aeropl.} & \rotatebox{90}{water} & \rotatebox{90}{face} & \rotatebox{90}{car} & \rotatebox{90}{bicycle} & \rotatebox{90}{flower} & \rotatebox{90}{sign} & \rotatebox{90}{bird} & \rotatebox{90}{book} & \rotatebox{90}{chair} & \rotatebox{90}{road} & \rotatebox{90}{cat} & \rotatebox{90}{dog} & \rotatebox{90}{body} & \rotatebox{90}{boat} & \rotatebox{90}{{\bf average}} & \rotatebox{90}{{\bf global}}\\ 
\hline
\hline
\cite{s:shotton08} & 49 & 88 & 79 & {\bf 97} & {\bf 97} & 78 & 82 & 54 & 87 & 74 & 72 & 74 & 36 & 24 & 93 & 51 & 78 & 75 & 35 & 66 & 18 & 67 & 72 \\
\cite{s:jiang09} & 53 & 97 & 83 & 70 & 71 & 98 & 75 & 64 & 74 & 64 & 88 & 67 & 46 & 32 & 92 & 61 & 89 & 59 & {\bf 66} & 64 & 13 & 68 & 78 \\
\cite{s:gonfaus10} & 60 & 78 & 77 & 91 & 68 & 88 & 87 & 76 & 73 & 77 & 93 & 97 & 73 & {\bf 57} & 95 & {\bf 81} & 76 & 81 & 46 & 56 & {\bf 46} & 75 & 77 \\
\cite{s:ladicky10} & 74 & 98 & 90 & 75 & 86 & {\bf 99} & 81 & 84 & {\bf 90} & 83 & 91 & {\bf 98} & 75 & 49 & 95 & 63 & {\bf 91} & 71 & 49 & {\bf 72} & 18 & 77.8 & {\bf 86.5}\\
\cite{s:koltun11} & 75 & {\bf 99} & {\bf 91} & 84 & 82 & 95 & 82 & 71 & 89 & {\bf 90} & {\bf 94} & 95 & {\bf 77} & 48 & 96 & 61 & 90 & 78 & 48 & 80 & 22  & 78.3 & 86.0\\
Ours & 71 & 98 & 90 & 79 & 86 & 93 & {\bf 88} & {\bf 86} & {\bf 90} & 84 & {\bf 94} & {\bf 98} & 76 & 53 & {\bf 97} & 71 & 89 & {\bf 83} & 55 & 68 & 17 & {\bf 79.3} & 86.2\\
\hline
\end{tabular}
\end{center}
}
\vspace{-2mm}
\caption{MSRC-21 segmentation results}
\label{tab:resultsAll}
\end{table*}

\begin{figure*}[htb!]
\vspace{-0.3cm}
\centering
\begin{minipage}{0.495\linewidth}
\includegraphics[width=\linewidth,height=3.92cm]{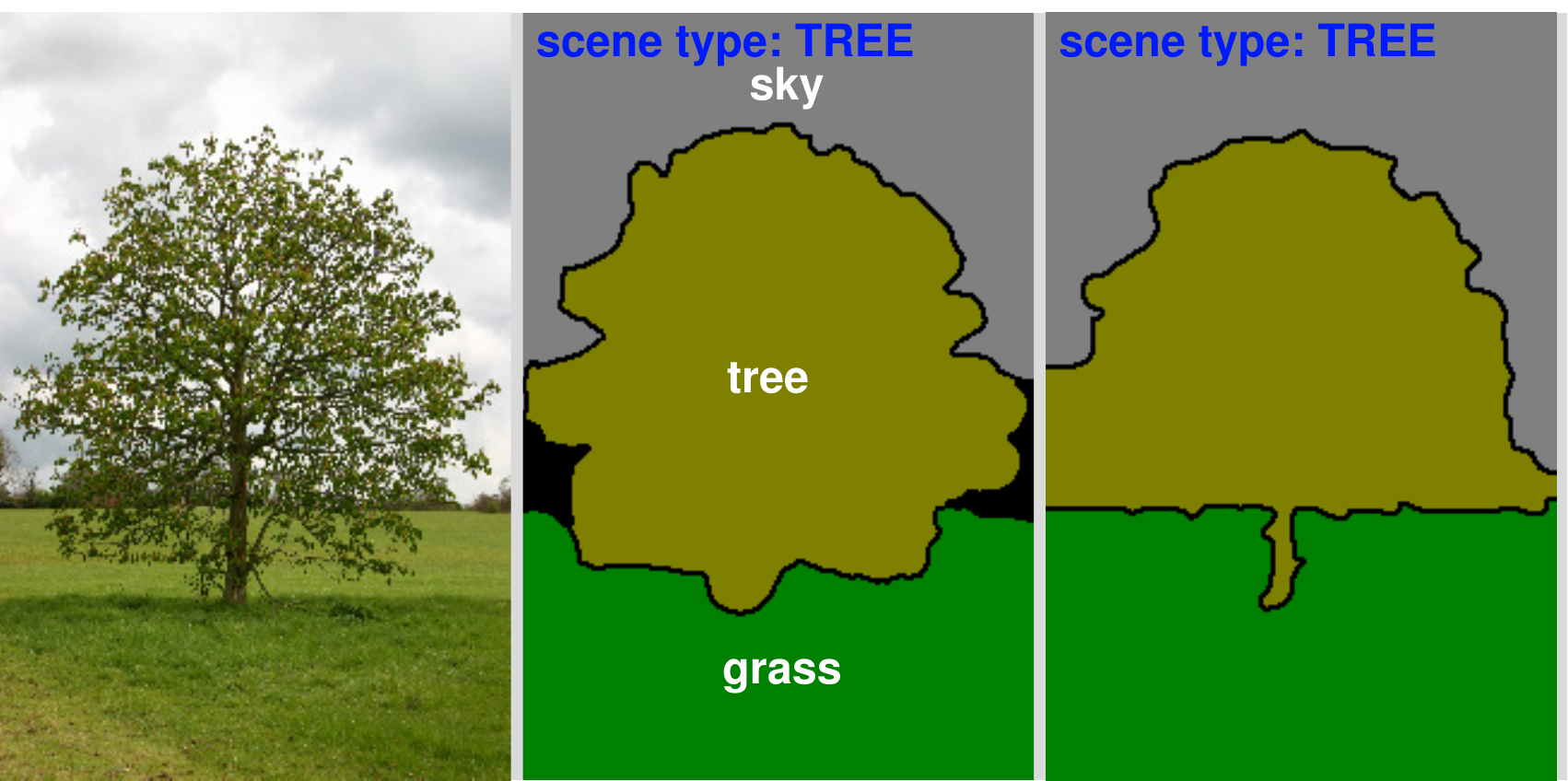}
\end{minipage}
\begin{minipage}{0.495\linewidth}
\includegraphics[width=\linewidth]{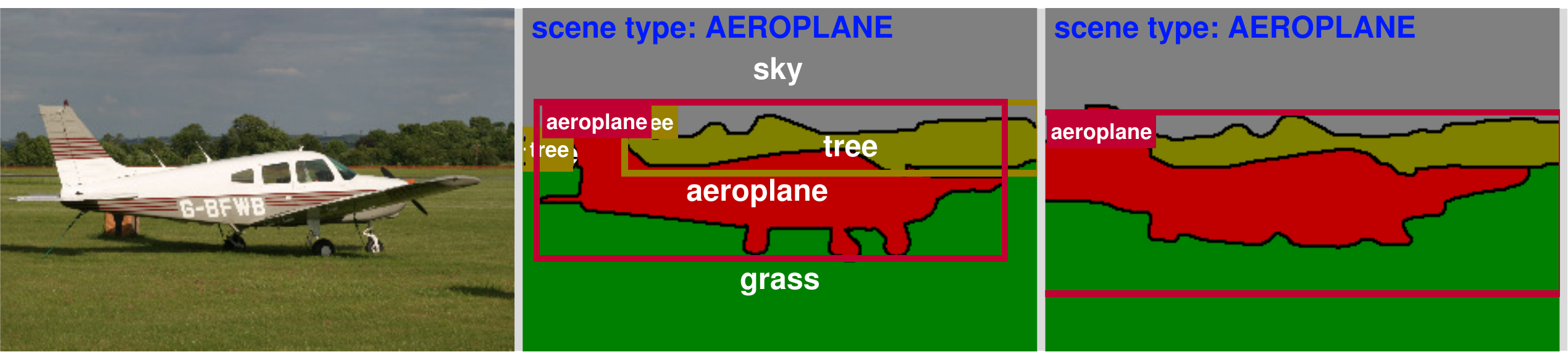}\\
\includegraphics[width=\linewidth]{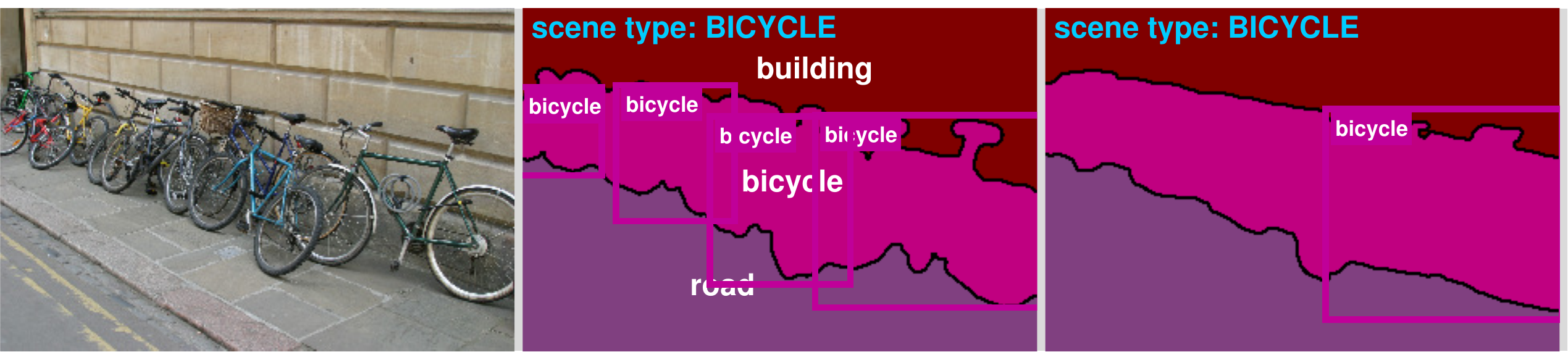}
\end{minipage}
\includegraphics[width=0.495\linewidth,height=1.94cm]{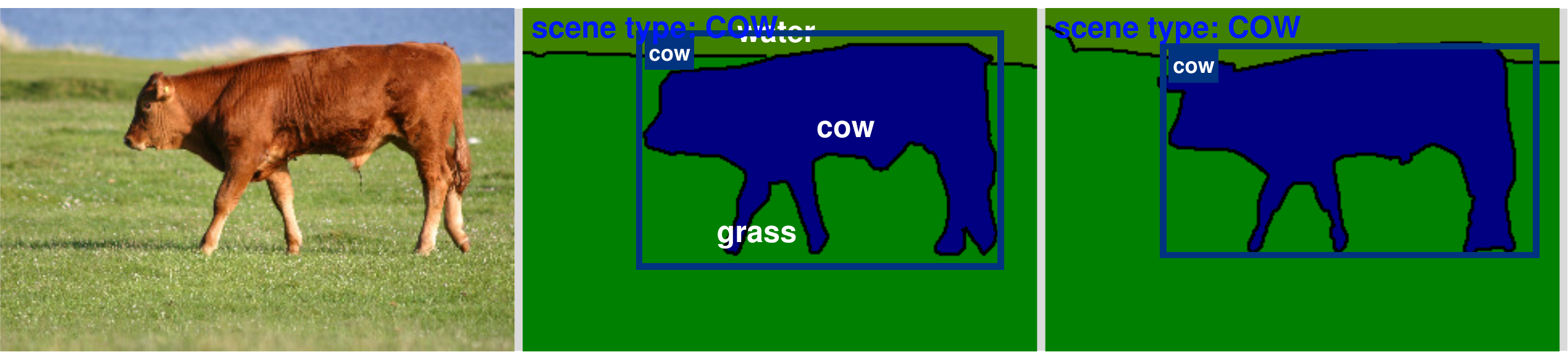}
\includegraphics[width=0.495\linewidth,height=1.94cm]{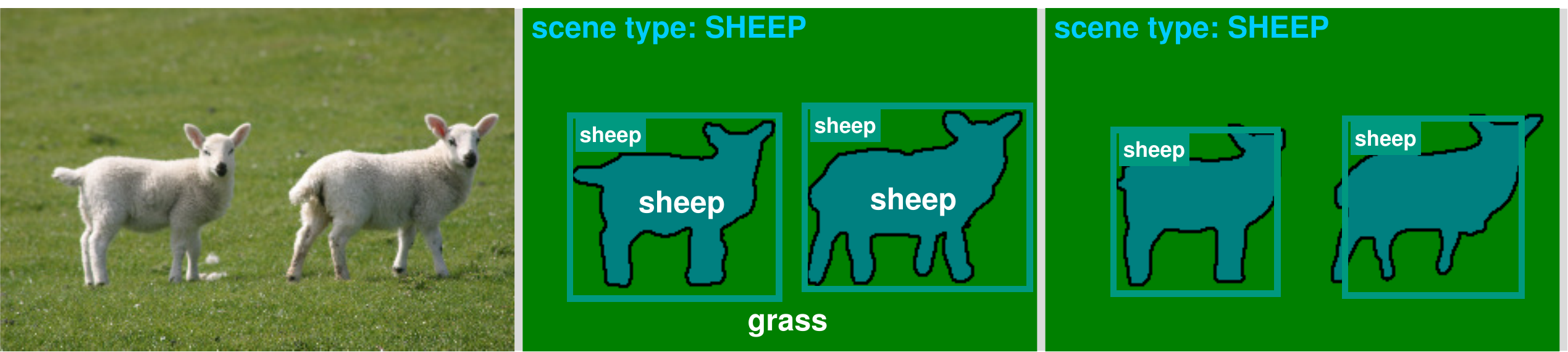}\\
(Successful  cases)\\
\includegraphics[width=0.495\linewidth]{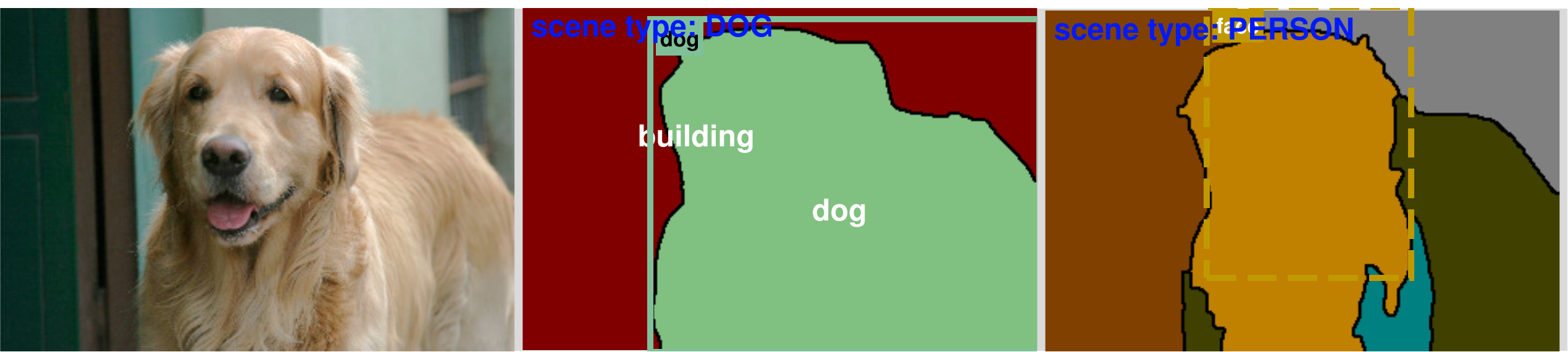}
\includegraphics[width=0.495\linewidth]{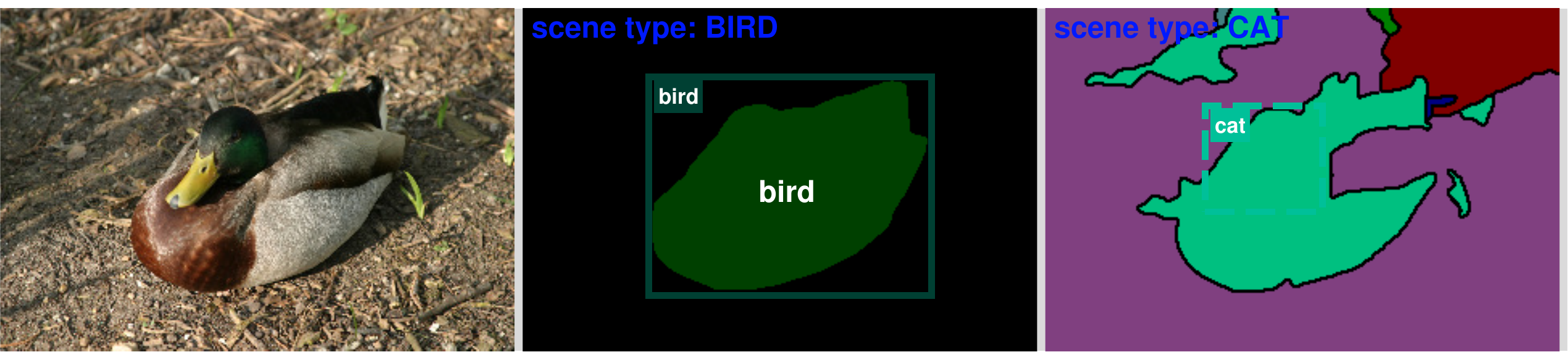}\\
(Failure modes)
\vspace{-0.1cm}
\caption{Segmentation examples: (image, groundtruth, our holistic scene model)}
 \label{fig:results-good}
 \vspace{-0.3cm}
\end{figure*}

Table~\ref{tab:resultsAll} reports the segmentation accuracy, along with the comparisons with the existing state-of-the-art.  Our joint model 
achieves the highest average accuracy reported
on this dataset to date. Furthermore,  the joint model not only improves segmentation accuracy but also significantly boosts object detection and scene classification. Scene classification improves from 79.5\% to 80.6\%, while detection improves from 44.6\% to 50.7\% recall at equal false positive rate.   The average precision of the detector also improves from 48.2\% to 49.3\%. This is notable as context re-scoring \cite{s:felzenswalb10} fails and reduces performance to 45.7\%. We conjecture that this is due to the small number of training examples. 
Fig. \ref{fig:results-good} shows some good segmentation examples, as well as some failure modes, which are due to very bad unary segmentation potentials or when several tasks agree on the wrong class. 

\subsection{3D indoor scene understanding}

\begin{figure}[t]
\vspace{-0.2cm}
\centering
\begin{minipage}{0.495\linewidth}
\includegraphics[width=4.5cm]{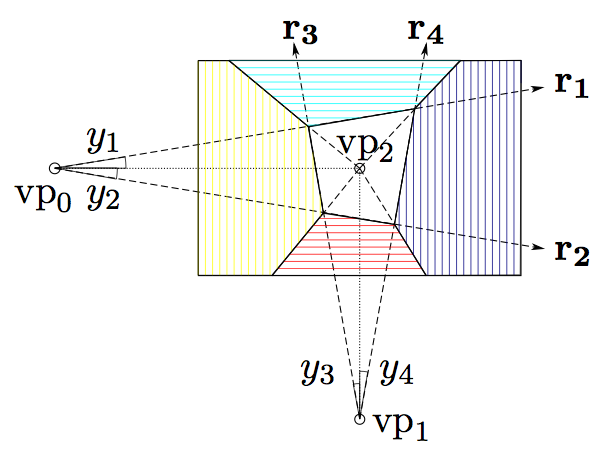}
\end{minipage}
\begin{small}
\begin{tabular}{|c||c|c|c|}
\hline
&  OM & GC & OM + GC \\ \hline \hline
\cite{Hoiem07}  &- &28.9&- \\ \hline
\cite{Hedau09} (a) & - &26.5& -\\ \hline
\cite{Hedau09} (b)  &-  &21.2 &-\\ \hline
\cite{Wang10}  & 22.2 &-&- \\ \hline
\cite{Lee10}  & 24.7 & 22.7 & 18.6\\ \hline
Ours & {\bf 18.6} & {\bf 15.4} & {\bf 13.6}\\ \hline
\end{tabular}
\end{small}
\caption{{\bf Layout estimation:} (Left) Parameterization of the problem. (Right) Comparison to the state-of-the-art that uses the same image information on the layout data set of~\cite{Hedau09}. Pixel classification error is given in \%.}
\label{tab:state_feat}
\vspace{-0.2cm}
\end{figure}

Most existing approaches to recovering the spatial layout of indoor scenes rely on the {\it Manhattan world} assumption, which states that there exist three  dominant vanishing points 
which are orthogonal. 
They typically formulate the problem 
as a structured prediction task, which estimates the 3D
box that best approximates the scene layout~\cite{Hedau09,Lee10,Wang10}. 
Two different  parameterizations have been proposed  for this problem, both  assuming that the three dominant vanishing points can be reliably detected. In~\cite{Hedau09,Lee10}, candidate 3D boxes are generated, and inference is formulated in terms of a single high dimensional discrete random variable. 
Hence, one state of such a variable denotes one candidate 3D layout. 
This limits significantly the amount of candidate boxes, e.g., only $\approx 1000$ candidates are employed in~\cite{Hedau09}.
In contrast, \cite{Wang10} parameterize the layout with four discrete random variables, 
 that correspond to the angles encoding the rays that originate from the respective vanishing points. 
 An illustration of this parameterization is shown in Fig.  \ref{tab:state_feat} (left).

Existing approaches employ potentials based on different image information. Geometric context~\cite{Hoiem07}, orientation maps~\cite{Lee09} as well as lines in accordance with vanishing points~\cite{Wang10} are amongst the most successful cues. 
The complexity of learning and inferece is determined by the order of the potentials - the number of variables involved and their size - that encode the image features. 
These potentials are typically unary, pairwise as well as higher-order (i.e., order four), and count for each face the number of pixels labeled with a particular label.
The order is even higher when reasoning about clutter in the form of hidden variables~\cite{Wang10} (i.e., order five) or objects present in the scene that restrict the hypothesis space~\cite{Lee10}.
While the aforementioned  approaches perform well in practice, to tractably handle learning and inference with both parameterizations, reductions on the search space were proposed and/or a limited amount of labelings was considered.

In contrast, in recent work \cite{Schwing12-cvpr} we have proposed a novel and efficient approach to discriminatively predict the 3D layout of indoor scenes. In particular, we generalize the concept of integral images to ``integral geometry,'' by constructing accumulators in accordance with the vanishing points.  We showed that utilizing this concept, as all potentials represent counts, their computation can be reduced to sums of pairwise potentials.
As a result, learning and inference is possible without further reduction of the search space.

\begin{figure*}
\centering
	\subfigure[Error: 1.25\%] {
		\includegraphics[width=2cm]{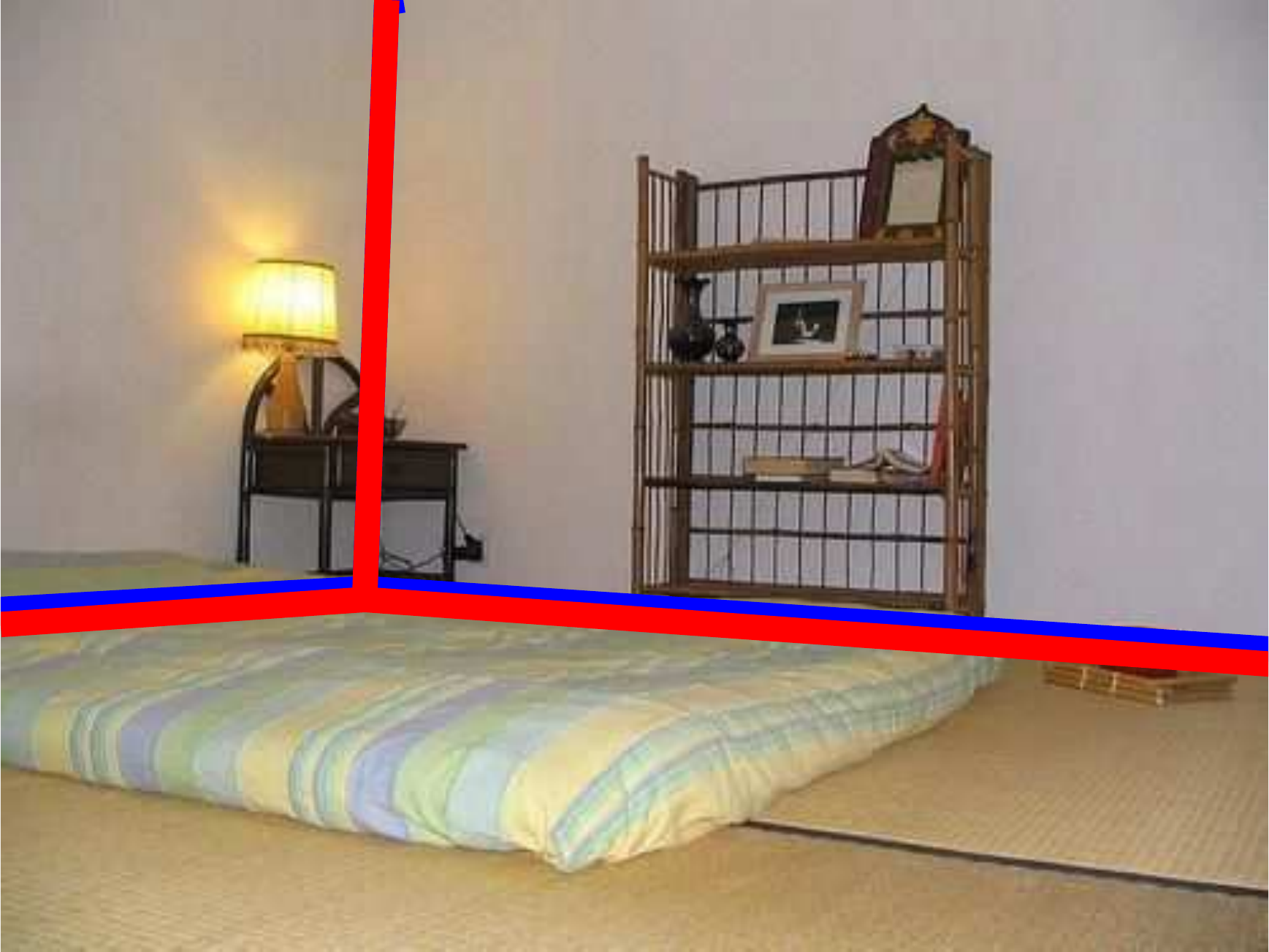}
		\includegraphics[width=2cm]{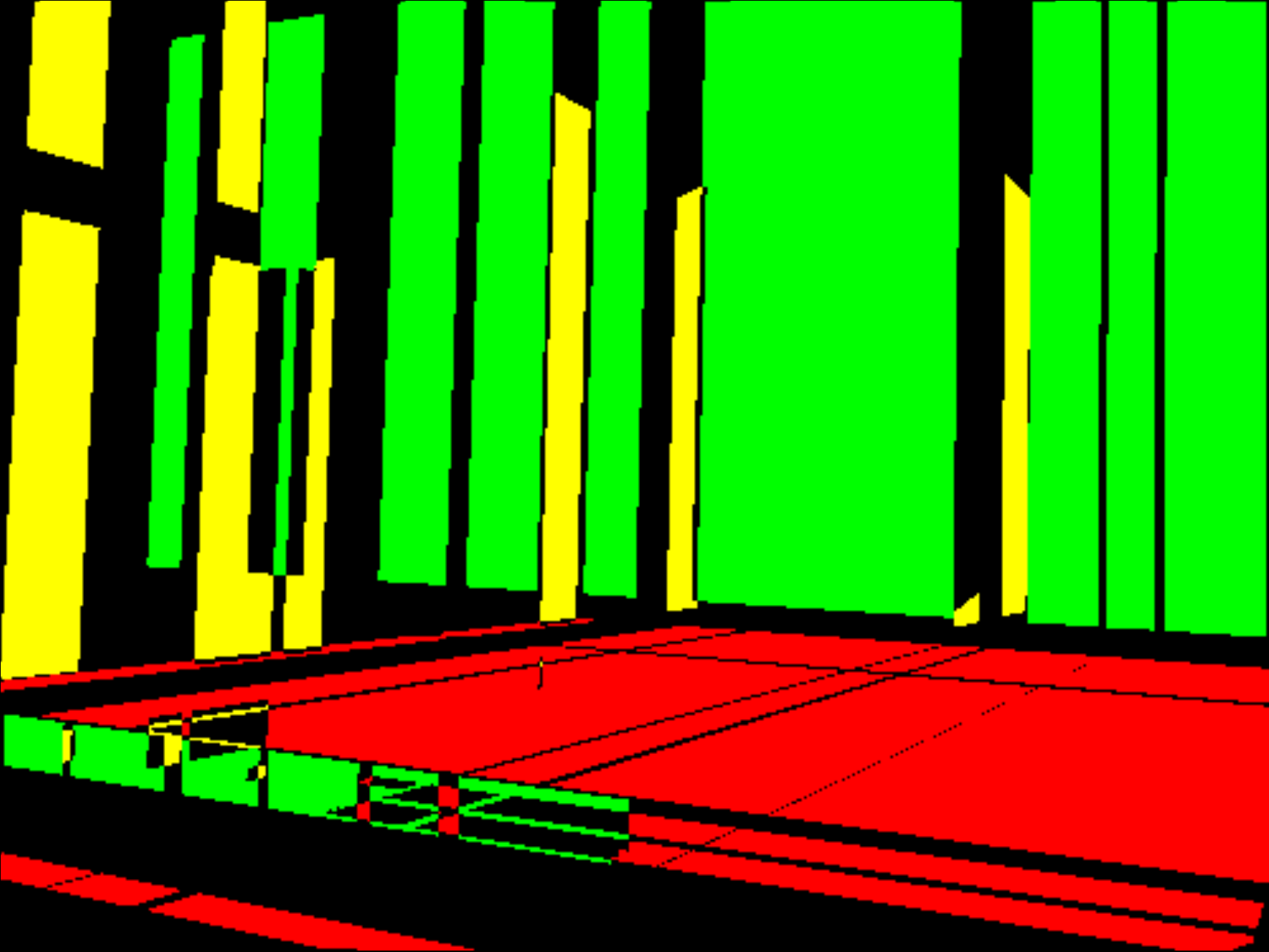}
		\includegraphics[width=2cm]{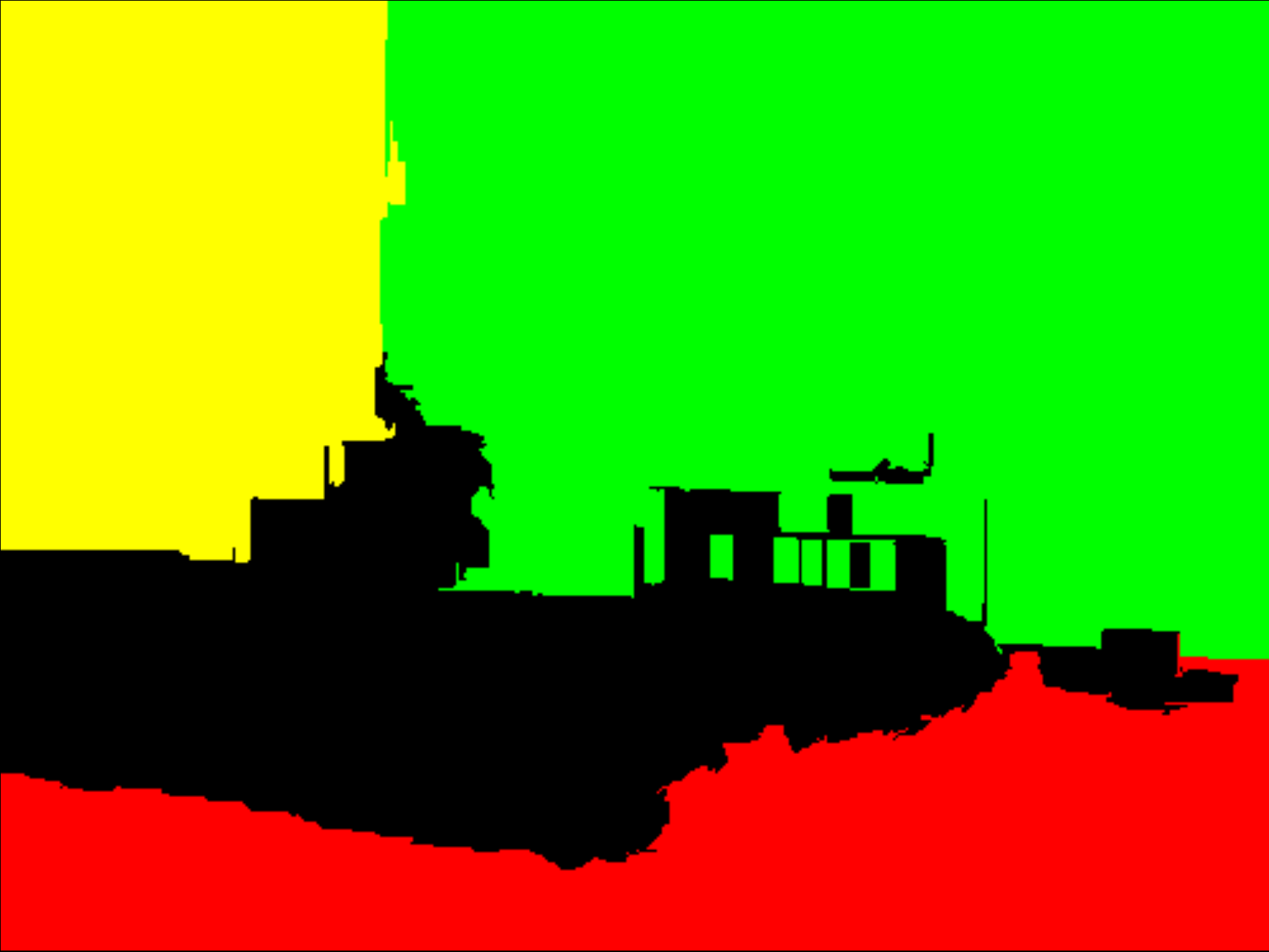}
		\label{fig:Qualfirst}
	}
	\subfigure[Error: 1.48\%] {
		\includegraphics[width=2cm]{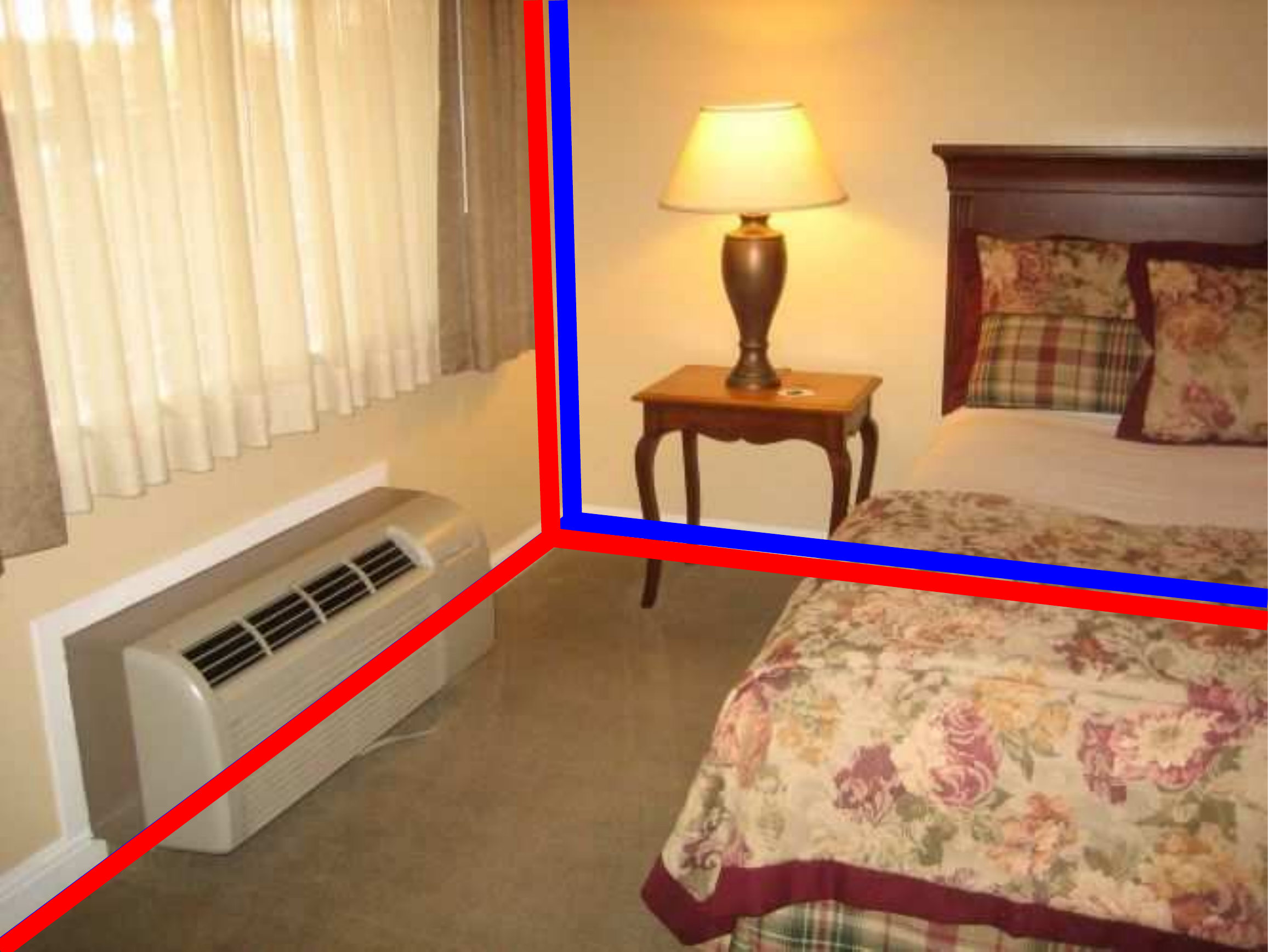}
		\includegraphics[width=2cm]{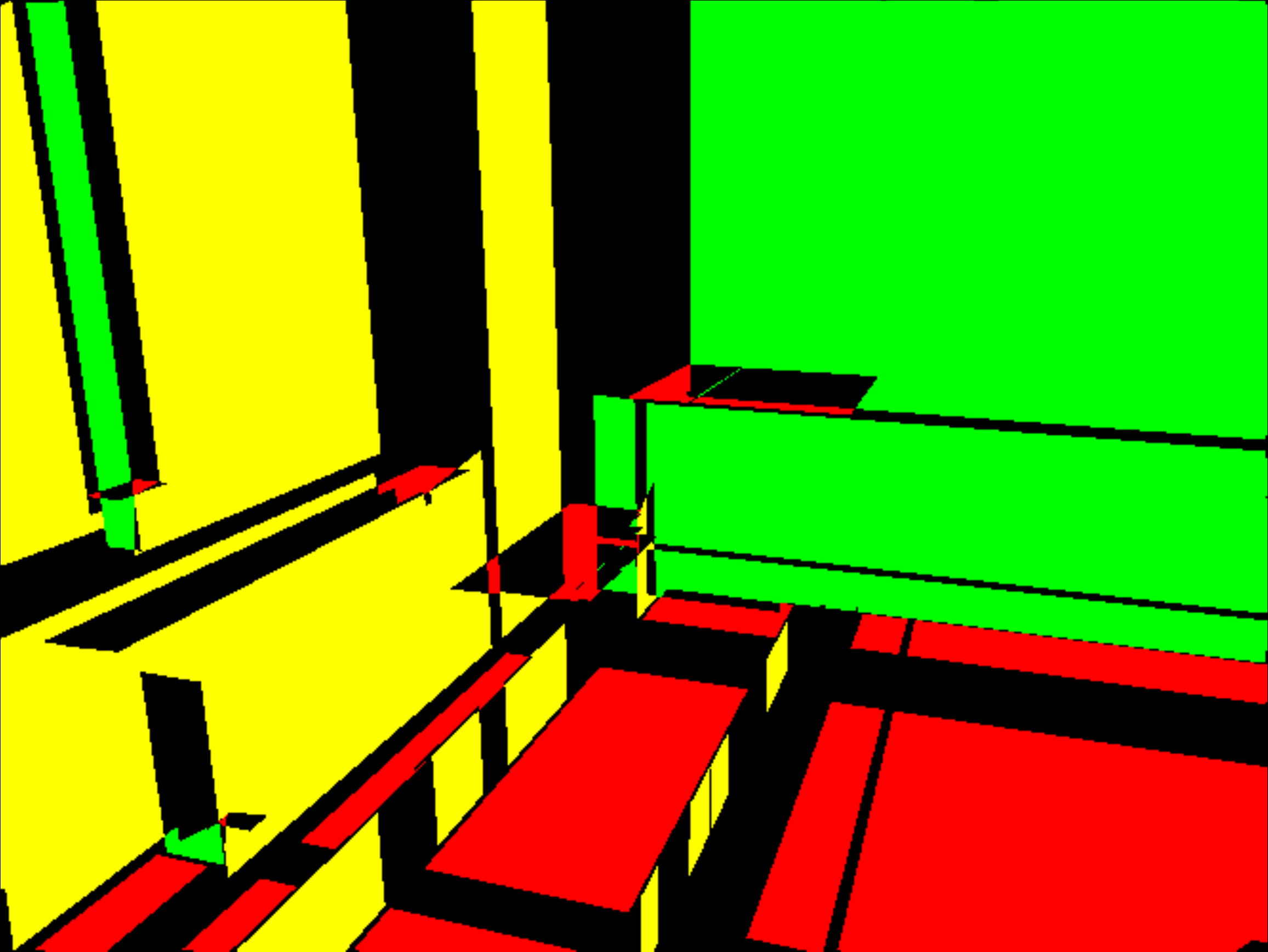}
		\includegraphics[width=2cm]{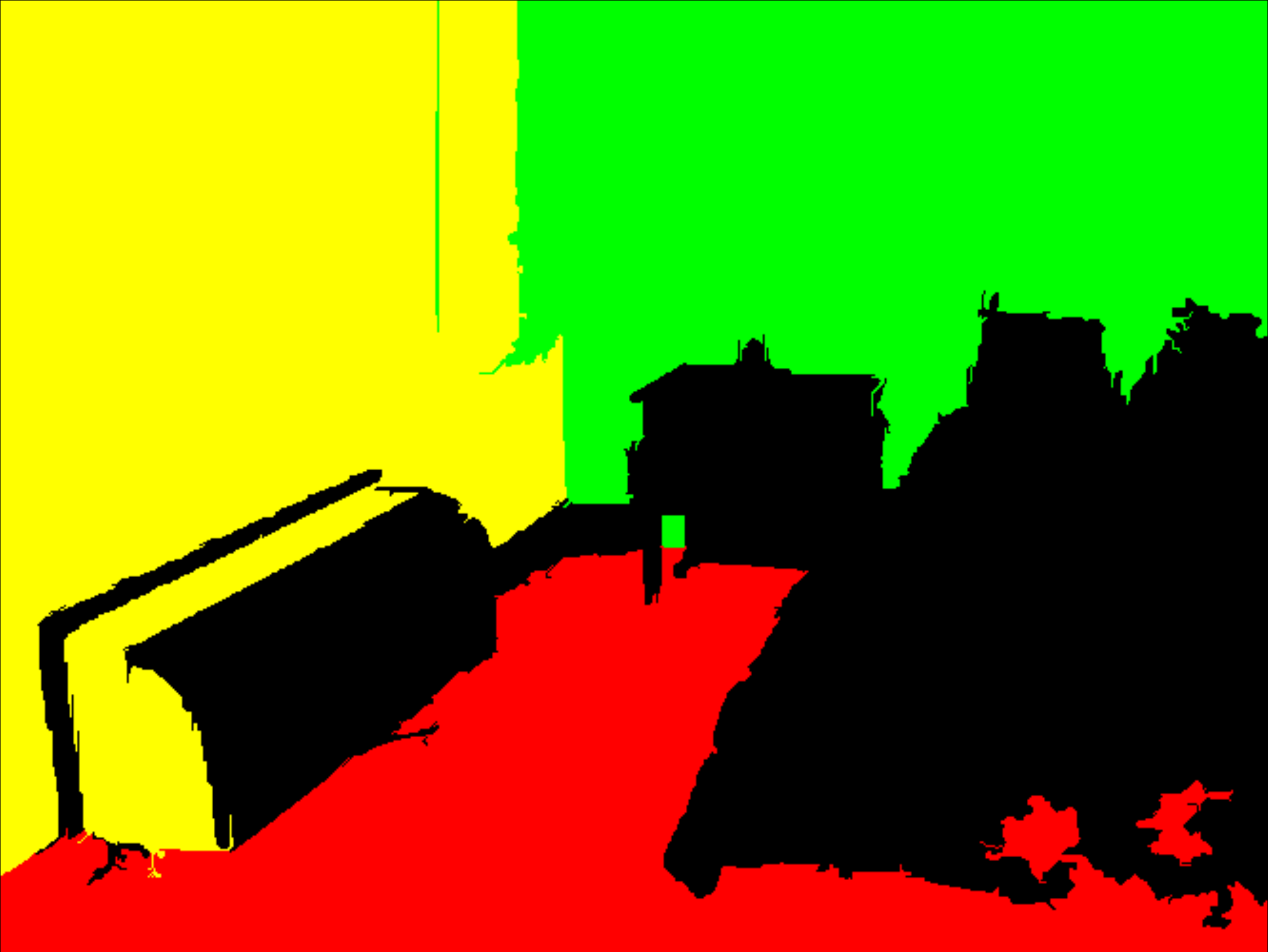}
	}\vspace{-0.2cm}
	\\
	\subfigure[Error: 1.50\%] {
		\includegraphics[width=2cm]{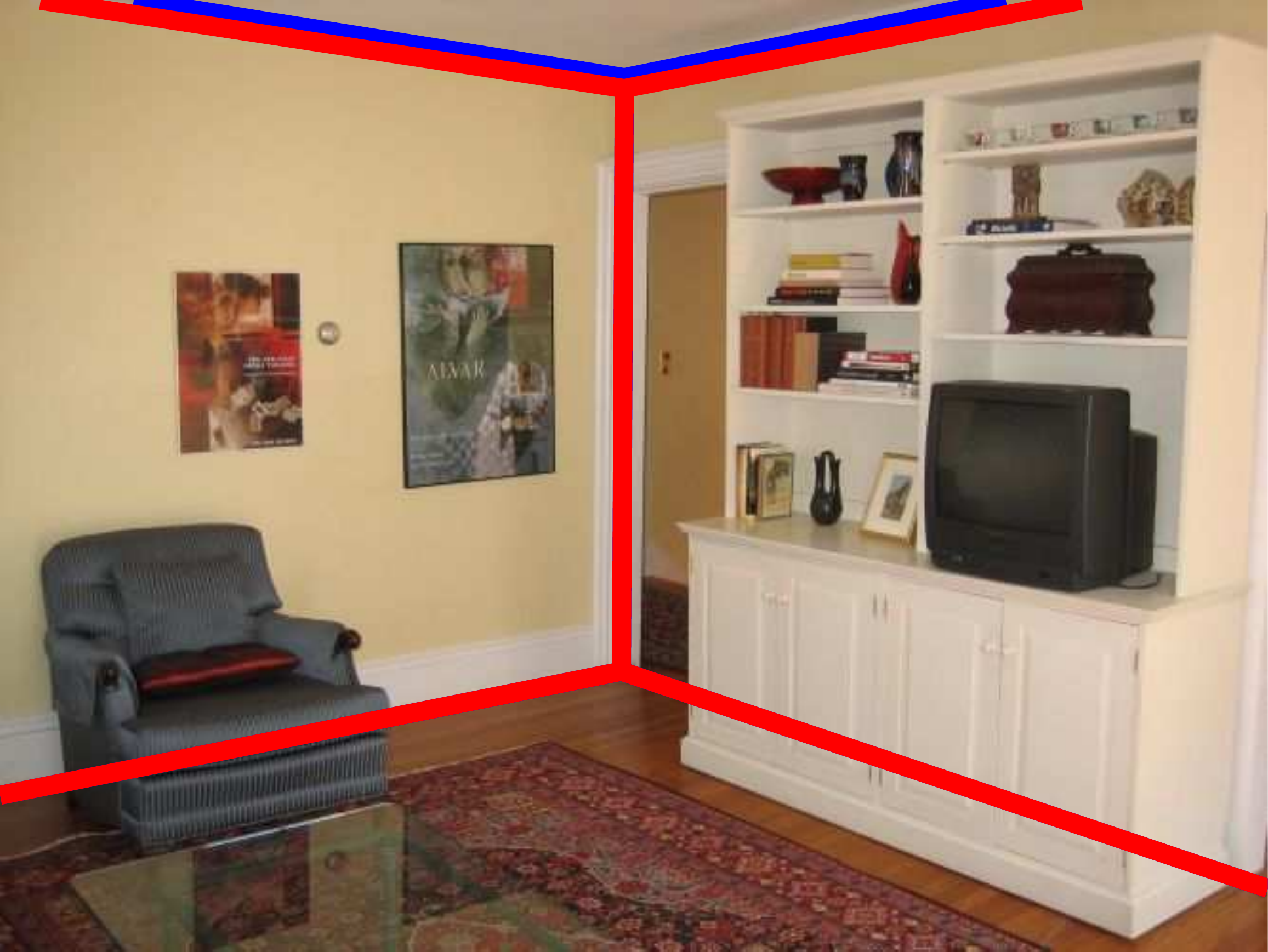}
		\includegraphics[width=2cm]{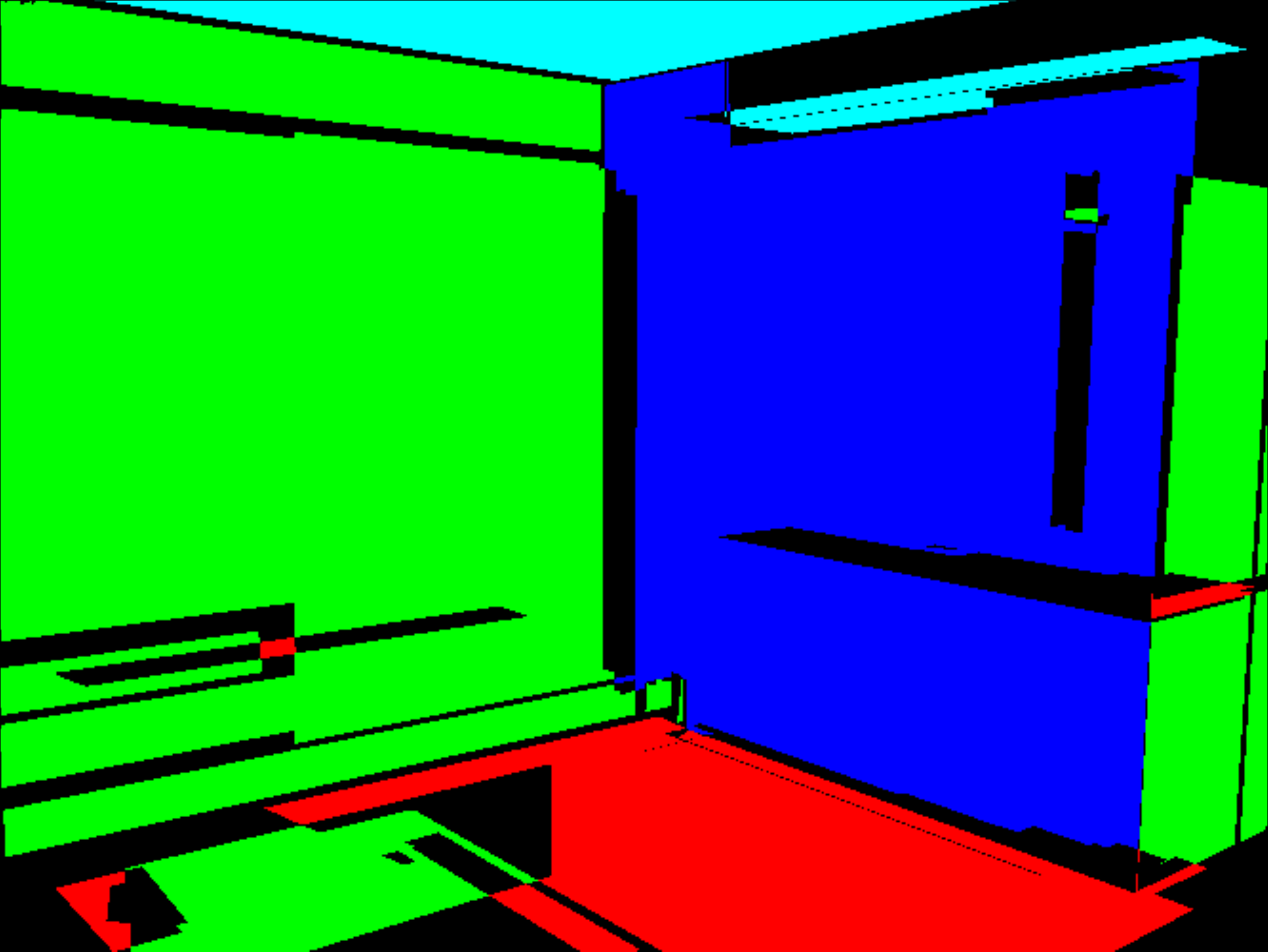}
		\includegraphics[width=2cm]{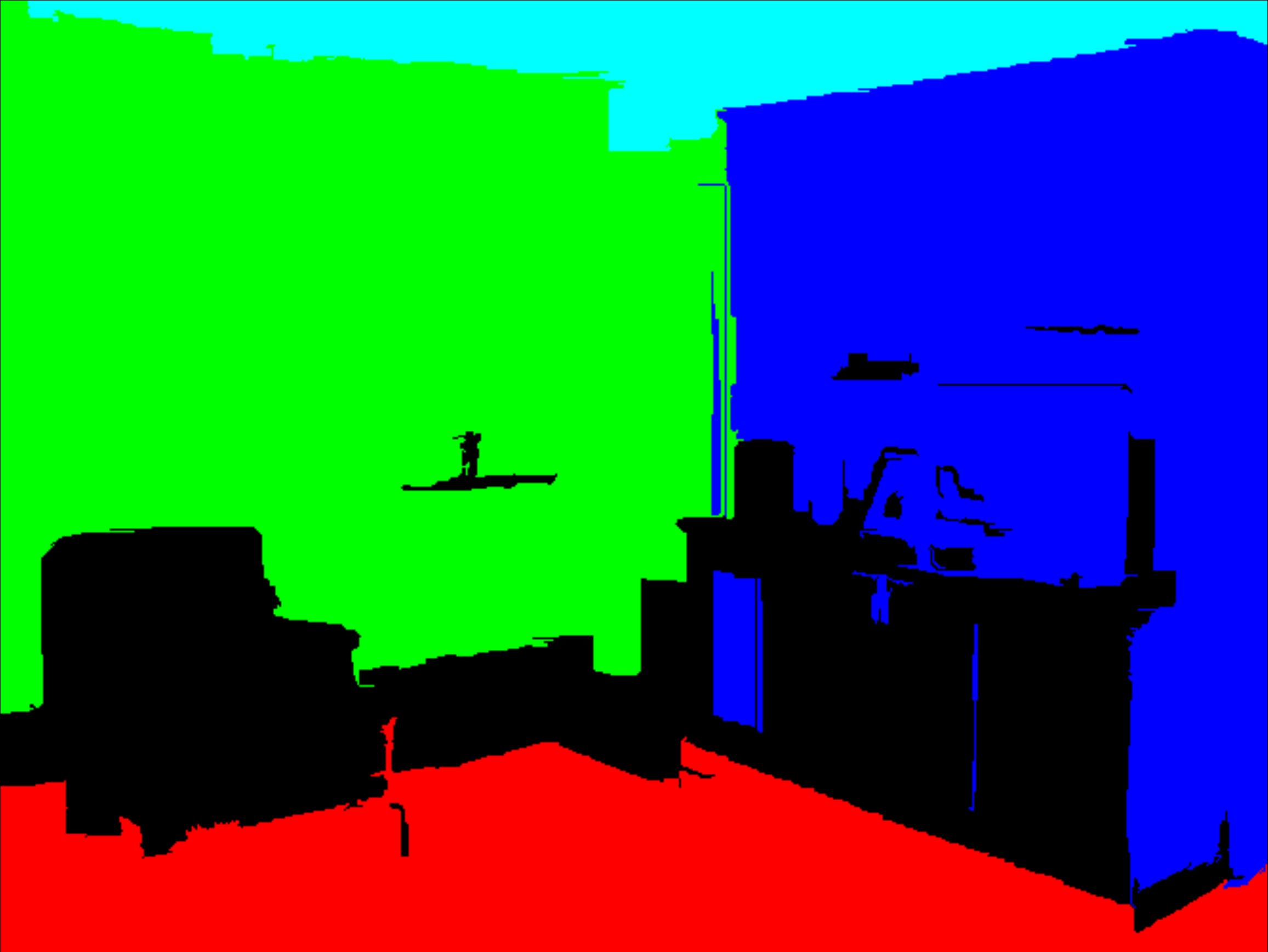}
	}
	\subfigure[Error: 1.60\%] {
		\includegraphics[width=2cm]{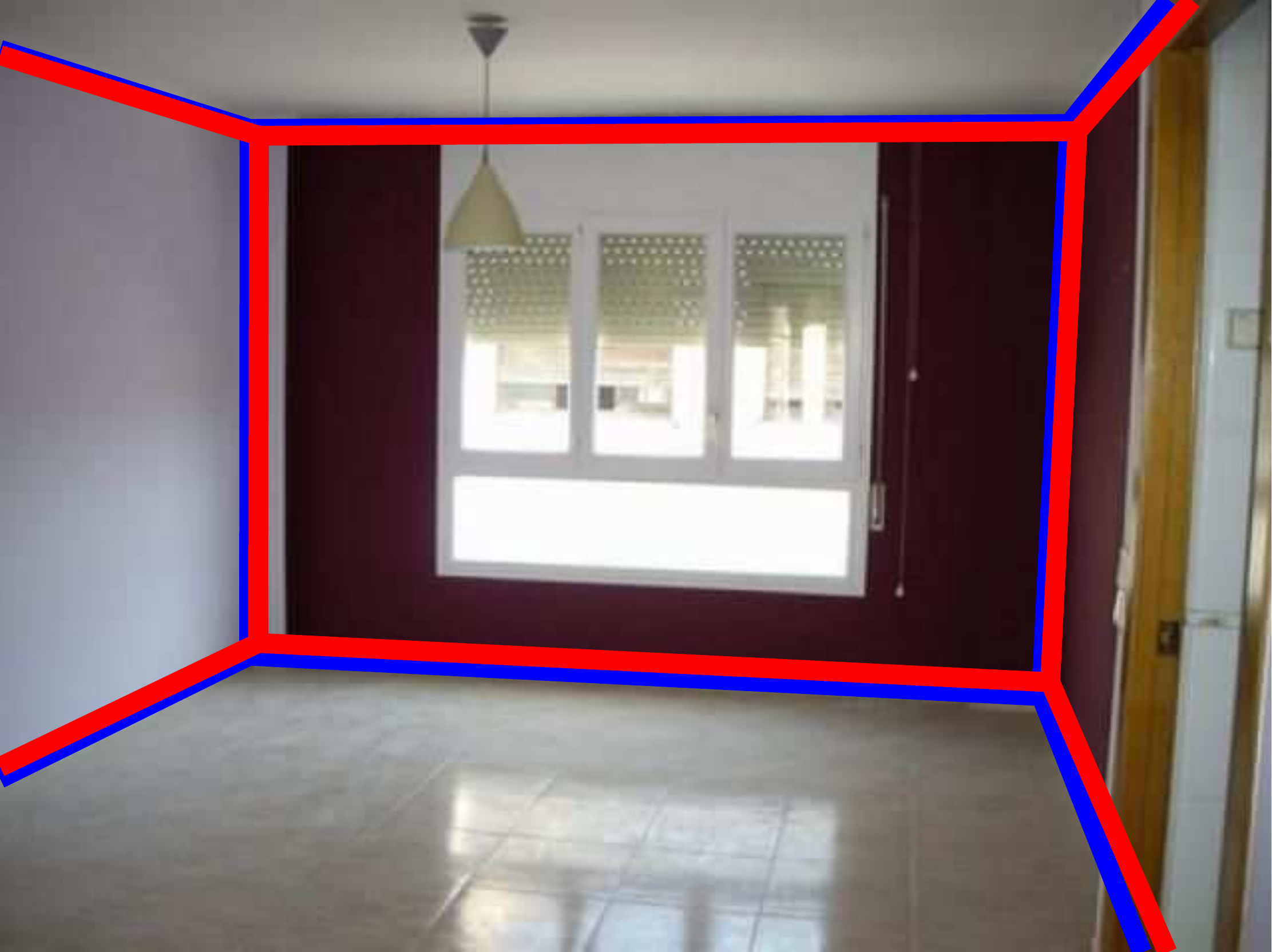}
		\includegraphics[width=2cm]{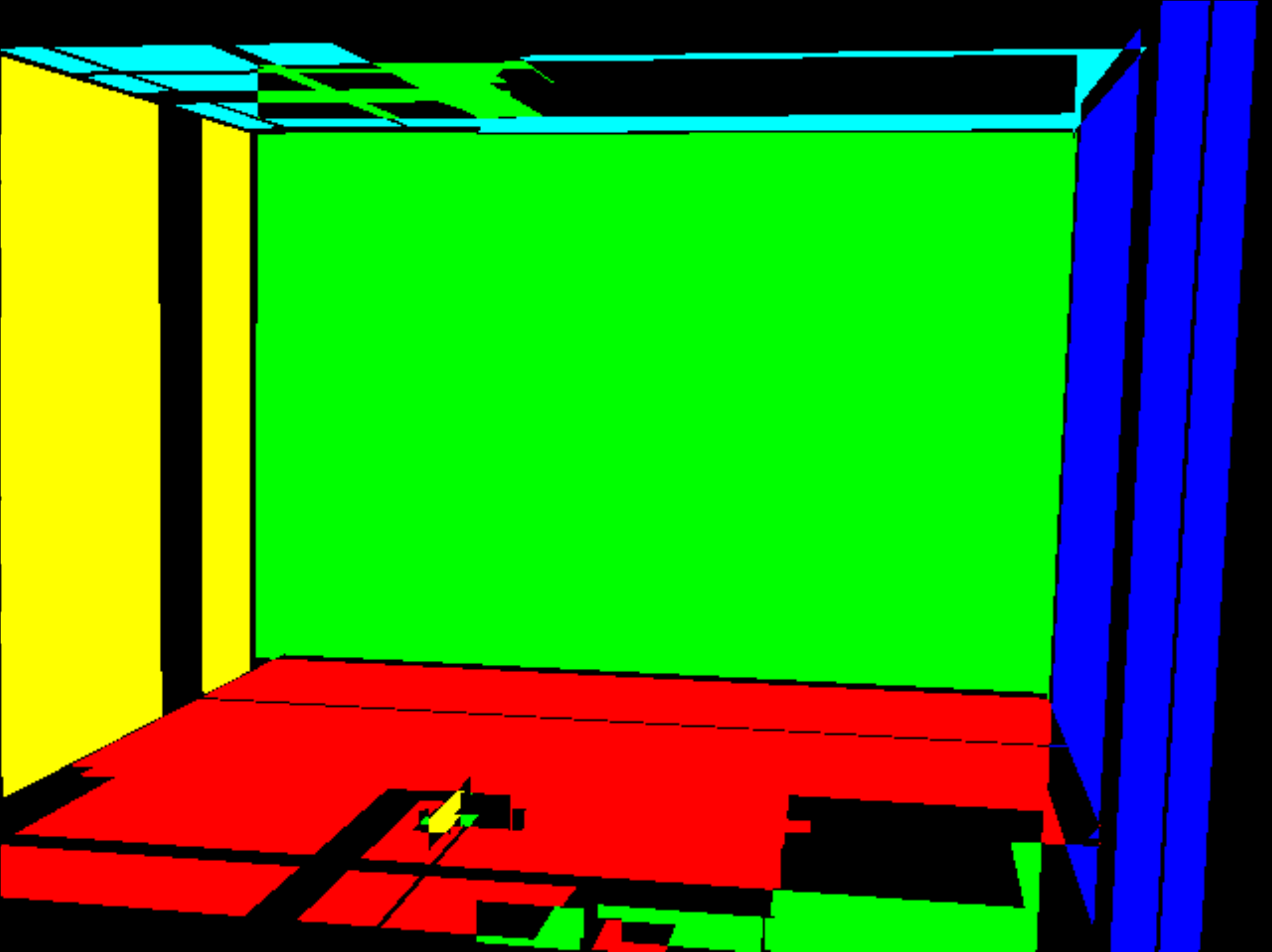}
		\includegraphics[width=2cm]{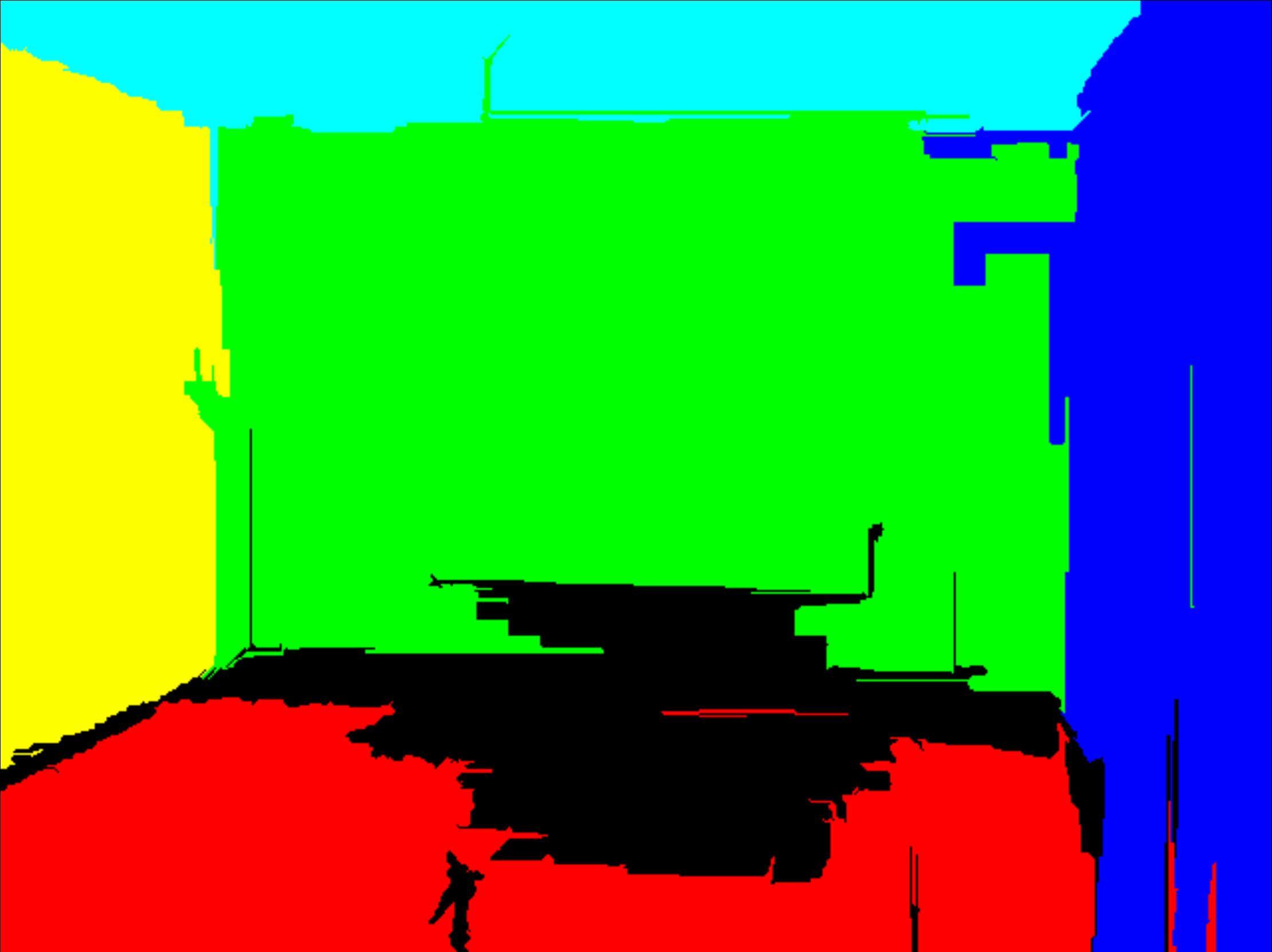}
		\label{fig:QualLastBedroom}
	}\vspace{-0.2cm}
	\subfigure[Error: 48.56\%] {
		\includegraphics[width=2cm]{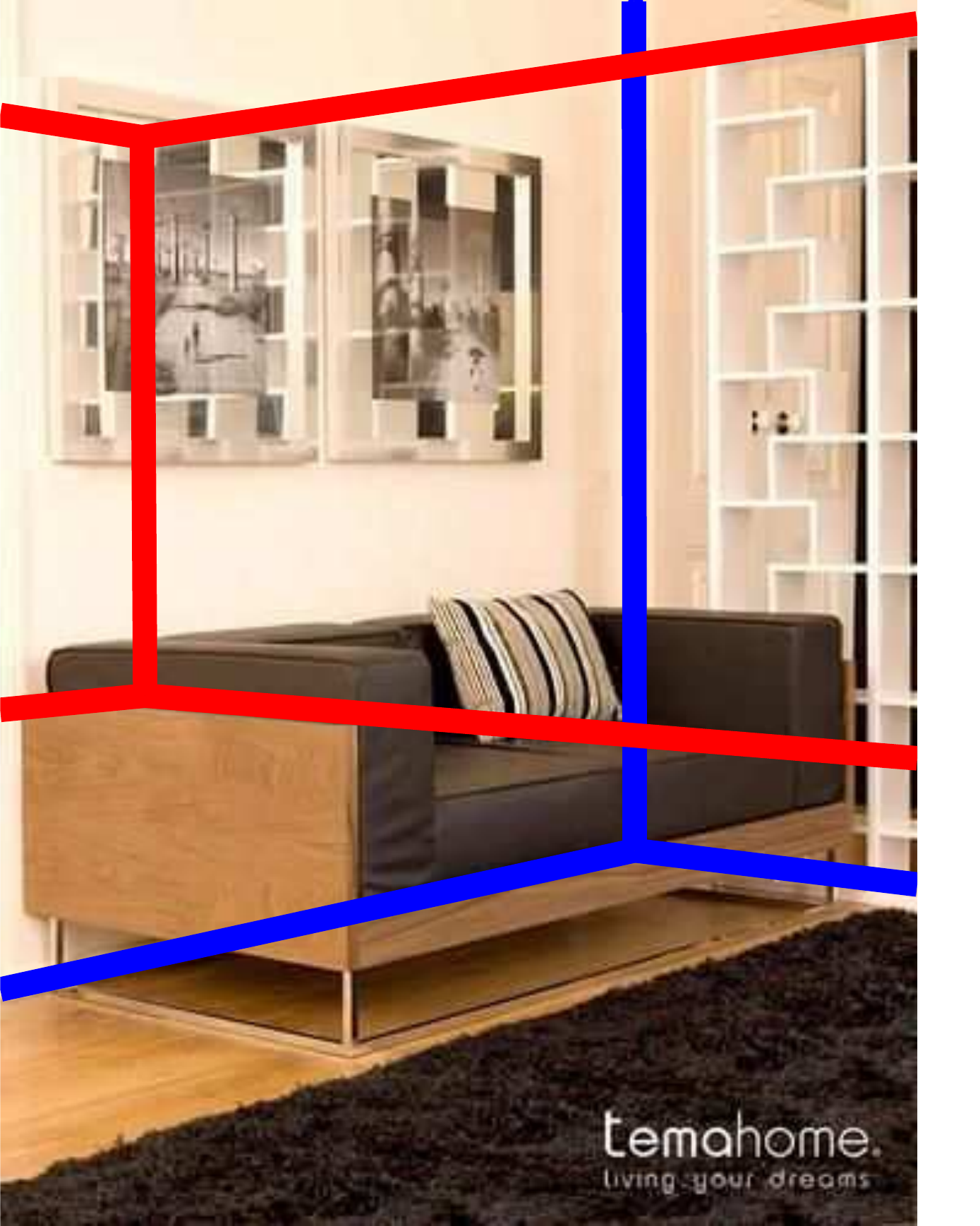}
		\includegraphics[width=2cm]{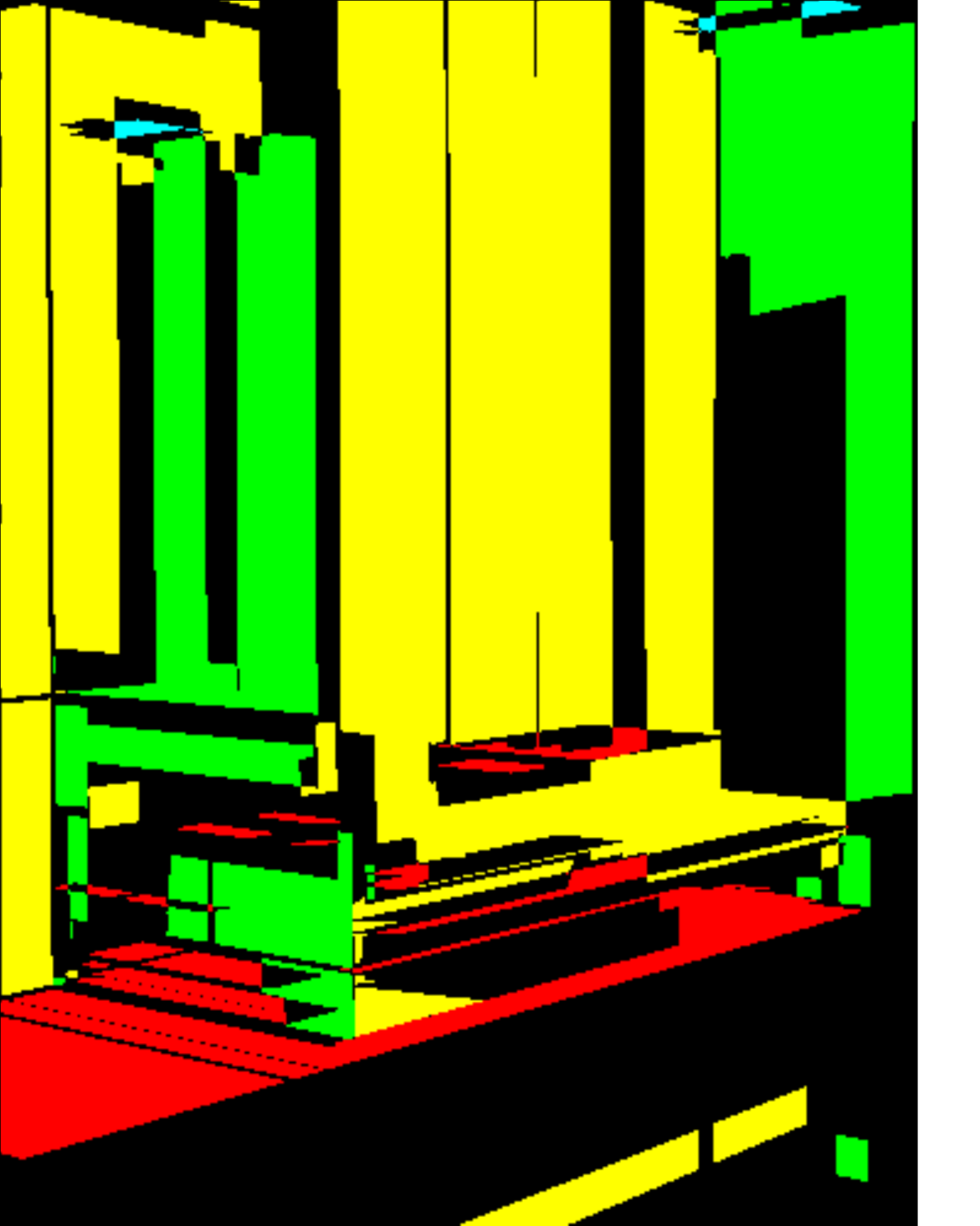}
		\includegraphics[width=2cm]{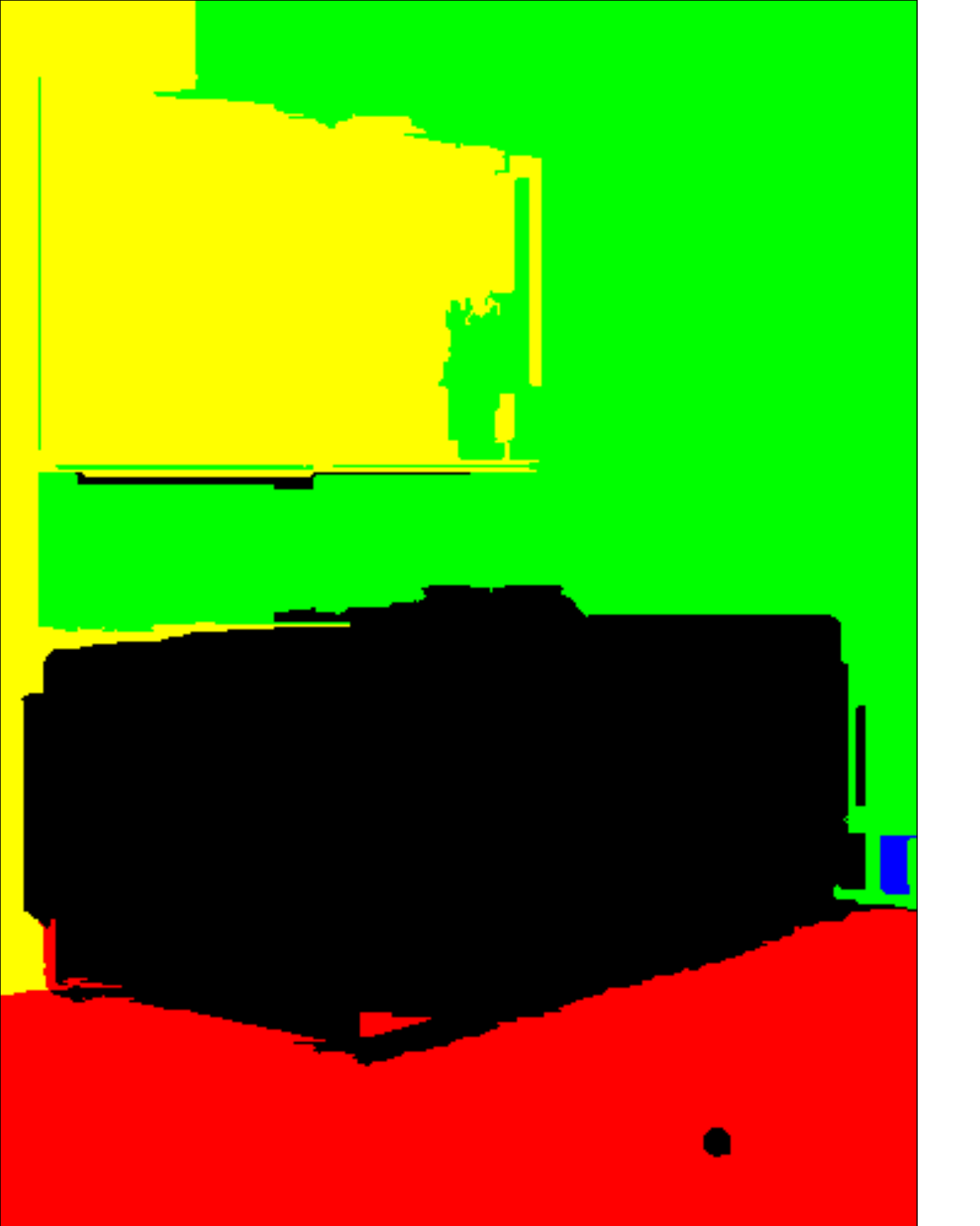}
		\label{fig:QualfirstFail}
	}
	\subfigure[Error: 42.11\%] {
		\includegraphics[width=2cm]{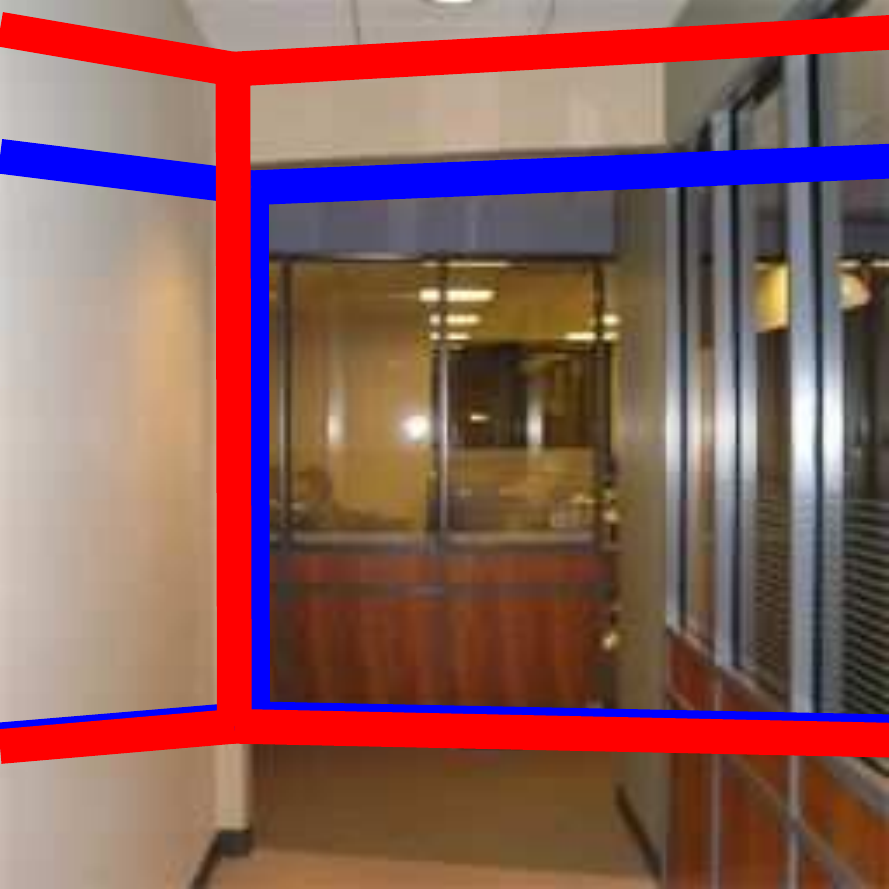}
		\includegraphics[width=2cm]{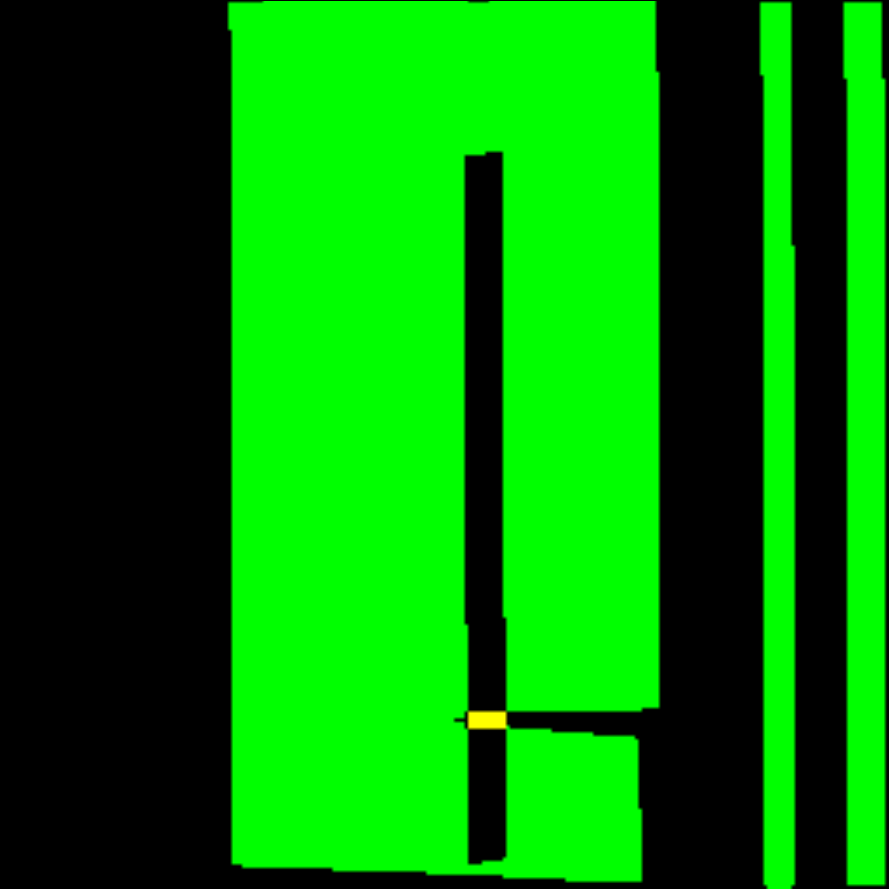}
		\includegraphics[width=2cm]{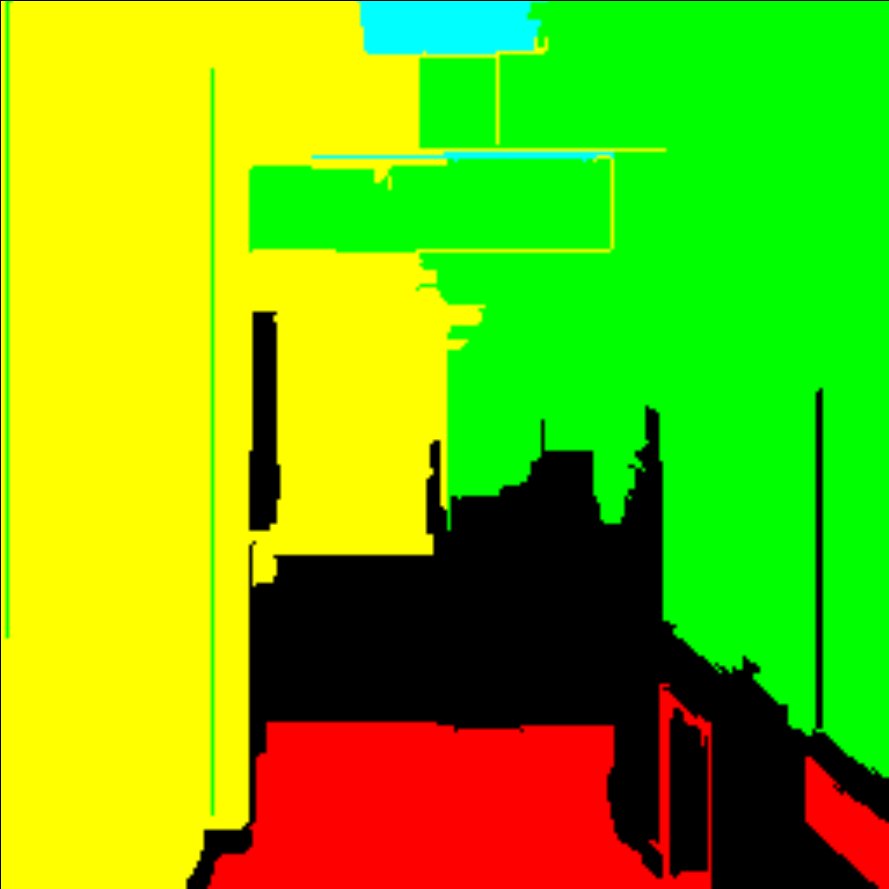}
		\label{fig:QuallastFail}
	}
\caption{Original image with estimated layout in red, (OM) and (GC) features. \subref{fig:Qualfirst}-\subref{fig:QualLastBedroom}: Examples of successful cases. \subref{fig:QualfirstFail}-\subref{fig:QuallastFail}: Failure modes.}
\label{fig:Visualization}
\vspace{-0.2cm}
\end{figure*}

 We evaluated our approach on the data set of~\cite{Hedau09}, which contains 314 images with ground truth annotation of layout faces. We employed the vanishing point detection of~\cite{Hedau09}, which failed in 9 training images and was successful for all test images, 105 in total. 
 We use a pixel based error measure, counting the percentage of pixel that disagree with the provided ground truth labeling.
 We compare our approach to the state-of-the-art .  Similar to~\cite{Lee10}, 
we report results when using different sets of image features, i.e., orientation maps (OM),  geometric context (GC), and both (OM+GC). We denote by~\cite{Hedau09} (a), when the GC features are used to estimate the layout, and by~\cite{Hedau09} (b), when the layout is used to re-estimate the GC features, and these new features are used to improve the layout.
As shown in Fig.  \ref{tab:state_feat} (right), our approach is able to significantly outperform the state-of-the-art in all scenarios, 
with our smallest error rate when using all features being $13.59\%$. 
We improve the state-of-the-art by $3.6\%$ for the OM features, by $5.8\%$ for the GC features and by $5.0\%$ when combining both feature cues.
Importantly inference is very efficient and takes on average $0.15$ seconds per image. 
Fig. \ref{fig:Visualization} depicts some successful examples, as well as failure modes. 
We refer the reader to \cite{Schwing12-cvpr} for more details and results.

\subsection{Shape reconstruction from monocular imaging}

Existing approaches to tackling monocular non-rigid surface reconstruction can be classified into (i) non-rigid structure-from-motion techniques ~\cite{Bregler00,Xiao05,Fayad10} that exploit the availability of multiple images of different deformations to reconstruct both 3D points and camera motion, and (ii) template-based methods \cite{Shen09,Perriollat10,Brunet10} that rely on a reference image with known 3D shape to perform reconstruction from a single additional image of the deformed surface. In most cases, the aforementioned methods are specifically designed to handle feature point correspondences, and as a consequence, cannot make use of richer image information, such as full surface texture, or surface boundaries. More importantly, these methods become unsuitable when too few feature points can be reliably detected and matched. 
Several attempts have been proposed to leverage more complex image likelihoods~\cite{Salzmann08a,Salzmann10a}. However, the resulting methods rely on gradient-based optimization schemes that can easily get trapped in the many local maxima of these complex, non-smooth likelihoods. As a consequence, these methods have only been used either for frame-to-frame tracking, where the previous frame provides a good initialization~\cite{Salzmann08a}, or when large amounts of training data are available to learn a discriminative predictor that produces a good initialization~\cite{Salzmann10a}.

In recent work, \citep{Salzmann12}, we have proposed to frame the problem as the one of inference in a graphical model. As this optimization is more global than gradient-based methods, it is also more robust to local maxima, thus yielding accurate reconstructions even in the absence of a good initialization. More specifically, we represent a surface as a triangulated mesh, and define the random variables in the graphical model to be the rotations and translations of the individual mesh facets. To handle such continuous variables, we adopt particle convex belief propagation~\cite{Peng11} as our inference algorithm: We iteratively draw random samples around the current solution for each variable, compute the MAP estimate of the discrete graphical model defined by these samples using convex belief propagation~\cite{Hazan10}, and update the current solution with this MAP estimate. This strategy lets us effectively explore the 3D shape space even when no good initialization is provided. 
We define potentials that encode surface boundary, facets coherence as well as template matching. We refer the reader to \cite{Salzmann12} for more details and results. 
We employ our approach to find the weights of the individual terms in the likelihood. 
To speed-up inference, we first define the graphical model over a coarse mesh and perform gradient descent on a finer mesh with initial point the MAP of the coarse mesh. 

We compare our results against two baselines. The first one, later denoted by Shen09, corresponds to~\cite{Shen09} initialized with the reference shape, with the extension of~\cite{Salzmann10a} to allow for more general image likelihoods than feature point reprojection error. The second baseline, later denoted by Salz10, follows the method of~\cite{Salzmann10a} and uses a Gaussian process (GP) predictor to initialize the shape before gradient-based optimization. To learn the GP predictor, we used the same training shapes as to learn the potential weights, and employ either noisy 2D point locations, or PHOG descriptors as input. To confirm that a simple coarse-to-fine optimization scheme is not enough to solve the problem, we also compare our results with a coarse-to-fine version of~\cite{Shen09}, denoted by Shen09 CTF. For all the baselines, we used the same image likelihoods as for our method, together with the weights learned with our CRF formulation.

\begin{figure}[t!]
\begin{center}
\hspace{-0.6cm}\includegraphics[width=0.37\linewidth]{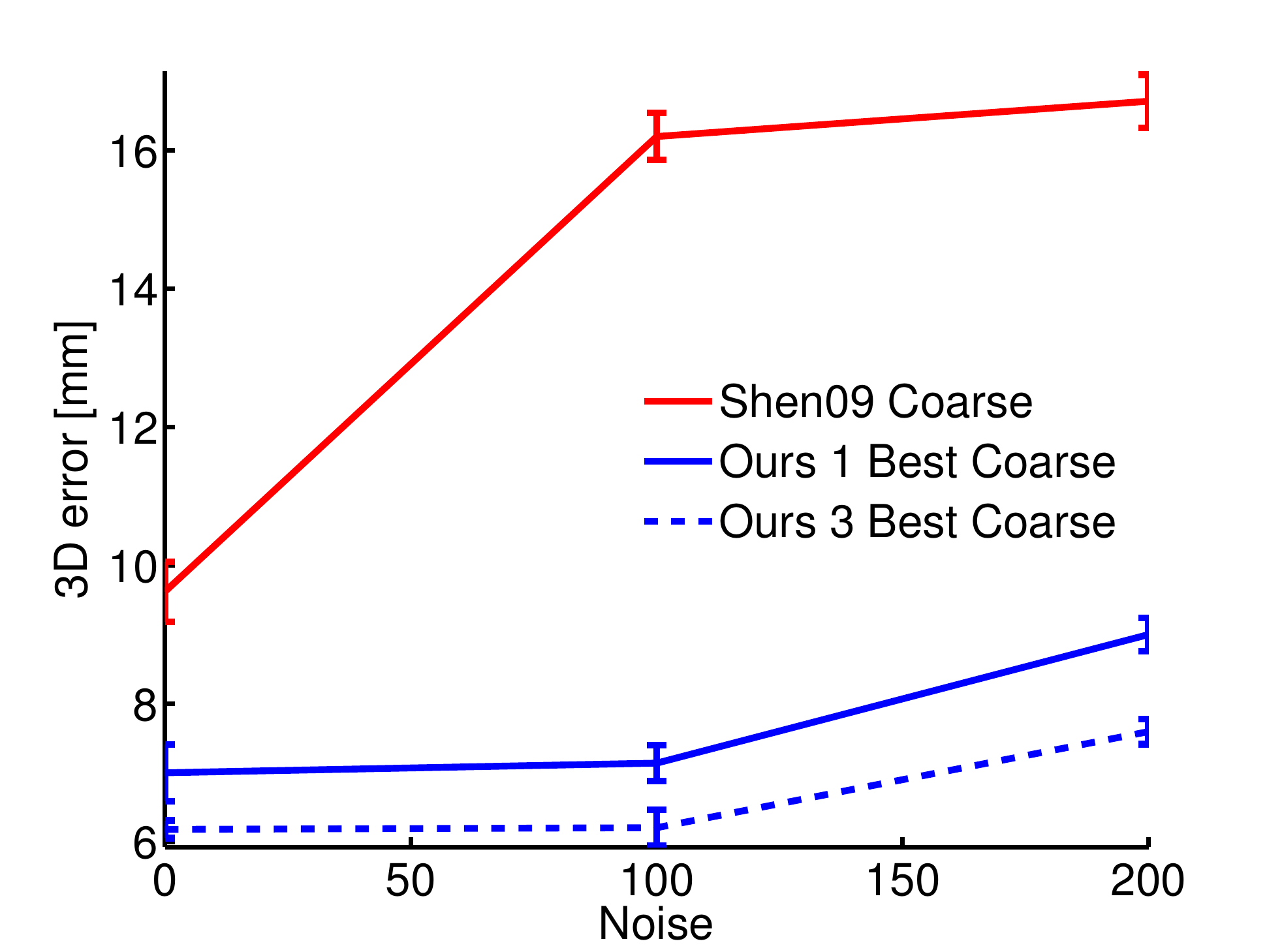} 
\hspace{-0.3cm}\includegraphics[width=0.37\linewidth]{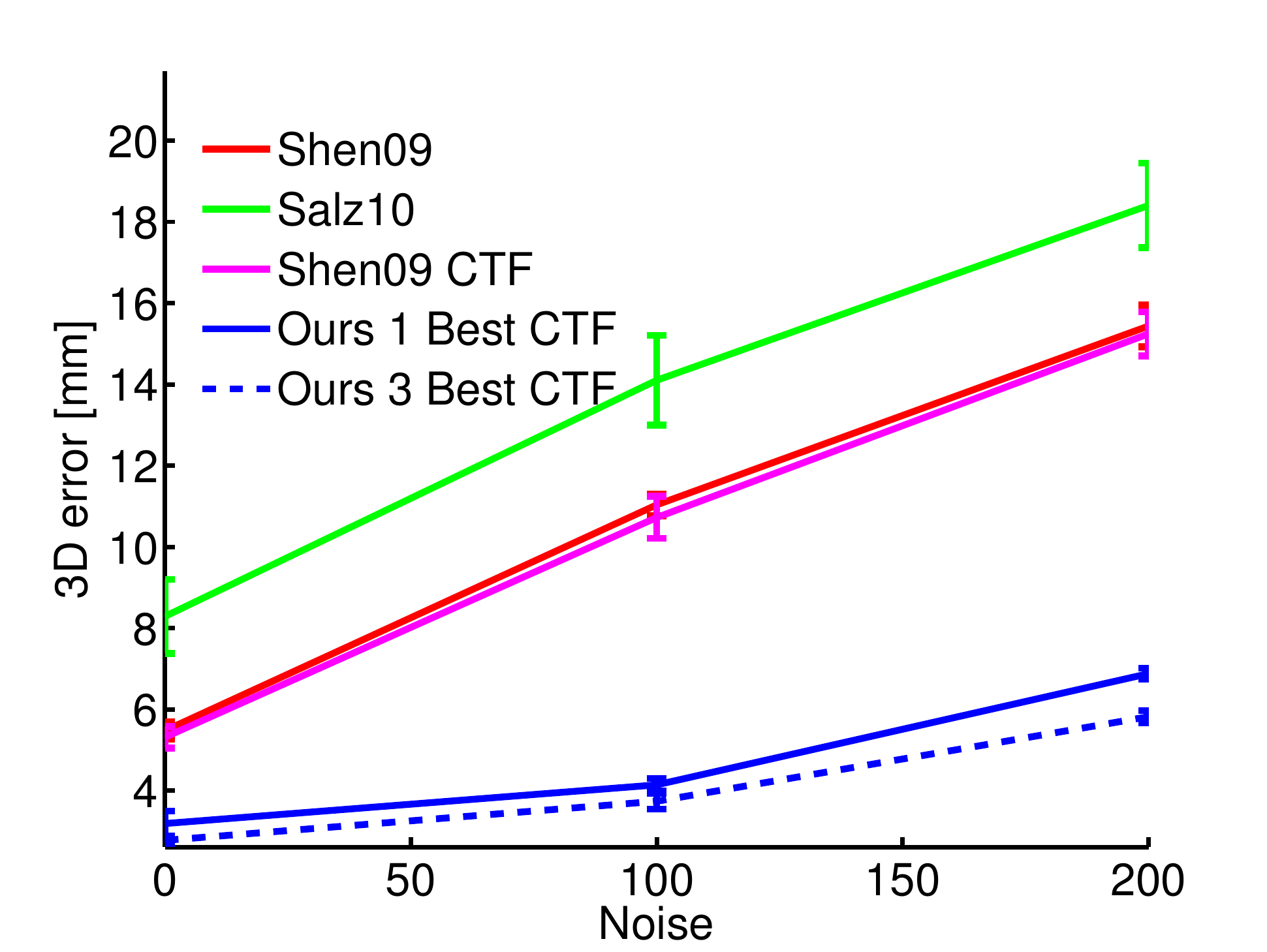}  
\end{center}
\vspace{-0.65cm}
\caption{{\bf Reconstructing a piece of cardboard from well-textured images.} 3D error when (a) using a coarse ($3 \times 3$) mesh and no smoothness, and (b) refining the results of (a) with a gradient-based method. 
Shen09 and Salz10 were directly obtained using a fine mesh. 
Note that our coarse results give a much better initialization for the refinement step.}
\label{fig:cardboard_dali}
\vspace{-0.3cm}
\end{figure}

We perform experiments  using data obtained with a motion capture system~\cite{DeformData}. The data consists of 3D reconstructions of reflective markers placed in a $9\times 9$ regular grid of $160\times 160$mm on a piece of cardboard deformed in front of 6 infrared cameras.  Since no images are provided with the 3D data, we synthesized well- and poorly-textured images as before.  We used 5 training examples to learn the potential weights. 
Fig.~\ref{fig:cardboard_dali}  depict the 3D errors 
with a coarse mesh and after refinement using a gradient descent approach. Our approach yields much more accurate reconstructions than the baselines. 
Interestingly, however, we outperform the baselines after refinement. This shows that our coarse results still provide a better initialization than the coarse version of Shen09. Note that with this poorly-textured surface, smoothness improves reconstruction, which seems natural since image information is much weaker. This, however, is not noticeably the case for the baselines.

\begin{figure}[t!]
\begin{center}
\begin{tabular}{cccccc}
\hspace{-0.2cm}\includegraphics[width=0.17\linewidth]{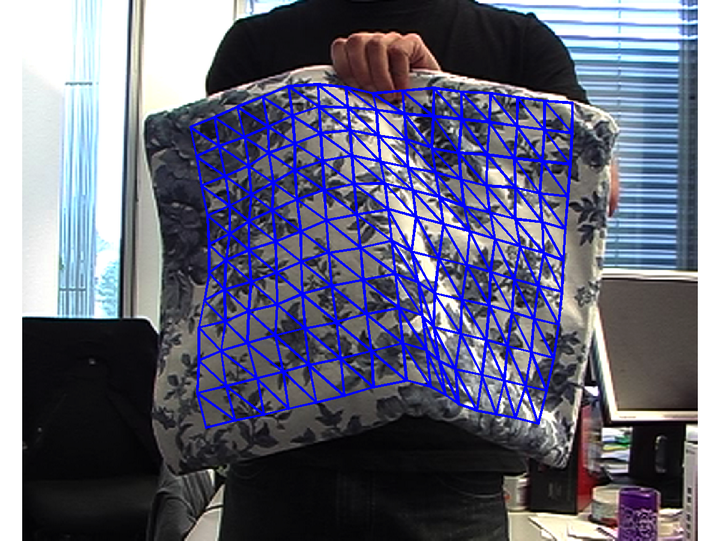} &
\hspace{-0.1cm}\includegraphics[width=0.17\linewidth]{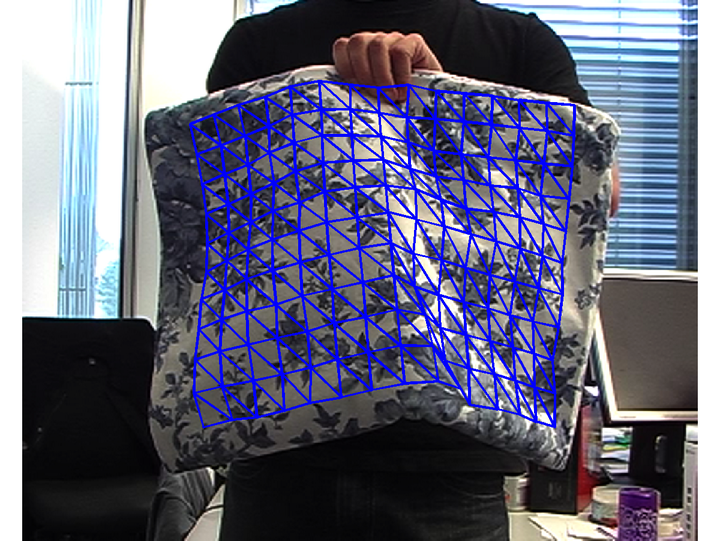}  &
\hspace{-0.1cm}\includegraphics[width=0.17\linewidth]{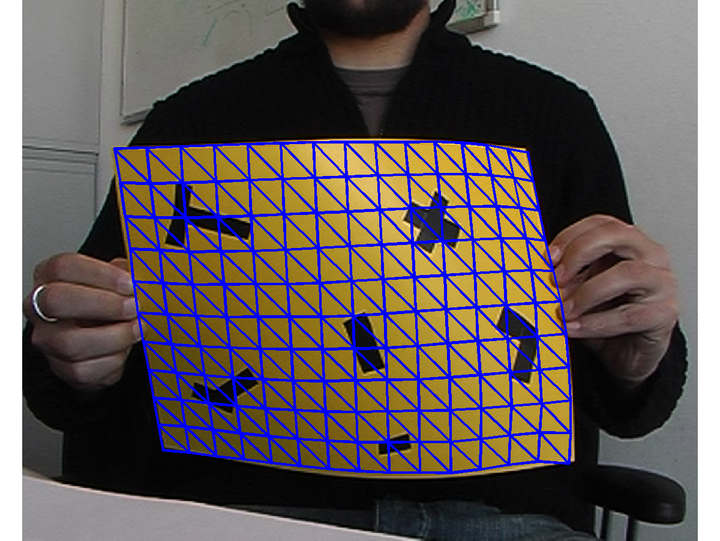}  &
\hspace{-0.1cm}\includegraphics[width=0.17\linewidth]{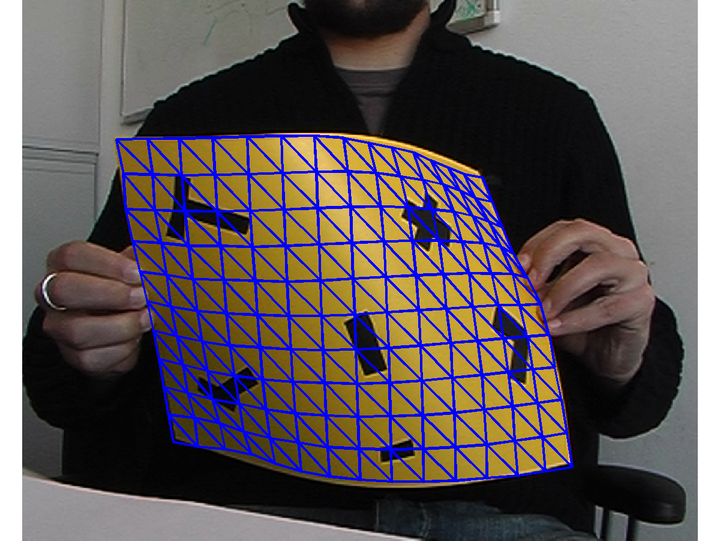} &
\hspace{-0.1cm}\includegraphics[width=0.17\linewidth]{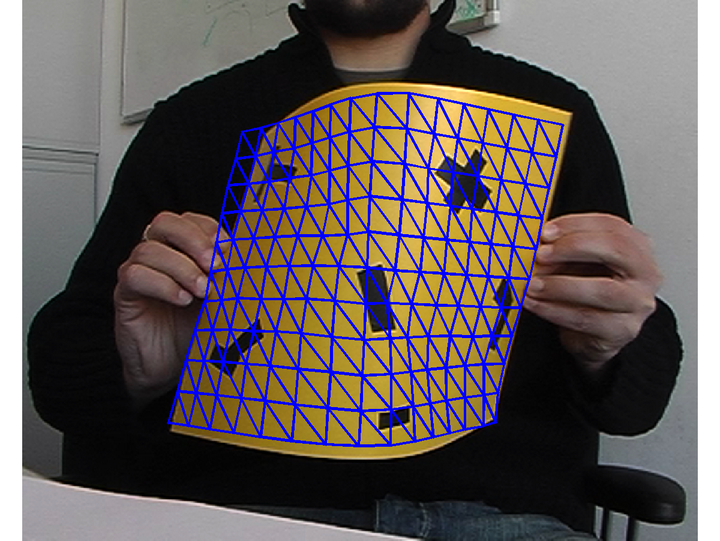}  &
\hspace{-0.1cm}\includegraphics[width=0.17\linewidth]{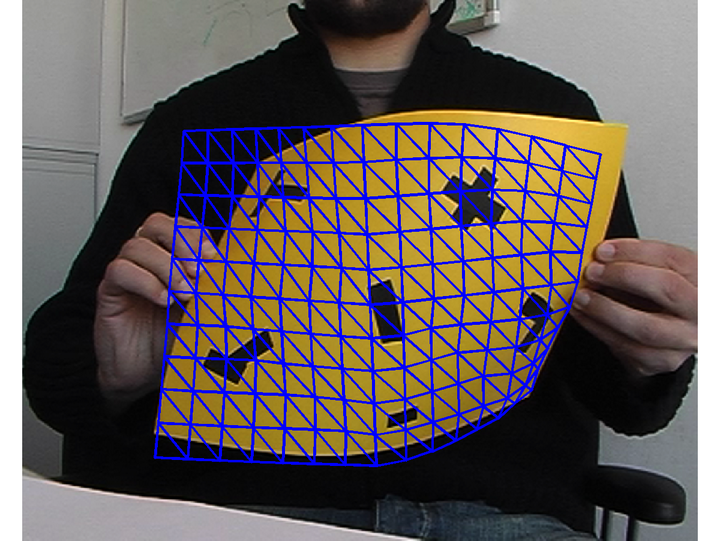}  \vspace{-0.1cm} \\
\hspace{-0.2cm}\includegraphics[width=0.17\linewidth]{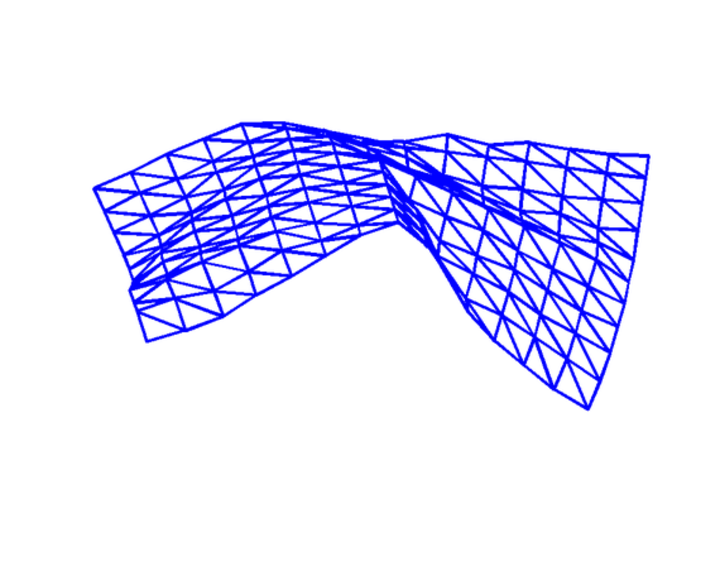} &
\hspace{-0.1cm}\includegraphics[width=0.17\linewidth]{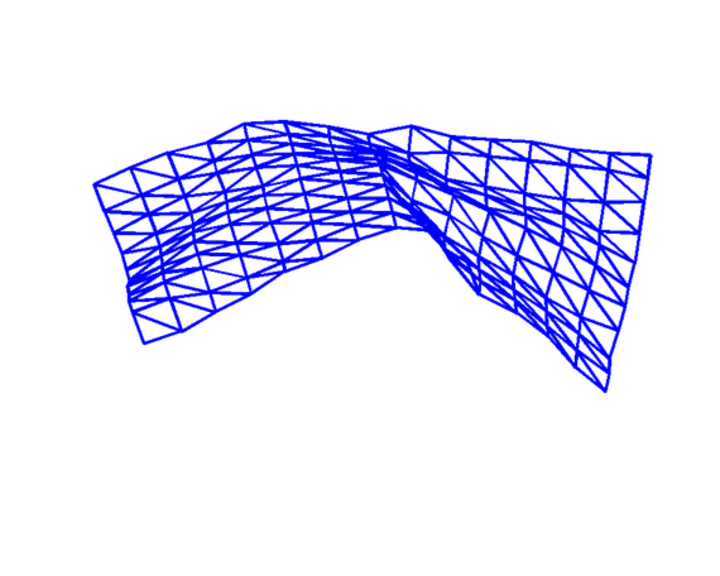}  &
\hspace{-0.1cm}\includegraphics[width=0.17\linewidth]{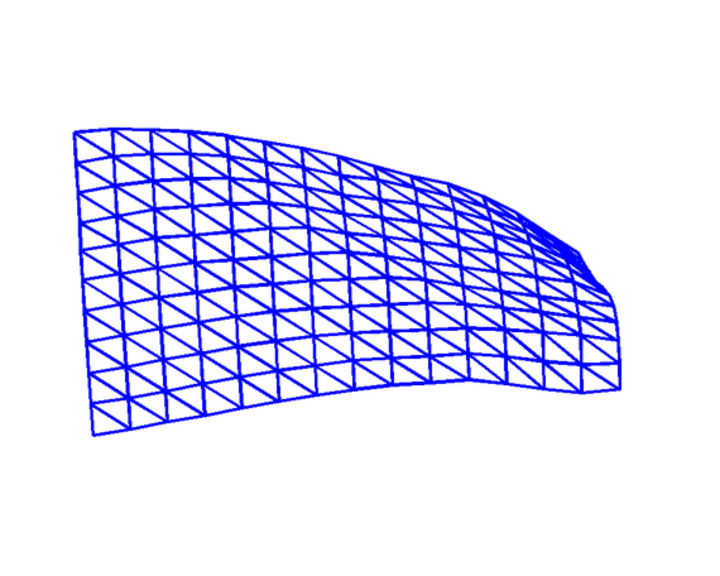}  &
\hspace{-0.1cm}\includegraphics[width=0.17\linewidth]{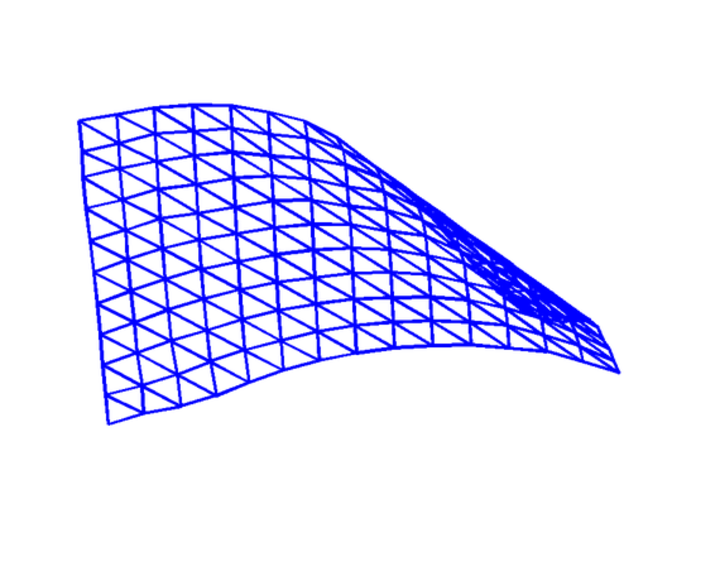} &
\hspace{-0.1cm}\includegraphics[width=0.17\linewidth]{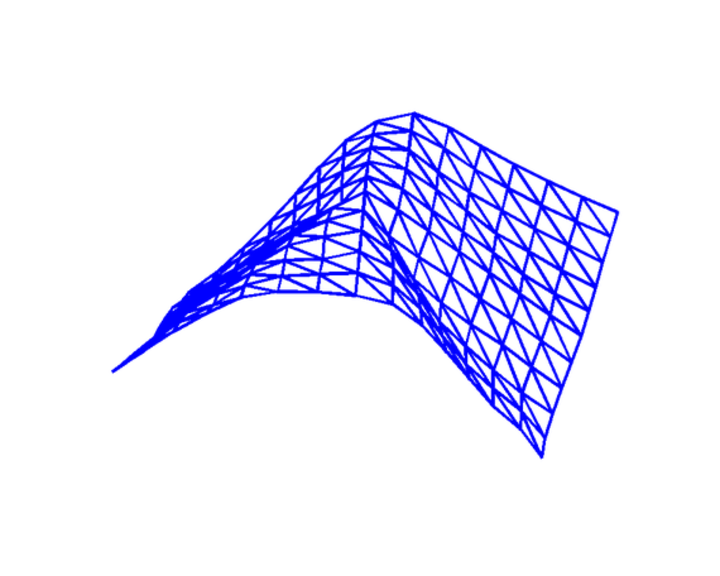}  &
\hspace{-0.1cm}\includegraphics[width=0.17\linewidth]{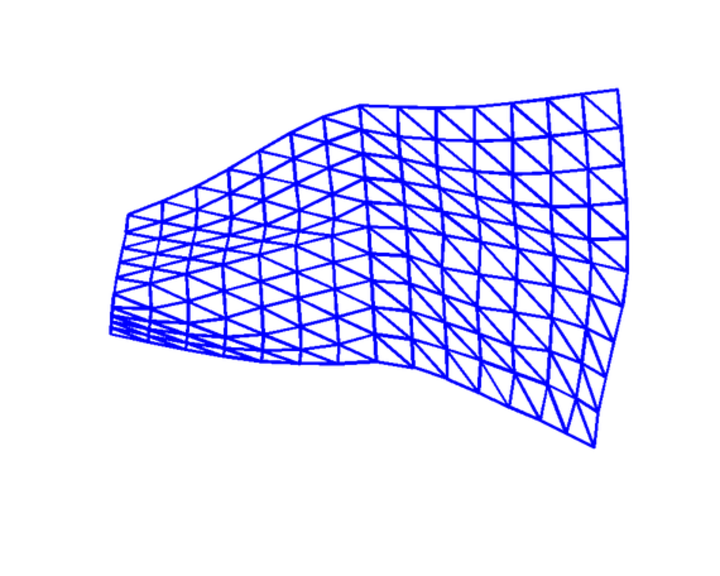}  \vspace{-0.3cm}\\
\hspace{-0.2cm}\includegraphics[width=0.17\linewidth]{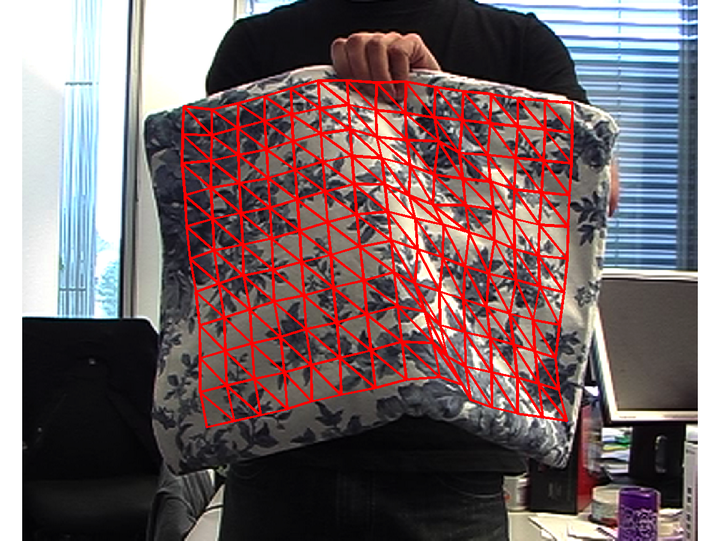} &
\hspace{-0.1cm}\includegraphics[width=0.17\linewidth]{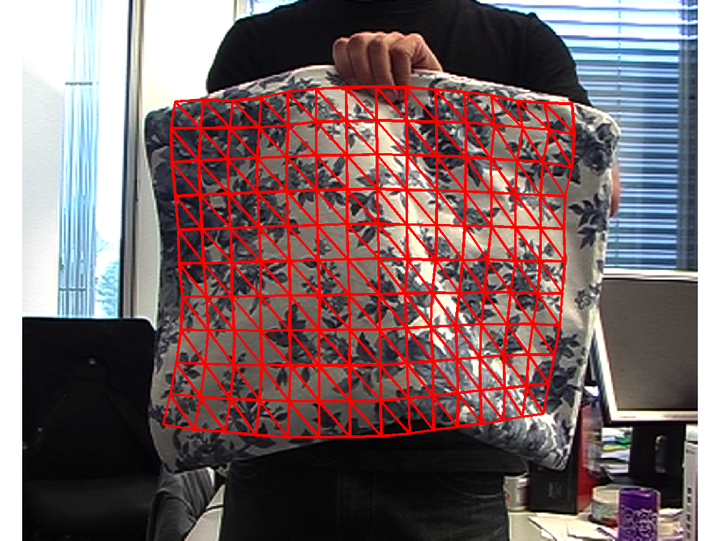}  &
\hspace{-0.1cm}\includegraphics[width=0.17\linewidth]{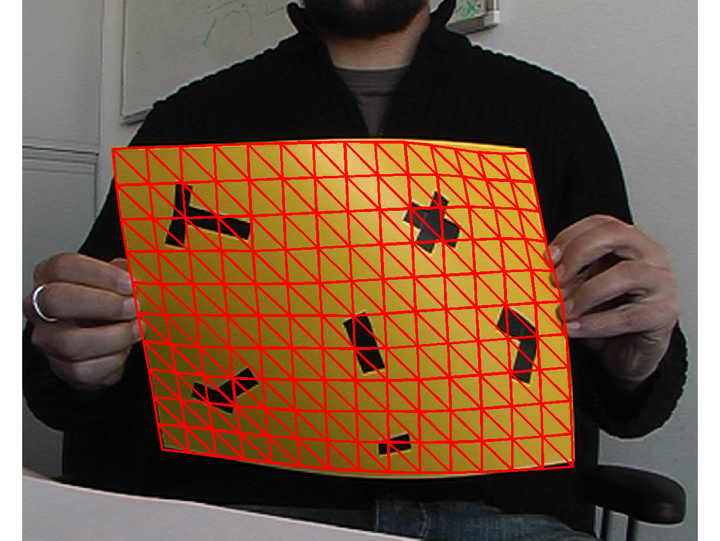}  &
\hspace{-0.1cm}\includegraphics[width=0.17\linewidth]{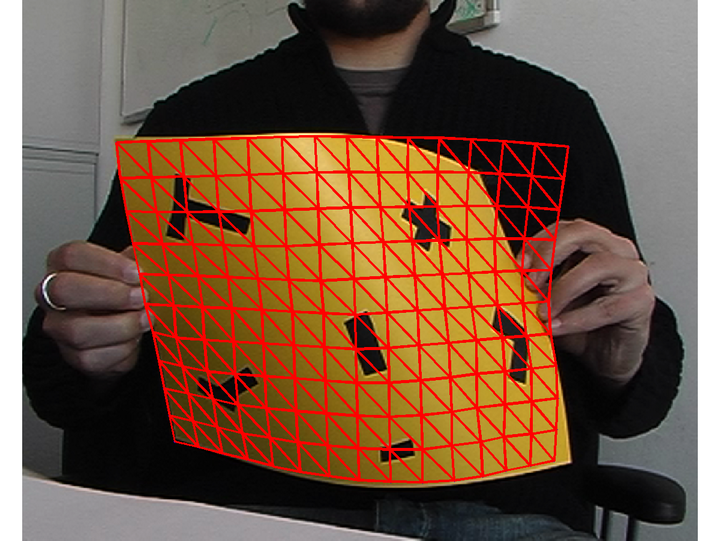} &
\hspace{-0.1cm}\includegraphics[width=0.17\linewidth]{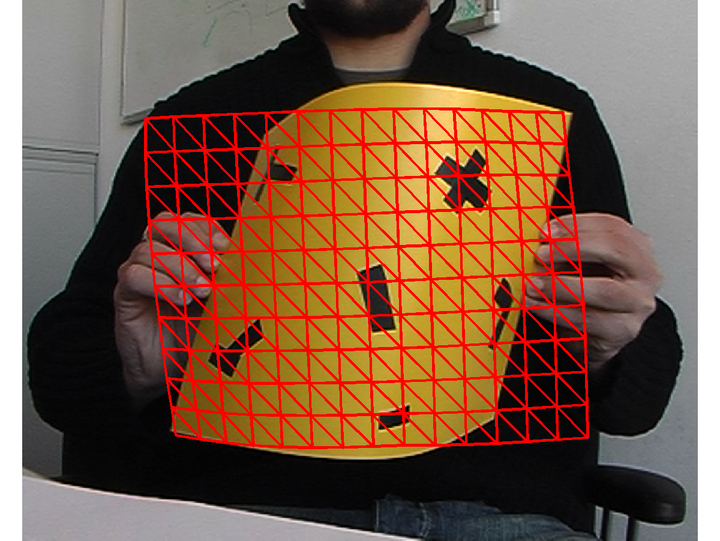}  &
\hspace{-0.1cm}\includegraphics[width=0.17\linewidth]{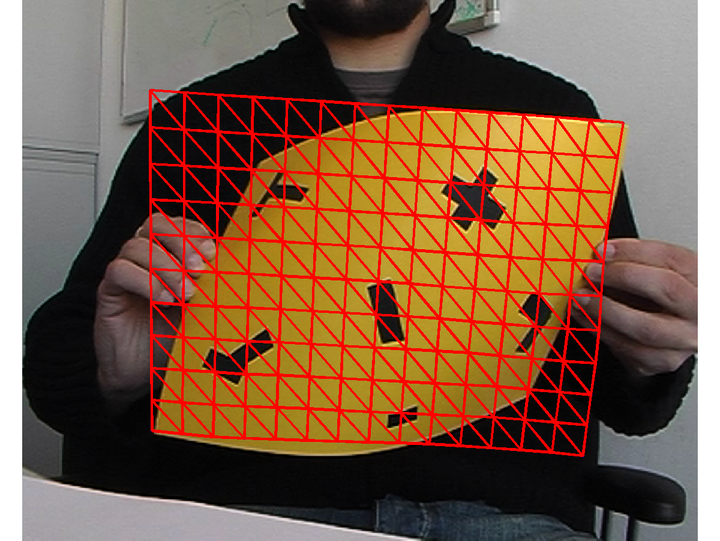}  \vspace{-0.1cm}\\
\hspace{-0.2cm}\includegraphics[width=0.17\linewidth]{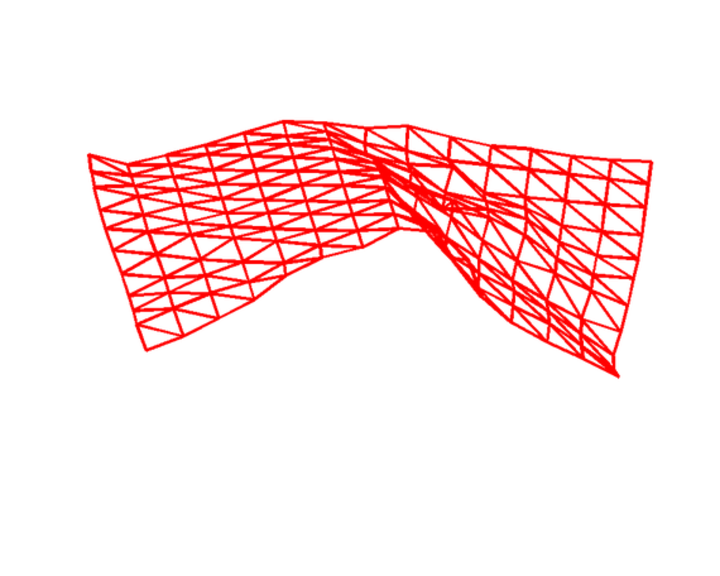} &
\hspace{-0.1cm}\includegraphics[width=0.17\linewidth]{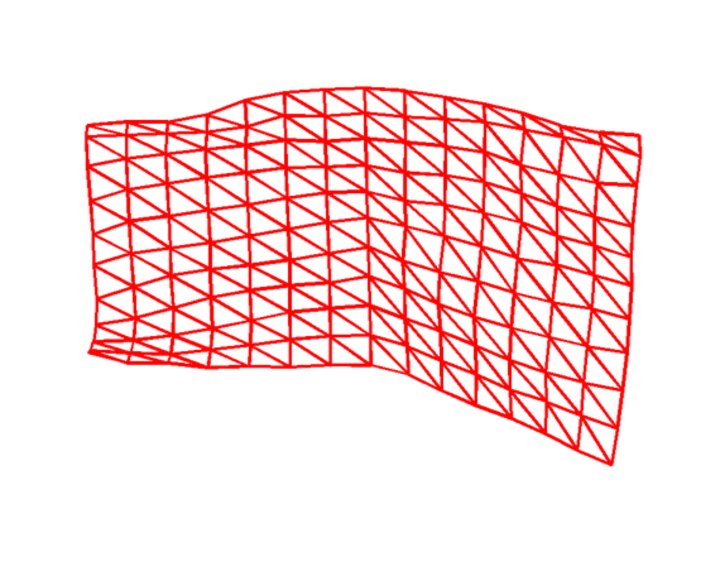}  &
\hspace{-0.1cm}\includegraphics[width=0.17\linewidth]{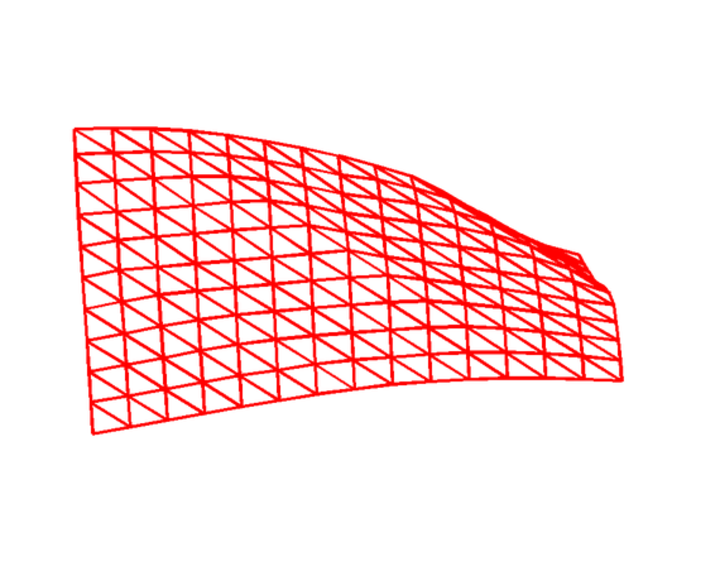}  &
\hspace{-0.1cm}\includegraphics[width=0.17\linewidth]{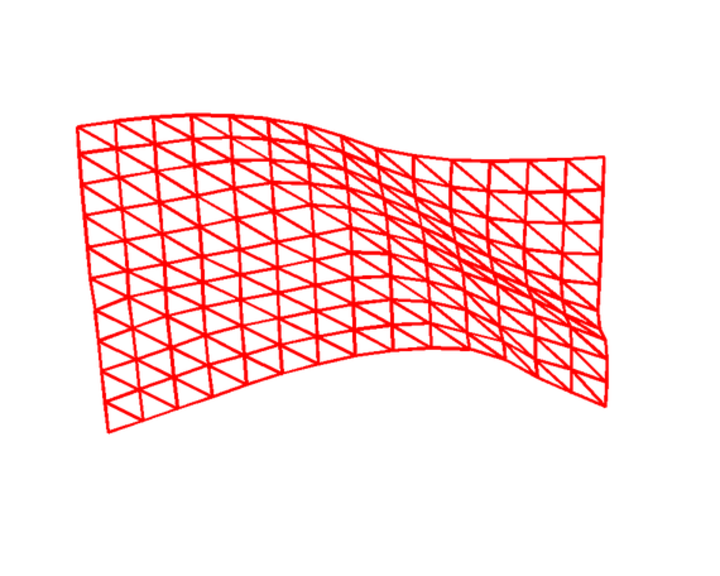} &
\hspace{-0.1cm}\includegraphics[width=0.17\linewidth]{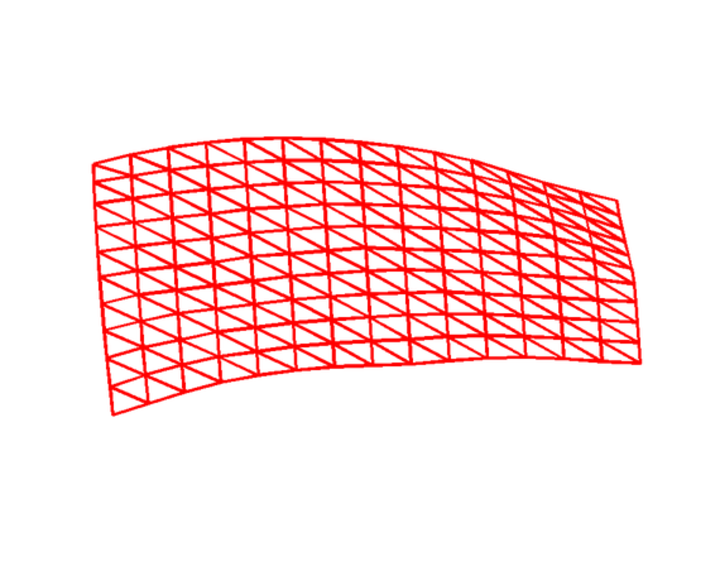}  &
\hspace{-0.1cm}\includegraphics[width=0.17\linewidth]{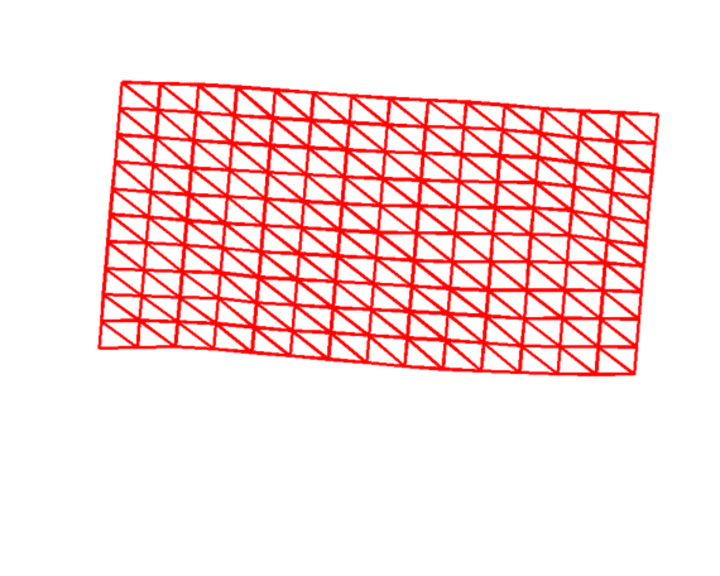}  
\end{tabular}
\end{center}
\vspace{-1cm}
\caption{{\bf Reconstructing surfaces from real images.} From top to bottom: Our reconstructions reprojected on the original images, side view of our reconstructions, reconstructions obtained with Shen09 CTF reprojected on the original images, side view of those reconstructions. For a well-textured surface, the baseline manages to reconstruct fairly large deformations, but is less consistent than our approach, as illustrated for two very similar frames. For a poorly-textured surface, the baseline only manages to reconstruct small deformations, whereas our approach can deal with much larger ones. The rightmost column shows a failure of our method due to an ambiguity in the facet reconstruction and to the use of a coarse mesh.}
\label{fig:tetris}
\vspace{-0.3cm}
\end{figure}

Finally, to show that our approach can also be applied to real images, we used two sequences of different deforming materials~\cite{DeformData}. While these are video sequences, all the images were treated independently and initialized from the template mesh to illustrate the fact that our approach can perform reconstruction from a single input image. Since no training data is available for these surfaces, we used a single training example consisting of the template mesh with reference image to learn the potential weights. In Fig.~\ref{fig:tetris}, we visually compare our reconstructions to those of Shen09. We do not show the results of Salz10, since with the template mesh as single training example, it would always predict the reference shape, and thus perform the same as Shen09. For the well-textured surface, Shen09 manages to reconstruct fairly large deformations. However, as illustrated by the two leftmost columns of the figure for two very similar frames, it is less consistent than our approach. For the poorly-textured surface, the baseline is completely unable to cope with large deformations. Our approach, however,  still manages to reconstruct the surface. In the rightmost column of the figure, we show a failure case of our approach, where the facet orientation is ambiguous. Furthermore, the topology of the coarse mesh makes it harder to bend the surface along this diagonal. Note, however, that as opposed to the baseline, we still recover some degree of surface deformation.

\section{The dual aspects of blending learning and inference}
\label{sec:duality}

CRFs and structured SVMs, as well as their one-parameter extension, are convex programs thus they have a dual program. Duality theory turned to be very effective in machine learning as it provides a principled way to decompose the different ingredients of the primal objective through its Lagrange multipliers. The dual decomposition in turn provides the means to efficiently estimate the different ingredients while maintaining their consistency using the dual objective.   

When dealing with convex programs one usually needs to consider the set of primal feasible solutions while constructing the dual function. We find it simpler to describe the primal program using extended real-valued convex functions, which are  functions that can get the value of infinity. Intuitively, by using extended real-valued functions we can ignore their domains, i.e., points for which a function gets the value of infinity, thus simplifying the derivations. The dual programs of extended real valued convex functions $g(\mu)$ are conveniently formulated in terms of their conjugate dual $$g^*(z) = \max_{\mu} \Big\{ \mu^\top z - g(\mu) \Big\}.$$ Throughout this work we use the following duality theorem, known as the  Fenchel duality (cf.  \cite{Fenchel51, Rockafellar70, Bertsekas03}):
\begin{theorem}
\label{theorem:dual}
Let $f_s: \R^Y \rightarrow \R$ and $h_t: \R \rightarrow \R$ be extended real-valued and convex functions, and let $a_{s,t},g_s$ be vectors of length $Y$, for every $s$. The following are primal and dual programs: 
\begin{eqnarray*}
&\mbox{(Primal)}& \min_{\nu} \hspace{0.5cm} \sum_s f_s \Big( \sum_t \nu_t  a_{s,t} + g_s \Big) - d^\top \nu + \sum_t h_t (-\nu_t) \\
&\mbox{(Dual)}& \max_{p} \hspace{0.5cm}  -\sum_s \Big(f^*(p_s) + p_s^\top g_s \Big) - \sum_t h_t^* \Big(\sum_s a_{s,t}^\top p_s - d_t \Big)  \\
\end{eqnarray*}
Strong duality holds if the functions satisfy $f_s(\mu_s), h_t(\nu_t) > -\infty$, their domains are defined with linear equalities and inequalities, they are continuous on their domains, their domains intersect and the primal optimal value is finite.
\end{theorem}
{\bf Proof:}  We use Lagrange duality theorem, minimizing the function  $\sum_s f_s(\mu_s + g_s) - d^\top \nu + \sum_t h_t (-\nu_t)$ subject to the constraints $\mu_s(y) = \sum_t \nu_t a_{s,t}(y)$. These equality constraints hold for every $y = 1,...,Y$, therefore correspond to Lagrange multipliers $p_s \in \R^Y$, for every $s$. The Lagrangian takes the form $$L(\mu,\nu,p) = \sum_s f_s(\mu_s + g_s) - d^\top \nu + \sum_t h_t (-\nu_t) - \sum_s p_s^\top \Big(\mu_s - \sum_t \nu_t a_{s,t} \Big).$$ By minimizing with respect to the primal variables $\min_{\mu,\nu} L(\mu,w,p)$ we get the dual function above. Strong duality holds by Theorems 6.2.5, 6.4.1, 6.4.2 in \cite{Bertsekas03} \eop  

The above duality theorem describes the relations between two types of functions through their conjugate dual functions. The learning problem in Equation (\ref{eq:reg-loss}) consists of two such functions, the loss function and the regularization. The extended log-loss is dominated by the normalizing constant of its loss adjusted Gibbs distribution, thus its conjugate dual is the entropy barrier function. The dual variables are then probability distributions $p_{(x,y)}(\hat y)$ and the dual program maximizes their entropy. The regularization consists of the square function, which is its own conjugate dual, therefore the learning dual tries to match the empirical moments $\sum_{(x,y)} \phi_k (x,y)$ using these probabilities. Hence the dual program for learning with extended log-loss balances between maximizing the entropy barrier function and fitting the moment matching constraints. 

\begin{corollary}
\label{cor:dual}
Let $\Delta_{\cal Y}$ be the probability simplex, i.e., the set of probability distributions over $\Y$. Define the entropy as a barrier function over the probability simplex, 
$$
\epsilon H(p) = \Bigg\{ 
\begin{array}{ll} 
-\epsilon \sum_y p(y) \log p(y) & \mbox{if} \;\; p \in \Delta_{\cal Y} \\
-\infty & \mbox{otherwise}
\end{array}
$$ 
Then following are primal and dual programs. 
\begin{eqnarray*}
&\mbox{(Primal)}& \min_{w} \sum_{(x,y) \in \S} \bar \ell_{\epsilon\mbox{-}log} (w,x,y) + \frac{C}{2} \|w\|_2^2 \\
&\mbox{(Dual)}& \max_{p_{(x,y)} \in \Delta_{\Y}} \sum_{(x,y) \in \S} \Big(\epsilon H(p_{(x,y)}) + \sum_{\hat y} p_{(x,y)}(\hat y) \ell(y, \hat y)  \Big) - \\ 
&&\hspace{3cm} \frac{1}{2C} \Big\| \sum_{(x,y) \in \S} \Big( \sum_{\hat y \in \Y} p_{(x,y)}(\hat y)\Phi(x,\hat y) - \Phi(x,y) \Big) \Big\|_2^2   
\end{eqnarray*}
In particular, $\ell_{\epsilon\mbox{-}log} (w,x,y)$ and $\|w\|_2^2$ satisfy the conditions in Theorem \ref{theorem:dual}, therefore strong duality holds. 
\end{corollary}
{\bf Proof:} Set $Z_\epsilon(w,x,y)$ to be the normalizing constant of $p_{(x,y)}(\hat y ; w, \epsilon)$. Thus the extended log-loss equals to $\log Z_\epsilon(w,x,y) - \sum_k w_k \phi_k(x,y)$. Thus the proof follows from Theorem \ref{theorem:dual} when setting $s=(x,y)$, and $\nu = w$, where the index $t$ is equivalent to the feature index $k$. Therefore $a_{s,k}(\hat y) = \phi_k(x, \hat y)$, $f_s(\sum_k w_k a_{s,k} + g_s) =  \log Z_\epsilon(w,x,y)$, $g_s(\hat y) = \ell(y,\hat y)$, $d_k = \sum_{(x,y)} \phi_k(x,y)$ and $h_k(-w_k) = w_k^2$, while noticing that the conjugate dual of $\log Z_\epsilon(w,x,y) $ is $\epsilon H(p_{(x,y)})$ (e.g., \cite{Wainwright08} Theorem 8.1) and the conjugate dual of $\frac{1}{2} w^2$ is $\frac{1}{2} z^2$ (e.g., \cite{Rockafellar70}, page 106). \eop \\

Both the primal and dual programs are well defined for $C=0$, where the primal regularization does not exist and the dual program enforces the moment matching as hard constraints. Generally, the parameter $C$ balances between the extended log-loss and the regularization. From dual perspective, the parameter $C$ balances between the entropy barrier function $\epsilon H(p)$ and the moment matching constraints.  One can observe that also the parameter $\epsilon$ balances between the entropy barrier and the moment matching. When considering learning with extended log-loss, restricted to $\ell(y, \hat y) \equiv 0$, the parameter $\epsilon$ affects the solution as $C$, since the dual program can be equivalently written as   
$$\ell \equiv 0 \hspace{0.5cm}  \Longrightarrow \hspace{0.5cm}  \epsilon \cdot \max_{p_{(x,y)}} \sum_{(x,y) \in \S} H(p_{(x,y)})  - \frac{1}{2C \epsilon} \Big\| \sum_{(x,y) \in \S} \Big( \sum_{\hat y \in Y} p_{(x,y)}(\hat y)\Phi(x,\hat y) - \Phi(x,y) \Big) \Big\|_2^2.$$ 
 
Restricting the dual program in Corollary \ref{cor:dual} to $\epsilon=1$ and $\ell(y,\hat y) \equiv 0$, it describes the well-known duality relation between the log-likelihood and the entropy, that is used in the context of CRFs by \cite{Lebanon02}. When $\epsilon=0$ we obtain the known dual formulation of structured SVM which emphasizes the duality between the max-function and the probability simplex (\cite{Taskar04, Tsochantaridis06, Collins08}). Thus, the seemingly different frameworks of CRFs and structured SVMs share the same moment matching perspective, and only differ by the selection rule for their probability distributions. Since these two formulations were proven to be successful in many cases of interest, we conclude that moment matching is important for learning the parameters of structured predictors . 


Considering structured labels $y=(y_1,...,y_n)$ over region graphs, the primal and dual learning programs complexities are exponential in $n$, as one needs evaluate the extended log-loss or the entropy barrier function respectively. However, the moment matching constraints, appearing in the dual program, are low dimensional and depend on the size of the regions. Thinking about the entropy barrier function as a selection rule that is independent of the moment matching constraints, we can reduce the complexity of the dual program. We match the moments with pseudo marginal probabilities, i.e., beliefs, while applying a low dimensional selection rule replacing the entropy barrier function.   

Restricting ourselves to graph based features, defined in Equation (\ref{eq:feature}), the moment matching constraints in the dual program of Corollary \ref{cor:dual} are taken with respect to the marginal probabilities, namely  
$$ \sum_{\hat y \in \Y} p_{(x,y)}(\hat y) \phi_k (x,\hat y) = \sum_{r \in {\cal R}} \sum_{\hat y_r \in \Y_r} p_{(x,y)}(\hat y_r) \phi_{k,r} (x,\hat y).$$
Thus the effective complexity of the moment matching constraints is the number of labels in a regions, namely $y_r \in \Y_r$. These averages can also be computed with beliefs $b_{(x,y),r}(y_r)$, i.e., probability distributions over regions labels that not necessarily come from a consistent distribution over all labels. To enforce local consistency between these averages we require these beliefs to agree on their overlapping labels. The selection rule we propose is the entropy barrier function for these beliefs. The pseudo moments matching provides the dual view for the low dimensional formulation for the extended log-loss that is described in Theorem \ref{theorem:upper}:  
\begin{theorem}
\label{theorem:approx}
The following are primal and dual programs. 
\begin{eqnarray*}
&\mbox{(Primal)}& \min_{w, \lambda} \sum_{(x,y) \in \S} \sum_{r \in {\cal R}} \bar \ell_{r,\epsilon\mbox{-}log} (w,x,y_r) + \frac{C}{2} \|w\|_2^2 \\
&\mbox{(Dual)}& \max_{b_{(x,y),r} \in \Delta_{\Y_r}} \sum_{(x,y) \in \S} \sum_{r \in {\cal R}} \Big(\epsilon H(b_{(x,y),r}) + \sum_{\hat y_r \in \Y_r} b_{(x,y),r}(\hat y_r) \ell_r(y_r, \hat y_r)  \Big) - \\ 
&&\hspace{3cm} \frac{1}{2C} \sum_k \Big( \sum_{(x,y) \in \S} \Big( \sum_{r \in {\cal R}_k} \sum_{\hat y_r \in \Y_r} b_{(x,y),r}(\hat y_r) \phi_{k,r}(x,\hat y_r) - \phi_k (x,y) \Big)\Big)^2 \\
&&\mbox{subject to}  \;\; \forall (x,y), r, \hat y_r, p \in P(r) \hspace{0.5cm}  b_{(x,y),r}(\hat y_r) = \sum_{\hat y_p \setminus \hat y_r} b_{(x,y),p}(\hat y_p)    
\end{eqnarray*}
Strong duality holds since  $\ell_{r,\epsilon\mbox{-}log} (w,x,y_r)$ and $\|w\|_2^2$ satisfy the conditions in Theorem \ref{theorem:dual}. 
\end{theorem}
{\bf Proof:} The proof follows from Theorem \ref{theorem:dual}, while we derive the primal program as the dual of the dual program. However, the indexing is a bit more involved. The index $s$ relates to the triplets of indexes $(x,y), r$, thus $g_s(\hat y_r) = \ell_r(y_r, \hat y_r)$.  The index $t$ either corresponds to a moment matching constraint index $k$ or to a marginalization constraint index $(x,y),r,y_r,p$. For $t=k$ we use $a_{s,t}(\hat y_r) = \phi_{k,r}(x,\hat y_r)$, and for $t = (x,y),r,\hat y_r,p$ we enforce marginalization constraints by setting $a_{s,t}(\hat y_p)=1$ if $s,t$ agree on the parent index in $t$ and $\hat y_p$ contains $\hat y_r$, and $a_{s,t}(\hat y_r)=-1$ if $s,t$ agree on the child index in $t$. When $t=k$ we set $d_k = \sum_{(x,y) \in \S} \phi_k(x,y)$ and zero otherwise. When $t=k$ we set $h_t^*(z) = \frac{1}{2}z^2$, whose conjugate dual is $h_t(w) = \frac{1}{2}w^2$. For $t=(x,y),r,\hat y_r,p$ we set $h_t^*(0) = 0$ and $h_t^*(z) = \infty$ otherwise, whose conjugate dual $h_t(\lambda) \equiv 0$. Since $f_s^*()$ is the entropy barrier function its conjugate dual is the normalizing constant of $b_{(x,y),r}(\hat y_r ; w, \lambda, \epsilon)$. We thus arrive to the final primal form, by adding to the linear term $d^\top \nu$ the quantity $$\sum_{r \in {\cal R}} \sum_{p \in P(r)} \lambda_{(x,y), r \rightarrow p}(y_r) - \sum_{r  \in {\cal R}} \sum_{c \in C(r)} \lambda_{(x,y), c \rightarrow r}(y_c) \equiv 0$$ which creates the numerator of the parametrized beliefs in the extended log-loss by multiplying with $\epsilon / \epsilon$ and exponentiating while taking the logarithm.
 \eop \\

Comparing the exteded log-loss formulation in Corollary \ref{cor:dual} to the low dimensional formulation in Theorem \ref{theorem:approx} we conclude that the difference between these two programs is in their probability models. In the low dimensional formulation we fit learning parameters $w$ to beliefs $b_{(x,y),r}(\hat y_r ; w, \lambda, \epsilon)$ whose local consistencies are governed by the inference variables $\lambda$. Using strong duality we are able to guarantee that the optimal beliefs are consistent with each other, since $\lambda$ are Lagrange multipliers of the marginalization constraints in the dual program. 

The connections of the inference variables $\lambda$ to the dual marginalization constraints suggest that the primal formulation in Theorem \ref{theorem:approx} is the objective function for approximate inference heuristics that recover parametrized beliefs that agree on their marginal probabilities, that are described in Section \ref{sec:bg-graphs}. These heuristics require running the norm-product belief propagation to convergence in order to obtain beliefs that agree on their marginal probabilities for updating the learning parameters $w$, thus they are computationally intractable. Our practical goal in this work is to overcome this computational difficulty and efficiently optimize this objective in large graphical models by blending learning and inference. For this purpose we show how to update the learning parameters $w$ using inferred beliefs that do not necessarily agree on their marginal probabilities throughout the algorithm run-time, but only when it converges. 

Strong duality holds for the pseudo moment matching and its corresponding low dimensional extended log-loss formulation. Therefore, one can either minimize the primal or maximize the dual to get the same results. Nevertheless, there are computational differences between these programs. The dual program is constrained and requires (sub)gradient descent methods that consider all variables. In contrast, the primal program is unconstrained, and one can perform block coordinate descent on its variables. Coordinate descent methods are appealing as they optimize small number of variables while holding the rest fixed, therefore they can be performed efficiently and can be easily parallelized. Moreover, coordinate descent for the primal program in Theorem \ref{theorem:approx} can be performed by sending messages over its region graph, thus can be efficiently applied to learn parameters of large graphical models. This approach is described in Figure \ref{fig:alg}.

\section{Extensions: entropies, regularizations and the penalty method}
\label{sec:ext}

Learning with low dimensional extended log-loss consists of two conjugate dual functions, one fits the moments and the marginalization constraints and the other provides a selection rule. Using these functions we are able to solve it efficiently while blending learning and inference. In this section we  extend our framework while maintaining the computational efficiency of our message-passing algorithm. 

We fit the moments using the square function. Since this function is strictly convex, its conjugate dual is smooth, thus we are able to perform a gradient descent step to optimize the learning parameters $w$. The selection rule we apply consists of entropies over region labels. This selection rule is frequently referred as an entropy approximation since it replaces the entropy function over all labels. Whenever the region graph is bipartite and without cycles, the Gibbs distribution can be described by its marginal probabilities $p(y_r) =  \sum_{y \setminus y_r} p(y)$, as described in Equation (\ref{eq:pr}). Therefore the entropy can be equivalently described by a weighted sum of local entropies $H(p) = \sum_r (1-|P(r)|)H(p(y_r))$, called the Bethe entropy. More generally, one can use fractional entropy approximation 
\begin{equation}
\label{eq:H}
H(p) \approx \sum_r c_r H(p(y_r)).
\end{equation}
The introduction of general functions for fitting the constrains and fractional entropy approximations selection rules to learn structured predictors parameters with low dimensional loss functions provides the following primal and dual programs: 
\begin{theorem}
\label{theorem:approx-gen}
Every nonnegative numbers $c_r \ge 0$, that fractionally cover $i=1,..,n$, namely $\sum_{r: i \in r} c_r \ge 1$ imply an upper bound for the extended log-loss, i.e., 
$$\bar \ell_{\epsilon\mbox{-}log} (w,x,y)  \le \sum_{r \in {\cal R}} \bar \ell_{r,\epsilon c_r \mbox{-}log} (w,x,y_r).$$ 
Also, for every $c_r \ge 0$ strong duality holds for the following primal and dual programs. 
\begin{eqnarray*}
&\mbox{(Primal)}& \min_{w, \lambda} \sum_{(x,y) \in \S} \sum_{r \in {\cal R}} \bar \ell_{r,\epsilon c_r \mbox{-}log} (w,x,y_r) + \sum_k h_1(-w_k)  + \sum_{(x,y) \in \S} \sum_{r,\hat y_r, p \in P(r)} h_2(\lambda)  \\
&\mbox{(Dual)}& \max_{b_{(x,y),r} \in \Delta_{\Y}} \sum_{(x,y) \in \S} \sum_{r \in {\cal R}} \Big(\epsilon c_r H(b_{(x,y),r}) + \sum_{\hat y_r} b_{(x,y),r}(\hat y_r) \ell_r(y_r, \hat y_r)  \Big) - \\ 
&&\hspace{2cm} \sum_k h^*_1 \Big( \sum_{(x,y) \in \S} \Big( \sum_{r \in {\cal R}_k} \sum_{\hat y_r \in \Y_r} b_{(x,y),r}(\hat y_r) \phi_{k,r}(x,\hat y_r) - \phi_k (x,y) \Big)\Big)  - \\
&&\hspace{2cm} \sum_{(x,y) \in \S} \sum_{r \in {\cal R}} \sum_{\hat y_r \in \Y_r} \sum_{p \in P(r)} h^*_2 \Big( \sum_{\hat y_p \setminus \hat y_r} b_{(x,y),p}(\hat y_p) - b_{(x,y),r}(\hat y_r)  \Big)   
\end{eqnarray*}
\end{theorem}
{\bf Proof:} The proof follows the same lines as Theorem \ref{theorem:upper} and Theorem \ref{theorem:approx}. Using the notation in Theorem \ref{theorem:upper} the extended log-loss upper bounds reduce to the following upper bounds:  
$$\sum_{\hat y} \prod_{r \in {\cal R}}  \exp(\theta_{(x,y),r}(\hat y_r ; w, \lambda)) \le \prod_{r \in {\cal R}} \Big( \sum_{\hat y_r} \exp(\theta_{(x,y),r}(\hat y_r ; w, \lambda) / \epsilon c_r) \Big)^{\epsilon c_r}$$
These upper bounds were previously described by \cite{Hazan12-uai}, Theorem 1. These bounds are based on fractional bounds to the entropy function by \cite{Friedgut04, Madiman10}.  
Strong duality follows along the same line as in Theorem \ref{theorem:approx}, while the differences are in the functions $f_s(\cdot)$ and $h_t(\cdot)$. We use the triplets $s=((x,y),r)$ and $f_s(b)$ is the log-partition function of $b_{(x,y),r}(\hat y_r ; w, \lambda, \epsilon c_r)$. For notional convenience we use the same regularization function for moment matching, i.e., $h_1(w)$, and for marginalization constraints, i.e., $h_2(\lambda)$.
 \eop \\
 
The above formulation describes a duality relation between the penalty functions $h^*_t(\cdot)$, for fitting the moments and marginalization constraints, and  the regularization $h_t(\cdot)$. Thus, one can use different penalty functions to influence the properties of the primal program. For example, one can use known relations between the weight of the penalty function to deduce the appropriate (dual) weight of the regularization function (cf. \cite{Bertsekas03} Section 5.5). Also, one can choose the penalty function to be strongly concave with Lipschitz continuous gradient, for which the primal block gradient descent has linear convergence rate. 

 
Using the fractional entropy approximations as our selection rule does not affect the computational properties of the programs, since its maximizing arguments, and hence the gradient of the extended log-loss, are beliefs which are (weighted) Gibbs distributions. The Gibbs distributions are computationally favorable since they have a log-linear form (e.g., Equation (\ref{eq:gibbs})) thus they provide an analytical block coordinate update rule, regardless of the the weights $c_r$. Moreover, since the block coordinate descent iterates over points with vanishing gradients it can also explain algorithms for optimizing weights with mixed signs, as happens with the Bethe entropy and the tree re-weighted entropy. Therefore we are able to extend the low dimensional learning framework while providing efficient message-passing algorithms: 
\begin{theorem}
\label{theorem:alg-gen}
Consider the primal and dual programs in Theorem \ref{theorem:approx-gen} with smooth regularization $h_k(-w_k)$, $h_t(\lambda) \equiv 0$, and the following algorithm:
\begin{eqnarray*}
&& \hspace{-0.8cm} \mu_{(x,y), p \rightarrow r}(\hat y_r) = \epsilon c_p \log \Big(\sum_{\hat y_p \setminus \hat y_r} \exp \big( (\theta_{(x,y),p}(\hat y_p; w) + \sum_{c \in C(p) \setminus r} \lambda_{(x,y),c \rightarrow p} (\hat y_c) - \sum_{p' \in P(p)} \lambda_{(x,y),p \rightarrow p'}(\hat y_p)) \big/ \epsilon c_p \big) \Big) \\ 
&& \hspace{-0.8cm}  \lambda_{(x,y),r \rightarrow p} (\hat y_r) =  \frac{c_p}{c_r + \sum_{p' \in P(r)} c_{p'}} \Big(\theta_{(x,y),r}(\hat y_r; w)  + \sum_{c \in C(r)} \lambda_{(x,y),c \rightarrow r}(\hat y_c) + \sum_{p' \in P(r)} \mu_{(x,y),p' \rightarrow r} (\hat y_r)  \Big) - \mu_{(x,y), p \rightarrow r}(\hat y_r) \\
&& \hspace{-0.8cm}  w_k =  w_k - \eta \Big(\sum_{(x,y) \in \S} \sum_{r \in {\cal R}}  \Big(\sum_{\hat y_r} b_{(x,y),r}(\hat y_r ; w, \lambda, \epsilon c_r) \phi_{k,r}(x,\hat y_r) - \phi_{k,r}(x,y_r) \Big) + \nabla h_k(-w_k)\Big)
\end{eqnarray*}
Whenever $\epsilon, c_r \ge 0$ these update rules monotonically decrease the primal objective, thus they are guaranteed to converge. Whenever $\epsilon, c_r > 0$ the beliefs generated by this algorithm converge to a dual optimal solution and the primal and dual objectives converge to their optimal values. Moreover, if $w, \lambda$ converge, then their limit point is a primal optimal solution. Whenever $\epsilon, c_r \gtrless 0$ the algorithm is not guaranteed to converge, but whenever it converges it recovers stationary points for both programs.    
\end{theorem}
{\bf Proof:} The proof follows the ones in Lemmas \ref{lemma:lambda}, \ref{lemma:theta} with some modifications. The primal program is unconstrained, therefore we describe the points for which the gradient vanishes. The gradient with respect to $\lambda_{(x,y),r \rightarrow p}(\hat y_p)$ relate to the disagreements of the marginal beliefs $\sum_{\hat y_p \setminus \hat y_r} b_{(x,y),p}(\hat y_p ; w, \lambda, \epsilon c_p ) - b_{(x,y),r}(\hat y_r ; w, \lambda, \epsilon c_r)$. Thus the gradient vanishes when  
$$\frac{\mu_{(x,y),p \rightarrow r}(\hat y_r) + \lambda_{(x,y),r \rightarrow p}(\hat y_r)}{c_p} = \frac{\theta_{(x,y),r}(\hat y_r; w) + \sum_{c \in C(r)} \lambda_{(x,y),c \rightarrow r}(\hat y_c)  - \sum_{p \in P(r)} \lambda_{(x,y),r \rightarrow p}(\hat y_r)}{c_r}$$
Multiplying both sides by $c_r c_p$ and summing both sides with respect to $p' \in P(r)$ we are able to isolate $\sum_{p' \in P(r)} \lambda_{(x,y),r \rightarrow p'}(\hat y_r)$. Plugin it into the above equation results in the desired inference update rule, i.e., $ \lambda_{(x,y),r \rightarrow p}(\hat y_r) $ for which the partial derivatives vanish. The convergence for $\epsilon, c_r \ge 0$ is guaranteed since the primal program is lower bounded by its dual. The optimality results for $\epsilon, c_r > 0$ are achieved by applying \cite{Tseng87}. Whenever $\epsilon, c_r \gtrless 0$ if the algorithm converges the primal gradient vanishes thus it recovers a primal stationary point. Considering the Lagrangian of the dual, given the Lagrange multipliers $\lambda, w$ their corresponding beliefs $b_{(x,y),r}(\hat y_r ; w, \lambda,  \epsilon c_r)$ satisfy the marginalization constraints, therefore we also recover a stationary point for the dual. \eop

The above theorem generalizes the learning-inference blending algorithm for low dimensional structured prediction in Fig. \ref{fig:alg}, that is attained by setting $c_r=1$. This provides a way to explain different heuristics for learning  structured predictors in graphical models. For example, setting the Bethe coefficients $c_r = 1- |P(r)|$ amounts to minimizing a non-convex program using belief propagation to approximate the marginal probabilities. Whenever the solver converges we are able to match the moments and expect the resulting learning parameters $w$  to be good in practice. This can explain the success of belief propagation when applied as a heuristic to estimate the marginal probabilities, since when it converges its beliefs agree on their marginal probabilities. However, since non-convex programs are harder to optimize, the algorithm might not result in beliefs that fit the moments and agree on their marginal probabilities. This in turn may explain the failures of the belief propagation heuristic as it is not guaranteed to converge and its resulting beliefs not necessarily agree on their marginal probabilities.

\section{Related work}

In this work we learn the parameters of region based structured predictors using pseudo moment matching and entropy approximations, or equivalently low-dimensional extended log-loss. We also construct an inference-learning blending algorithm and show how it achieves state-of-the-art results in stereo estimation (\cite{Yamaguchi12}), semantic segmentation (\cite{Yao12}), shape reconstruction (\cite{Salzmann12}), and indoor scene understanding (\cite{Schwing12-cvpr}). We also provide an efficient C++ implementation with a Matlab wrapper. This work extends the framework of \cite{Hazan10-nips} while simplifying its theoretical and practical concepts. Theoretically, it extends \cite{Hazan10-nips} to general region graph, introduces the notion of extended log-loss and investigates the penalty method in message-passing. Practically, it emphasizes the importance of graph based predictors and show how to use them to achieve state-of-the-art results in several computer vision applications.  

The extended log-loss reduces to the hinge-loss, described by \cite{Taskar04, Tsochantaridis06}, and the log-loss of \cite{Sha07,Gimpel10}. This extension is described by \cite{Hazan10-nips, Pletscher10}, using a linear term and a temperature parameter in the logarithm of the partition function, which is referred as soft-max (cf. \cite{Vontobel06, Johnson07, Hazan10}). In this work we present the extended log-loss as the logarithm of the inferred beliefs.  

The inference techniques we are using in this work rely on the region graph message-passing of \cite{Heskes06} and the region norm-product algorithm of \cite{Hazan12-uai}. Section \ref{sec:bg-graphs} describes how to use these inference algorithms in a black-box manner while learning, and their computational disadvantages. The main purpose of this work is to blend the inference and learning in general graphical models.  

The extended log-loss minimization, as appears in Corollary \ref{cor:dual}, reduces to CRFs when setting $\epsilon=1$ and $\ell(y, \hat y) \equiv 0$. CRFs, defined by \cite{Lafferty01, Lebanon02}, are widely applied in machine learning. Whenever the labels are in a discrete product space, i.e., $y=(y_1,...,y_n)$, the gradient is exponentially hard to compute. Although approximate inference techniques can be used to estimate the gradient (e.g., \cite{Levin06, Yanover07}), as described in Section \ref{sec:bg-graphs}, this approach requires running an approximate inference algorithm for every gradient step thus it is computationally intractable in general. To apply CRFs efficiently in discrete product spaces, some works focus on the practical aspects of low dimensional approximations for learning CRFs parameters. \cite{Sutton09} present the piecewise training approach which uses a low dimensional log-loss while ignoring the consistency messages $\lambda$. In our setting, these messages are used to enforce the marginalization constraints of the dual program. \cite{Ganapathi08} approximate CRFs using the non-concave Bethe entropy and execute a double loop algorithm while showing that it requires only few outer-loop iterations. The inner loops of this algorithm use concave entropy approximations, and are computed using BFGS. Our work is different as it blends the inference and learning through convexity. We show in the experiments that this significantly improves the run-time of the algorithm and the quality of the solution. Other works focus on the theoretical aspects of fractional entropy approximations. \cite{Wainwright06} prove that whenever one wishes to learn parameters jointly with prediction, it is preferred to choose concave entropy approximations since their parameters are stable with respect to their prediction. The theoretical foundations of concave entropy approximations in parameter learning and pseudo moment matching appear in  \cite{Wainwright03}. Our  work provides a detailed derivation of the primal and dual programs for graph based features. In addition, our work blends the learning and inference updates, thus is able to attain state-of-the-art results in several computer vision applications. 

The extended log-loss minimization, as appears in Corollary \ref{cor:dual}, reduces to structured SVMs for $\epsilon=0$. Structured SVMs are defined in \cite{Taskar04, Tsochantaridis04} and are motivated by the structured perceptron of \cite{Collins02}. Structured SVMs are very popular in machine learning, but in general the gradient computation requires to solve the max-function over a discrete product space, thus it is NP-hard to compute. Therefore in general setting, straight forward subgradient methods, such as \cite{Collins02, Roth05, Ratliff07, Shalev07}, cannot be applied as they require to solve a NP-hard problem for each gradient step. Instead, one can use a low dimensional max-function, while ignoring the consistency messages $\lambda$ (e.g., \cite{Punyakanok05}). We use these messages to enforce the marginalization constraints and in general it results in better performance. Alternatively, one can relax the NP-hard max-function in a similar manner to our dual program, introducing beliefs and marginalization constraints. For $\epsilon=0$ this approach boils down to running a linear program solver for every gradient step, e.g., \cite{Kulesza07, Finley08}, thus it is computationally intractable in general. To relax the computational burden of the max-function, \cite{Tsochantaridis06} use the cutting plane method. It was shown that the number of added constraints is polynomial (e.g., \cite{Joachims09}), but in practice finding a cutting plane may be hard and the number of added constraints may be large. Taskar and collaborators use a different approach to deal with the computational complexity of the max-function. Specifically, they consider structured SVM which can be solved efficiently using Lagrange multipliers and duality. \cite{Taskar04} introduce the structured SMO to solve the structured SVM dual. In our work we avoid solving the dual program since it is a constrained optimization problem. In contrast, our primal program is unconstrained thus can usually be solved faster using block coordinate steps, and can be distributed and parallelized easily, as described in \cite{Schwing11}. \cite{Taskar05, Anguelov05} consider the hinge loss conjugate dual and integrate it into the primal program, thus effectively replacing the max-function of the hinge-loss with a min-function. Although these formulations are applied to settings for which the maximum can be computed efficiently (e.g., associative networks and matchings), this dual-primal concepts play an important role in our learning and inference blending derivation of the low dimensional primal program. \cite{Meshi10} further improve this idea and their primal program as well as their algorithm are similar to our primal program and message-passing algorithm. Specifically, their inference step apply the norm-product belief propagation update rule thus also utilize the soft-max function. However, their setting is restricted to pairwise graphical models while we consider graph-based features for general regions, which are important when applying these methods to real-life problems. Their parameter learning update for the $w$ parameters, motivated by the case $\epsilon=0$, is using subgradient steps (\cite{Meshi10}, Section 3.2). Despite the theoretical guarantees of subgradient methods, these methods tend to be slow in many cases of interest and for different settings of $C$, e.g, \cite{Shalev07}. Restricting to $
\epsilon=0$, it is hard to verify the subgradient algorithm reaches an optimal solution: since the max-function is non-smooth, it is hard to recover a dual feasible solution to verify a small duality gap. In general, we find that setting $\epsilon>0$ gives better results faster than setting $\epsilon=0$, and it might be better to solve structured SVM without the shortcomings of subgradient methods by simply setting $\epsilon \rightarrow 0$.

\section{Conclusion and Discussion}
\label{sec:discussion}

In this paper we have related CRFs and structured SVMs through the extended log-loss formulation, thus showing how CRFs approximate smoothly the structured SVMs.We have also proposed low dimensional loss formulation which decomposes according to general regions in a graphical model, and its dual program corresponds to pseudo moments matching and fractional entropy approximations. We have derived an efficient message-passing algorithm for learning the parameters of graph based structured predictors and have demonstrated the effectiveness of our approach, achieving state-of-the-art results in several computer vision applications. We believe it is interesting to show in the future if this algorithm provides state-of-the-art performance in domains other than computer vision, or whether the statistics in computer vision are used by this approach in a special manner.  

The computational complexity of our algorithm depends on norm-products over the labels of regions. Therefore, efficient techniques over large regions in inference can be applied as sub-procedure in our algorithm, e.g., \cite{Kohli09, Batra10, Tarlow10, Tarlow11, Tarlow12}.   

The extended log-loss introduces a weight parameter $\epsilon$ which controls the characteristics of the loss, e.g., for $\epsilon=0$ we recover the hinge loss for structured SVMs and for $\epsilon=1$ we recover the log-loss for CRFs. The learning program also considers a constant $C$ which controls the tradeoff between the extended log-loss and the regularization. We have shown that whenever the true loss is equivalently zero, these two parameters influence equally the learned parameters, and an important open problem is their influence in general loss settings. 

In our framework, we can enforce the moment matching constraints through general concave functions. These function translate to a regularization in the primal. For computational efficiency we choose the square function but we did not investigate the different moment matching and regularization functions. Moreover, we enforce the marginalization constraints through indicator functions, in order to obtain closed-form solution in the primal block coordinate descent. However, we have shown that using the penalty method we can enforce the marginalization constraints with different convex functions. We leave the affect of general convex functions on moment matching and regularization, as well as marginalization constraints and efficient message-passing for future research.

Interestingly, our approach confirms that the parameters of graph based structured predictors can be efficiently learned in many real-life problems. This validates the intuition behind the theoretical results of \cite{Wainwright03, Wainwright06} which asserts that whenever learning and inference occur together one can use pseudo moment matching for learning the parameters. This concept was put forward in the general framework of learning to reason by \cite{Roth97} and we leave for future research to find different frameworks which have similar learning-prediction robustness that such algorithms might be effective.


\bibliography{blend-jmlr13}

\end{document}